\documentclass[10pt,journal,compsoc]{IEEEtran}

\ifCLASSOPTIONcompsoc
  \usepackage[nocompress]{cite}
\else
  \usepackage{cite}
\fi

\usepackage{xcolor}
\usepackage{graphicx}
\usepackage{amsmath}
\usepackage{acronym}
\usepackage{amssymb}
\usepackage{algorithmic}

\usepackage{subfigure}
\usepackage{subfloat}
\usepackage{multirow}
\usepackage{booktabs}

\usepackage{makecell}
\usepackage{array}
\usepackage{url}
\usepackage{commath}
\usepackage{textcomp}
\usepackage{siunitx}

\usepackage[pagebackref=true,breaklinks=true,colorlinks,bookmarks=false]{hyperref}
\usepackage{marvosym}
\usepackage[normalem]{ulem}
\usepackage[symbol]{footmisc}

\usepackage{xspace}
\usepackage{ragged2e}
\usepackage[labelfont=bf, font=footnotesize, justification=justified, format=plain]{caption}

\usepackage{rotating}

\makeatletter
\DeclareRobustCommand\onedot{\futurelet\@let@token\@onedot}
\def\@onedot{\ifx\@let@token.\else.\null\fi\xspace}

\def\eg{\emph{e.g}\onedot} 
\def\ie{\emph{i.e}\onedot}

\def\etal{\emph{et al}\onedot}
\makeatother

\definecolor{ao}{rgb}{0.0, 0.5, 0.0}
\definecolor{blue}{rgb}{0.0, 0.0, 0.0} 
\usepackage{colortbl}
\definecolor{seen}{RGB}{255,255,179}
\newcommand{\seen}[0]{\cellcolor{seen} }

\newcommand{\numdatasets}{40\xspace}

\newcommand{\name}{SMPLer-X\xspace}
\newcommand{\namepami}{SMPLest-X\xspace}
\newcommand{\synhand}{SynHand\xspace}
\newcommand{\Fig}{Fig.\xspace}
\newcommand{\Tab}{Table\xspace}
\newcommand{\Sec}{Sec.\xspace}
\newcommand{\Supp}{Supplementary Material}

\begin{document}
\title{\namepami: Ultimate Scaling for Expressive \\ Human Pose and Shape Estimation}


\author{Wanqi Yin$^{\star}$ \quad
Zhongang Cai$^{\star}$ \quad
Ruisi Wang \quad
Ailing Zeng \quad
Chen Wei \\
Qingping Sun \quad
Haiyi Mei \quad
Yanjun Wang \quad
Hui En Pang \quad
Mingyuan Zhang \\
Lei Zhang \quad
Chen Change Loy \quad
Atsushi Yamashita \quad
Lei Yang$^{\href{mailto:yanglei@sensetime.com}{\textrm{\Letter}}}$ \quad
Ziwei Liu$^{\href{mailto:ziwei.liu@ntu.edu.sg}{\textrm{\Letter}}}$%
\IEEEcompsocitemizethanks{
    \IEEEcompsocthanksitem $\star$ indicates equal contributions.
    \IEEEcompsocthanksitem The corresponding authors are Lei Yang: yanglei@sensetime.com and Ziwei Liu: ziwei.liu@ntu.edu.sg
}%
}

\markboth{Journal of \LaTeX\ Class Files,~Vol.~XX, No.~XX, MM~YYYY}%
{Shell \MakeLowercase{\textit{et al.}}: Bare Advanced Demo of IEEEtran.cls for IEEE Computer Society Journals}


\IEEEtitleabstractindextext{
\begin{abstract}
\justifying
Expressive human pose and shape estimation (EHPS) unifies body, hands, and face motion capture with numerous applications. Despite encouraging progress, current state-of-the-art methods focus on training innovative architectural designs on confined datasets. 
\textcolor{blue}{In this work, we investigate the impact of scaling up EHPS towards a family of generalist foundation models.}
\textit {1) For data scaling}, we perform a systematic investigation on \textcolor{blue}{\numdatasets} EHPS datasets, encompassing a wide range of scenarios that a model trained on any single dataset cannot handle. More importantly, capitalizing on insights obtained from the extensive benchmarking process, we optimize our training scheme and select datasets that lead to a significant leap in EHPS capabilities. Ultimately, we achieve diminishing returns at \textcolor{blue}{10M} training instances from diverse data sources.
\textit {2) For model scaling,} we take advantage of vision transformers (up to ViT-Huge as the backbone) to study the scaling law of model sizes in EHPS. To exclude the influence of algorithmic design, we base our experiments on \textcolor{blue}{two minimalist architectures: \textbf{\name}, which consists of an intermediate step for hand and face localization, and \textbf{\namepami}, an even simpler version that reduces the network to its bare essentials and highlights significant advances in the capture of articulated hands.}
With big data and the large model, the foundation models exhibit strong performance across diverse test benchmarks and excellent transferability to even unseen environments. Moreover, our finetuning strategy turns the \textit{generalist} into \textit{specialist} models, allowing them to achieve further performance boosts.
Notably, our foundation models consistently deliver state-of-the-art results on seven benchmarks such as AGORA, UBody, EgoBody, \textcolor{blue}{and our proposed SynHand dataset for comprehensive hand evaluation. (Code is available at: \url{https://github.com/wqyin/SMPLest-X}).}

\end{abstract}
}


\maketitle
\IEEEdisplaynontitleabstractindextext
\IEEEpeerreviewmaketitle

\section{Introduction}

The innovative realm of expressive human pose and shape estimation (EHPS) from monocular images or videos offers transformative applications for the animation, gaming, and fashion industries. This task typically uses parametric human models (\eg, SMPL-X \cite{pavlakos2019expressive}) as a powerful representation of the human body, face, and hands. With a flurry of diverse datasets entering the scene in recent years~\cite{black2023bedlam,cai2024playing,yang2023synbody,zhang2022egobody,lin2023one,bhatnagar2022behave,fan2023arctic,fieraru2020three,yi2023generating,renbody}, providing the community new opportunities to study various aspects such as capture environment, pose distribution, body visibility, and camera views. Yet, the state-of-the-art methods channel their attention towards advancements in architectural designs and remain tethered to a limited selection of these datasets, creating a bottleneck in performance across varied scenarios and hindering the ability to generalize to unseen situations. 

Our first mission in this study is to explore existing data resources extensively, providing key insights crucial for establishing robust, universally applicable models for EHPS. Accordingly, we establish a systematic benchmark for EHPS, utilizing \numdatasets datasets and evaluating their performance across five major benchmarks. Our study reveals that there are significant inconsistencies among benchmarks, revealing the overall complicated landscape of EHPS, and calling for data scaling to combat the domain gaps between scenarios. 
This detailed examination emphasizes the need to reassess the utilization of available datasets for EHPS, advocating for a paradigm shift towards a more inclusive data selection that offers better in-the-wild performances. 
Particularly, we propose a new metric, Mean Primary Error (MPE) \textcolor{blue}{for both whole-body and hand pose}, to gauge model performance across a basket of key benchmarks, in order to provide a holistic measurement of generalization capabilities. Our study underscores the importance of harnessing a multitude of datasets to capitalize on their complementary nature.
\textcolor{blue}{Moreover, we contribute a new dataset, SynHand, to provide the community with a long-awaiting benchmark for comprehensive hand pose evaluation in a whole-body setting. SynHand features diverse hand poses in close-up human shots, accurately annotated as part of the whole-body SMPL-X labels.}
%



\begin{figure*}
  \centering
  \includegraphics[width=\linewidth]{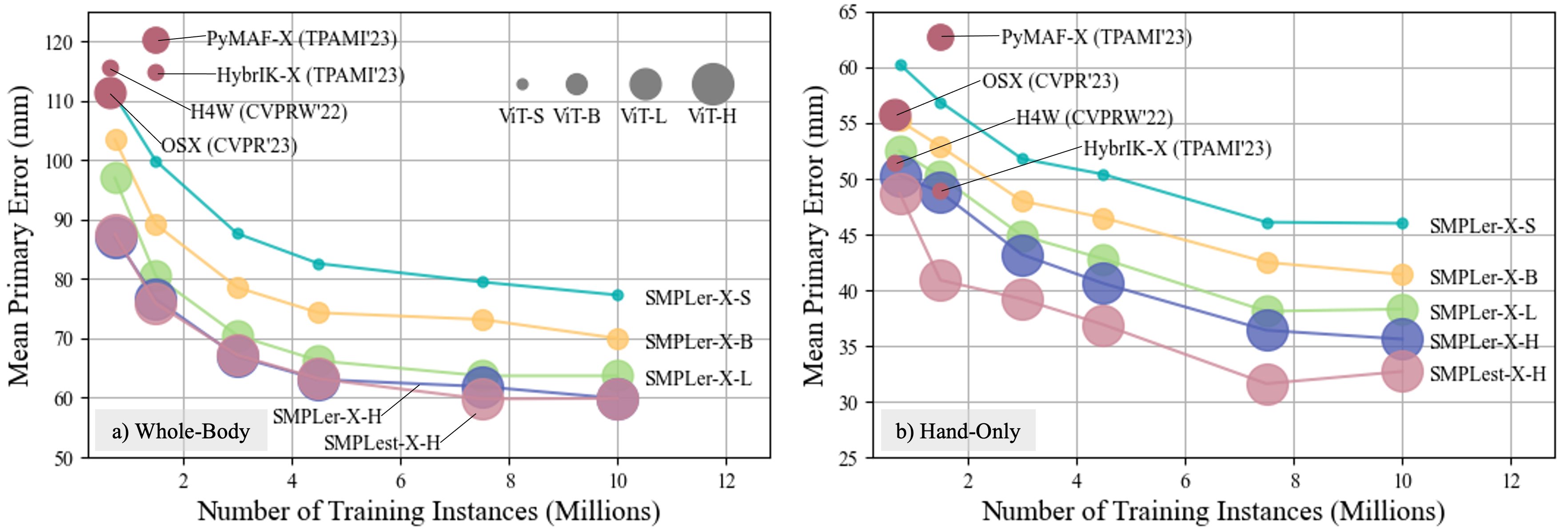}
  \caption{\textbf{Scaling up EHPS.} a) Whole-body and b) hand-only mean primary error (MPE) indicate both data and model scaling are effective in reducing mean errors on primary metrics across key benchmarks for: AGORA~\cite{patel2021agora}, UBody~\cite{lin2023one}, EgoBody~\cite{zhang2022egobody}, 3DPW~\cite{von2018recovering} and EHF~\cite{Pavlakos_2019smplx}. OSX~\cite{lin2023one} and HybrIK-X~\cite{li2023hybrik} are SOTA methods. Area of the circle indicates model size, with ViT variants as the reference (top right in the left figure).}
  \label{fig:teaser}
\end{figure*}

Equipped with the knowledge acquired from the benchmarks, we proceed to exhibit the strength of massively selected data in crafting \textcolor{blue}{\name and \namepami, a family of} \textit{generalist} foundation models that are trained using a diverse range of datasets and achieve exceptionally balanced results across various scenarios. To decouple from algorithmic research work, we intentionally design \textcolor{blue}{\name and \namepami} with a minimalist mindset: they are reduced to very simple architectures. \textcolor{blue}{Between them, \name opts for a relatively conventional intermediate step for component (hands and face) guiding, while \namepami, as its name suggests, only retains the ultimately essential components for EHPS (an encoder, a decoder, and regression heads). 
Despite that \name and \namepami are designed to facilitate massive data and parameter scaling instead of a rigorous investigation into the algorithmic aspect, we highlight that their simplicity does not undermine their overall performance, and surprisingly, \namepami even demonstrates a special edge on hand estimation in a one-stage manner.
}

Experiments with various data combinations and model sizes lead us to a well-rounded model that excels across all benchmarks that contests the community norm of limited-dataset training. Notably, our foundation models demonstrate significant performance boost through both data scaling and model size scaling, reducing the whole-body mean primary errors on five major benchmarks (AGORA~\cite{patel2021agora}, UBody~\cite{lin2023one}, EgoBody~\cite{zhang2022egobody}, 3DPW~\cite{von2018recovering}, and EHF~\cite{Pavlakos_2019smplx}) from over 110 mm to below \textcolor{blue}{60 mm and the hand primary errors on six major benchmarks from over 62 mm to 31 mm} (demonstrated in Fig.~\ref{fig:teaser}), and showcases impressive generalization capabilities by effectively transferring to new scenarios, such as DNA-Rendering~\cite{renbody} and ARCTIC~\cite{fan2023arctic}. Furthermore, we validate the efficacy of finetuning our \textit{generalist} foundation models to evolve into domain-specific \textit{specialists}, delivering outstanding performance on all benchmarks. Specifically, we follow the same data selection strategy that empowers our specialist models to set new records on the AGORA leaderboard by being the first model to hit \textcolor{blue}{96.2 mm} in NMVE, \textcolor{blue}{31.1 mm in hand MVE} and achieving SOTA performance on key benchmarks such as EgoBody and UBody.

\textcolor{blue}{
Our contributions are three-fold. 
\textbf{1)} We systematically study existing EHPS datasets. This includes building the first systematic benchmark on EHPS datasets to provide critical guidance for scaling up the training data toward robust and transferable EHPS, and proposing \synhand as a better whole-body benchmark that focuses on hand evaluation.  
\textbf{2)} We design minimalistic foundation models (\name and \namepami) that are tailored for scaling up. Despite their simplicity, they demonstrate capabilities to achieve strong whole-body performance and \namepami further shows promising hand performance in a pure one-stage manner.
\textbf{3)} We explore the impact of scaling in EHPS. As a result, we build \textit{generalist} foundation model that delivers balanced, strong results across various scenarios and extends successfully to unseen datasets, finetuning further elevates our foundation model into a potent \textit{specialist} across a multitude of key benchmarks.
}
\begin{table*}
  \caption{\textbf{Benchmarking EHPS datasets.} For each dataset, we train a model on its training set and evaluate its performance on the \textit{testing} sets of five major benchmarks: AGORA, UBody, EgoBody (EgoSet), 3DPW, and EHF. Datasets are then ranked by mean primary error (MPE). Top-1 values are bolded, and the rest of Top-5 are underlined. \#Inst.: number of instances used in training. ITW: in-the-wild. EFT~\cite{HanbyulJoo2022eft}, NeuralAnnot (NeA)~\cite{moon2022neuralannot} and UP3D~\cite{lassner2017unite} produce pseudo labels. Unit: mm.}
  \label{tab:single_datasets}
  \centering
  \resizebox{\textwidth}{!}{
  \begin{tabular}{lccccc|cccccccccc|c}
    \toprule
    &  &  &  &  &  &   
    \multicolumn{2}{c}{AGORA~\cite{patel2021agora}} &
    \multicolumn{2}{c}{UBody~\cite{lin2023one}} &
    \multicolumn{2}{c}{EgoBody~\cite{zhang2022egobody}} &
    \multicolumn{2}{c}{3DPW~\cite{von2018recovering}} &
    \multicolumn{2}{c}{EHF~\cite{Pavlakos_2019smplx}}\\
    
    \cmidrule(lr){7-8} \cmidrule(lr){9-10} \cmidrule(lr){11-12} \cmidrule(lr){13-14}  \cmidrule(lr){15-16} 
    
    Dataset & \#Inst. & Scene & Real/Synthetic & SMPL & SMPL-X &
    PVE$\downarrow$ & $\bigstar$ &
    PVE$\downarrow$ & $\bigstar$ &
    PVE$\downarrow$ & $\bigstar$ &
    MPJPE$\downarrow$ & $\bigstar$ &
    PVE$\downarrow$ & $\bigstar$ &
    \textbf{MPE$\downarrow$} 
    \\
    \midrule
BEDLAM~\cite{black2023bedlam} & 951.1K & ITW & Syn & - & Yes & \underline{164.7} & 4 & 132.5 & 8 & \underline{109.1} & 2 & \underline{98.1} & 2 & \textbf{81.1} & 1 & \textbf{117.1} \\
SynHand & 462.8K & ITW & Syn & - & Yes & 182.3 & 6 & 141.2 & 10 & \underline{111.0} & 3 & 108.35 & 9 & \underline{99.1} & 4 & \underline{128.4} \\
SynBody~\cite{yang2023synbody} & 633.5K & ITW & Syn & - & Yes & \underline{166.7} & 5 & 144.6 & 12 & 136.6 & 8 & 106.5 & 7 & 112.9 & 8 & \underline{133.5} \\
InstaVariety~\cite{kanazawa2019learning} & 2184.8K & ITW & Real & NeA & - & 195 & 11 & \underline{125.4} & 4 & 140.1 & 13 & \underline{100.6} & 4 & 110.8 & 7 & \underline{134.3} \\
GTA-Human~\cite{cai2024playing} & 1802.2K & ITW & Syn & - & Yes & \underline{161.9} & 3 & 143.7 & 11 & 139.2 & 12 & 103.4 & 6 & 126 & 15 & \underline{134.8} \\
MSCOCO~\cite{lin2014microsoft} & 149.8K & ITW & Real & EFT & NeA & 191.6 & 10 & \underline{107.2} & 2 & 139 & 11 & 121.2 & 15 & 116.3 & 10 & 135.0 \\
AGORA~\cite{patel2021agora} & 106.7K & ITW & Syn & Yes & Yes & \textbf{124.8} & 1 & 128.4 & 6 & 138.4 & 10 & 131.1 & 17 & 164.6 & 30 & 140.6 \\
EgoBody-MVSet~\cite{zhang2022egobody} & 845.9K & Indoor & Real & Yes & Yes & 190.9 & 9 & 191.4 & 22 & \underline{127.0} & 4 & \underline{99.2} & 3 & \underline{101.8} & 5 & 142.1 \\
IDEA400~\cite{lin2023motionx} & 1362.9K & Indoor & Real & - & Yes & 213.4 & 23 & 193.7 & 23 & \underline{129.4} & 5 & 106.6 & 8 & \underline{91.4} & 2 & 146.9 \\
WHAC-A-Mole~\cite{yin2025whac} & 2500.4K & ITW & Syn & - & Yes & 184.6 & 7 & 150.7 & 14 & 130.8 & 6 & 117.7 & 13 & 152.8 & 26 & 147.3 \\
EgoBody-EgoSet~\cite{zhang2022egobody} & 90.1K & Indoor & Real & Yes & Yes & 207.1 & 19 & \underline{126.8} & 5 & \textbf{103.1} & 1 & 134.4 & 23 & 121.4 & 13 & 147.5 \\
RICH~\cite{huang2022capturing} & 243.4K & ITW & Real & - & Yes & 195.6 & 12 & 168.1 & 19 & 137.9 & 9 & 115.5 & 12 & 127.5 & 16 & 148.9 \\
MPII~\cite{andriluka14cvpr} & 28.9K & ITW & Real & EFT & NeA & 202.1 & 13 & \underline{123.9} & 3 & 155.5 & 19 & 131.9 & 19 & 140.8 & 19 & 150.8 \\
MuCo-3DHP~\cite{Mehta2018SingleShotM3} & 465.3K & ITW & Real & Yes & - & 187.7 & 8 & 185.4 & 21 & 146.4 & 16 & 119.4 & 14 & 134.7 & 18 & 154.7 \\
EMDB~\cite{kaufmann2023emdb} & 109.8K & ITW & Real & Yes & - & 205.9 & 17 & 202.8 & 25 & 136.1 & 7 & \textbf{97.8} & 1 & 148.4 & 23 & 158.2 \\
PROX~\cite{hassan2019resolving} & 88.5K & Indoor & Real & - & Yes & 204.1 & 15 & 180.3 & 20 & 151.8 & 17 & 132.5 & 22 & 122.5 & 14 & 158.2 \\
UBody~\cite{lin2023one} & 683.3K & ITW & Real & - & Yes & 207 & 18 & \textbf{78.7} & 1 & 145.6 & 15 & 149.4 & 29 & 132.1 & 17 & 158.5 \\
SPEC~\cite{kocabas2021spec} & 72.0K & ITW & Syn & Yes & - & \underline{161.5} & 2 & 146.1 & 13 & 154.8 & 18 & 139.7 & 26 & 197.8 & 34 & 160.0 \\
CrowdPose~\cite{li2019crowdpose} & 28.5K & ITW & Real & NeA & - & 207.1 & 20 & 129.8 & 7 & 156.9 & 20 & 156.3 & 31 & 154.5 & 27 & 160.9 \\
MPI-INF-3DHP~\cite{mehta2017monocular} & 939.8K & ITW & Real & NeA & NeA & 221.5 & 26 & 166.7 & 18 & 142.7 & 14 & 131.6 & 18 & 155.5 & 28 & 163.6 \\
HumanSC3D~\cite{fieraru2021learning} & 288.4K & Studio & Real & - & Yes & 215.2 & 24 & 237.8 & 28 & 167.3 & 21 & 113 & 11 & 107.1 & 6 & 168.1 \\
PoseTrack~\cite{andriluka2018posetrack} & 28.5K & ITW & Real & EFT & - & 218.1 & 25 & 161 & 15 & 180.8 & 26 & 150.2 & 30 & 149.9 & 25 & 172.0 \\
BEHAVE~\cite{bhatnagar2022behave} & 44.4K & Indoor & Real & Yes & - & 208.3 & 21 & 205.8 & 26 & 175.8 & 24 & 132 & 20 & 145 & 21 & 173.4 \\
CHI3D~\cite{fieraru2020three} & 252.4K & Studio & Real & - & Yes & 203.3 & 14 & 264.7 & 32 & 175.7 & 23 & 122.6 & 16 & 121 & 12 & 177.5 \\
DAMON~\cite{tripathi2023deco} & 4.4K & ITW & Real & Yes & - & 222.2 & 27 & 161.2 & 16 & 174.8 & 22 & 173 & 32 & 164.4 & 29 & 181.5 \\
Human3.6M~\cite{ionescu2013human3} & 312.2K & Studio & Real & Yes & NeA & 226 & 28 & 276.1 & 33 & 200.6 & 30 & 112.3 & 10 & 120.8 & 11 & 187.2 \\
Hi4D~\cite{yin2023hi4d} & 186.5K & Studio & Real & Yes & - & 213.2 & 22 & 330.5 & 37 & 204.1 & 32 & \underline{102.1} & 5 & \underline{93.5} & 3 & 188.7 \\
DNA-R-HiRes~\cite{renbody} & 998.1K & Studio & Real & - & Yes & 230 & 29 & 278.2 & 34 & 179.2 & 25 & 134.5 & 24 & 149.7 & 24 & 194.3 \\
SLOPER4D~\cite{dai2023sloper4d} & 33.2K & ITW & Real & - & Yes & 205.3 & 16 & 255.1 & 29 & 190 & 28 & 140.3 & 27 & 189.2 & 32 & 196.0 \\
3DPW~\cite{von2018recovering} & 22.7K & ITW & Real & Yes & NeA & 234 & 30 & 259.3 & 30 & 192.6 & 29 & 140.6 & 28 & 142.9 & 20 & 207.2 \\
ARCTIC~\cite{fan2023arctic} & 1539.1K & Studio & Real & - & Yes & 308.5 & 36 & 200.7 & 24 & 186.4 & 27 & 202.5 & 33 & 182.5 & 31 & 216.1 \\
DNA-R~\cite{renbody} & 3992.0K & Studio & Real & - & Yes & 274.7 & 33 & 341.5 & 38 & 214.4 & 35 & 138.4 & 25 & 115.5 & 9 & 216.9 \\
UP3D~\cite{lassner2017unite} & 7.1K & ITW & Real & UP3D & - & 257.5 & 31 & 224.1 & 27 & 216.6 & 36 & 211.5 & 34 & 194.8 & 33 & 220.9 \\
Talkshow~\cite{yi2023generating} & 3326.9K & Indoor & Real & - & Yes & 286.4 & 34 & 133.2 & 9 & 203.6 & 31 & 291.3 & 36 & 201.9 & 36 & 223.3 \\
SignAvatars~\cite{yu2025signavatars} & 4975.1K & Studio & Real & - & Yes & 317.4 & 37 & 164.6 & 17 & 207.2 & 33 & 306.82 & 38 & 201.47 & 35 & 239.5 \\
FIT3D~\cite{fieraru2021aifit} & 1779.3K & Studio & Real & - & Yes & 329.7 & 38 & 404 & 39 & 213.8 & 34 & 132.1 & 21 & 148.1 & 22 & 245.5 \\
MTP~\cite{muller2021self} & 3.2K & ITW & Real & Yes & Yes & 272.7 & 32 & 284.9 & 35 & 273.2 & 37 & 265.2 & 35 & 244.6 & 37 & 268.1 \\
OCHuman~\cite{zhang2019pose2seg} & 2.5K & ITW & Real & EFT & - & 307.1 & 35 & 263.3 & 31 & 279.3 & 38 & 293.4 & 37 & 281.7 & 38 & 285.0 \\
LSPET~\cite{Johnson2011LearningEH} & 2.9K & ITW & Real & EFT & - & 365.7 & 39 & 292.6 & 36 & 340.1 & 40 & 339.8 & 40 & 316.3 & 39 & 330.9 \\
MoYo~\cite{tripathi2023ipman} & 1668.4K & Studio & Real & - & Yes & 481.8 & 40 & 436.4 & 40 & 301.7 & 39 & 335 & 39 & 421.1 & 40 & 395.2 \\
SSP3D~\cite{Sengupta2020SyntheticTF} & 311.0K & ITW & Real & Yes & - & 549.8 & 41 & 522.4 & 41 & 548.1 & 41 & 439 & 41 & 539.5 & 41 & 519.8 \\

    \bottomrule
  \end{tabular}}
\end{table*}

\section{Related Work}

\noindent \textbf{Expressive Human Pose and Shape Estimation (EHPS).}
Due to the erupting expressive avatar applications and the advent of whole-body parametric model (e.g., SMPL-X~\cite{Pavlakos_2019smplx}), capturing the expressive human pose and shape (EHPS), which estimates body, hand, and face together from images, has attracted increasing attention~\cite{Pavlakos_2019smplx,Xiang_2019_wholebody3d,PavlakosGeorgios2020expose,Rong_2021frank,Zhou_2021_full,Feng_2021_pixie, sun2022learning,HongwenZhang2022PyMAFXTW,lin2023one}. 
Optimization-based methods (e.g., SMPLify-X~\cite{Pavlakos_2019smplx}) detect 2D features corresponding to the whole body and fit the SMPL-X model. However, they suffer from slow speed and poor performance. Hence, learning-based models are proposed.
One of the key challenges of EHPS is the low resolution of hands and face compared with the body-only estimation, making the articulated hand pose estimation and high-quality expression capture hard. Accordingly, mainstream whole-body models first detect and crop the hands and face image patches, then resize them to higher resolutions and feed them into specific hand and face networks to estimate the corresponding parameters~\cite{PavlakosGeorgios2020expose,Rong_2021frank,Zhou_2021_full,Feng_2021_pixie,sun2022learning,GyeongsikMoon2020hand4whole,HongwenZhang2022PyMAFXTW,li2023hybrik}.
Due to the highly complex multi-stage pipelines, they inevitably cause inconsistent and unnatural articulation of the mesh and implausible 3D wrist rotations, especially in occluded, truncated, and blurry scenes. 
\textcolor{blue}{OSX~\cite{lin2023one} proposes the first one-stage framework based on ViT-based backbone~\cite{dosovitskiy2020image}, followed by recent methods such as AiOS~\cite{sun2024aios} and Multi-HMR~\cite{multi-hmr2024}. This line of work does not require a separate detection stage; it integrates localization and parameter estimation into a single process. As a result, this architecture alleviates the issues in previous multi-stage pipelines, and enhances EHPS performance in scenes prone to imperfect detection.}
Despite the concise design, they only use confined training datasets for a fair comparison and do not explore the combination of more data to achieve generalizable and precise EHPS.

\noindent \textbf{Large-scale Training for Human-centric Vision.}
Recent efforts use multiple datasets to pretrain a general model for a wide range of downstream human-centric tasks. 
For example, HumanBench~\cite{tang2023humanbench} leverages 37 datasets, whereas UniHCP~\cite{ci2023unihcp} utilizes 33 datasets for tasks such as ReID, pedestrian detection, and 2D pose estimation. However, these works have only evaluated the efficacy of 2D tasks. 
S{\'a}r{\'a}ndi \etal~\cite{sarandi2023learning} take advantage of 28 datasets in training a strong model for 3D keypoint detection, which recovers only the skeleton of subjects without estimating body shapes and meshes.
Pang \etal~\cite{pang2022benchmarking} analyze 31 datasets for human pose and shape estimation (\ie, SMPL estimation). Nonetheless, hands and face estimation is not included, and only fewer than ten datasets are used concurrently in the most diverse training.
\textcolor{blue}{
HaMeR~\cite{pavlakos2024reconstructing} consolidates 10 datasets, up to 2.7M training examples for training a transformer model for 3D hand-only reconstruction.
NLF~\cite{sarandi24nlf} is capable of mixing data sources with different annotation formats, including parametric annotations, 3D keypoint annotations, 2D keypoint annotations and DensePose annotations. Despite this, it suffers from inaccurate hand pose due to the limited sample points on hands.
}
This paper targets the more difficult and essential EHPS task to scale training data and model size that accurately recovers the expressive pose and shape of the human body, hands, and face.

\section{Benchmarking EHPS Datasets}

\subsection{Preliminaries}
\label{sec:preliminaries}

\textbf{SMPL-X.}
We study expressive human pose and shape estimation via 3D parametric human model SMPL-X \cite{pavlakos2019expressive}, which models the human body, hands, and face geometries with parameters. Specifically, our goal is to estimate pose parameters $\theta\in\mathbb{R}^{55\times3}$ that include body, hands, eyes, and jaw poses; joint body, hands and face shape $\beta\in\mathbb{R}^{10}$, and facial expression $\psi\in\mathbb{R}^{10}$. The joint regressor $\mathcal{J}$ is used to obtain 3D keypoints from parameters via $R_{\theta}(\mathcal{J}(\beta))$ where $R_{\theta}$ is a transformation function along the kinematic tree.

\textbf{Evaluation Metrics.}
We use standard metrics for EHPS. PVE (per-vertex error) and MPJPE (mean per-joint position error) measure the mean L2 distances between vertices and regressed joints, respectively. The ``PA" prefix indicates Procrutes Alignment is conducted before error computation. AGORA Leaderboard~\cite{patel2021agora} introduces NMVE (normalized mean vertex error) and NMJE (normalized mean joint error) that take detection performance F1 score into consideration. Moreover, we propose MPE (mean primary error) that takes the mean of multiple primary metrics (MPJPE for 3DPW~\cite{von2018recovering} test, and PVE for AGORA-val, UBody, EgoBody, and EHF) to gauge generalizability. \textcolor{blue}{Similar to the whole-body counterpart, we propose hand-MPE (the mean of hand PVE) for wrist-aligned hand pose and hand-PA-MPE (the mean of hand PA-PVE) for local pose of hands from multiple datasets (AGORA-val, UBody, EgoBody, EHF, ARCTIC and SynHand) to evaluate the performance of hand pose estimation across key benchmarks. As 3DPW~\cite{von2018recovering} provides the annotation in SMPL format for body-only evaluation, we replace it with ARCTIC~\cite{fan2023arctic} for hand-relevant metrics.} All errors are reported in millimeters (mm). 

\subsection{Overview of Data Sources}
In this work, we gather \textcolor{blue}{\numdatasets datasets} from three major types. 
1) motion capture datasets that leverage optical~\cite{ionescu2013human3, fieraru2020three, fan2023arctic, fieraru2021learning, fieraru2021aifit, mehta2017monocular, yin2023hi4d} or vision-based~\cite{zhang2022egobody, hassan2019resolving, renbody, tripathi2023ipman} multi-view motion capture systems, are typically collected in a studio environment. However, it is possible to include an outdoor setup, or utilize additional sensors such as IMUs \textcolor{blue}{or LiDARs}~\cite{von2018recovering, kaufmann2023emdb, dai2023sloper4d}. These datasets generally provide high-quality 3D annotations but are less flexible due to physical constraints, especially those built with optical motion capture systems. 
2) pseudo-annotated datasets~\cite{lin2014microsoft, andriluka2018posetrack, li2019crowdpose, lin2023one,  Mehta2018SingleShotM3, kanazawa2019learning, zhang2019pose2seg, andriluka14cvpr, Johnson2011LearningEH, muller2021self, yi2023generating, Sengupta2020SyntheticTF, lin2023motionx, yu2025signavatars, tripathi2023deco, Joo2020ExemplarFF, moon2022neuralannot} that re-annotate existing image datasets with parametric human annotations. These datasets take advantage of the diversity of 2D datasets, and the pseudo-3D annotations, albeit typically not as high-quality, have been proven effective~\cite{pang2022benchmarking}. 
3) synthetic datasets~\cite{black2023bedlam, cai2024playing, kocabas2021spec, patel2021agora, yang2023synbody, yin2025whac} that are produced with renderings engines (\eg, Unreal Engine). These datasets produce the most accurate 3D annotations and can easily scale up with high diversity. However, the synthetic-real gap is not fully addressed. Key attributes of the datasets are included in \Tab\ref{tab:single_datasets}.

\subsection{Whole-body Evaluation Benchmarks}
To evaluate the EHPS capability across diverse scenarios, we select multiple key datasets to form a comprehensive benchmark. They should possess the desirable traits such as 1) having accurate SMPL or SMPL-X annotations, 2) being representative of certain aspects of real-life scenarios, 3) being widely used, but this requirement is relaxed for the new datasets which are released within two years, and 4) has a clearly defined test set. To this end, five datasets (AGORA~\cite{patel2021agora}, UBody~\cite{lin2023one}, EgoBody~\cite{zhang2022egobody}, 3DPW~\cite{von2018recovering}, and EHF~\cite{Pavlakos_2019smplx}) representing different aspects are selected as the evaluation datasets \textcolor{blue}{for whole-body pose evaluation. We briefly introduce the evaluation datasets and the rest in the \Supp.}
\textbf{AGORA} is the most widely-used benchmark for SMPL-X evaluation. It is a synthetic dataset featuring diverse subject appearances, poses, and environments with high-quality annotation. We evaluate on both validation and test set (leaderboard) as the latter has a monthly limit of submissions. 
\textbf{UBody} is the latest large-scale dataset with pseudo-SMPL-X annotations that covers fifteen real-life scenarios, such as talk shows, video conferences, and VLOGs, which mainly present the upper body in images. We follow the intra-scene protocol in training and testing, where all scenarios are seen.
\textbf{EgoBody} captures human motions in social interactions in 3D scenes with pseudo-SMPL-X annotations. It comprises a first-person egocentric set (EgoSet) and a third-person multi-camera set (MVSet). We test on the EgoSet with heavy truncation and invisibility.
%
\textbf{3DPW} is the most popular in-the-wild dataset with SMPL annotations. Since SMPL-X annotation is not available, we transform the estimation in SMPL-X to SMPL topology following~\cite{Pavlakos_2019smplx} for evaluation. 
\textbf{EHF} is a classic dataset with 100 curated frames of one subject in an indoor studio setup, with diverse body poses and especially hand poses annotated in SMPL-X vertices. It has a test set but no training or validation sets. Hence, it is only used to evaluate cross-dataset performance.

\begin{figure*}[t!]
  \centering
  \includegraphics[width=\linewidth]{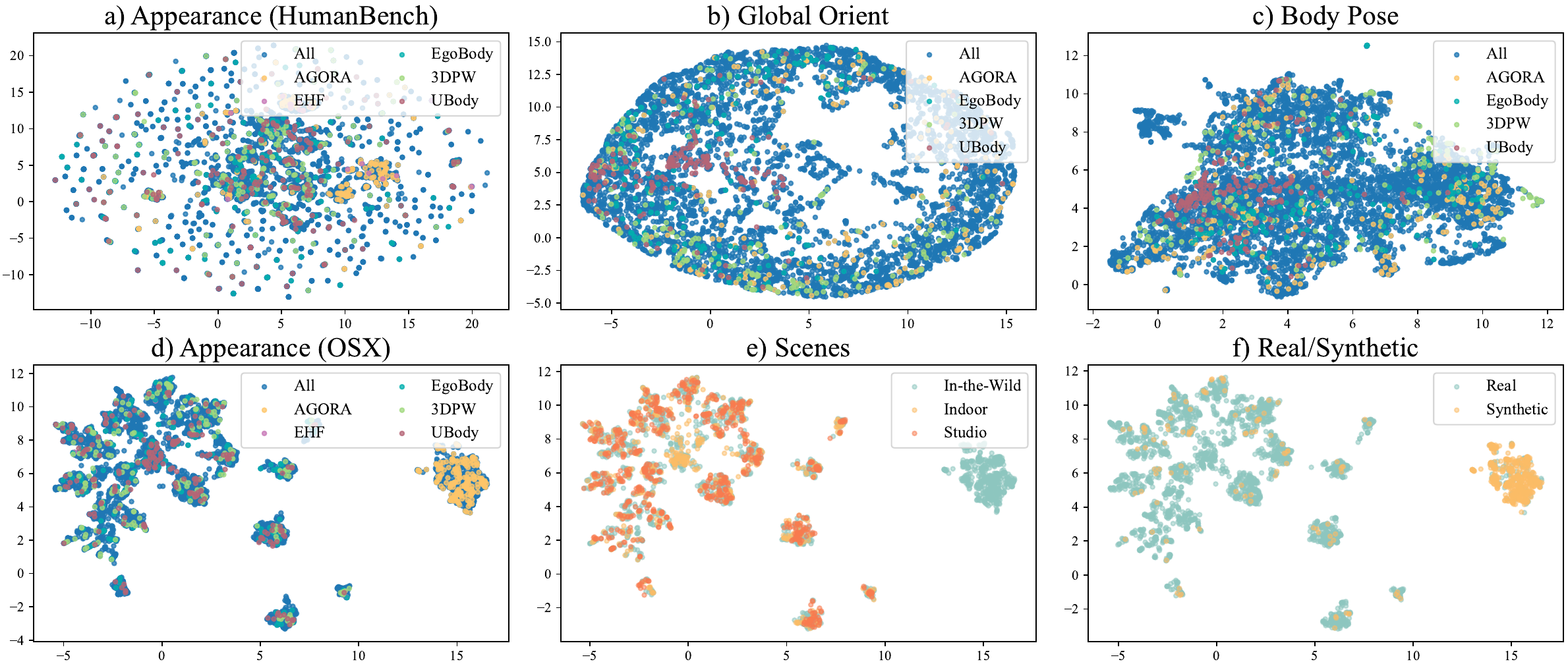}
  \caption{\textbf{Dataset attribute distributions.} a) and d) are image feature extracted by HumanBench~\cite{tang2023humanbench} and OSX~\cite{lin2023one} pretrained ViT-L backbone. b) Global orientation (represented by rotation matrix) distribution. c) Body pose (represented by 3D skeleton joints) distribution. Both e) scenes and f) Real/eot are drawn on the same distribution as d). All: all datasets. UMAP~\cite{mcinnes2018umap} dimension reduction is used in all visualization with the x and y-axis as the dimensions of the embedded space.  }
  \label{fig:benchmark_distributions}
\end{figure*}

Besides being popular or the latest evaluation sets for EHPS, we further analyze if these five datasets collectively provide wide coverage of existing datasets. In \Fig\ref{fig:benchmark_distributions}, we randomly downsample all datasets to equal length (1K examples) and employ UMAP~\cite{mcinnes2018umap} to visualize several key aspects. We use pretrained ViT-L from HumanBench~\cite{tang2023humanbench} and OSX~\cite{lin2023one} to process patch tokens flattened as feature vectors from images cropped by bounding boxes. HumanBench is trained for various human-centric tasks (\eg, Re-ID, part segmentation, and 2D pose estimation), whereas OSX is an expert model on EHPS. As for global orientation, it is closely associated with camera pose as we convert all data into the camera coordinate frame; we plot its distribution by using flattened rotation matrix representations. Moreover, we follow~\cite{rong2019delving, cai2024playing, pang2022benchmarking} to represent poses as 3D keypoints regressed from the parametric model. Specifically, we flatten 21 SMPL-X body keypoints, and 15 hand keypoints from each hand, regressed with mean parameters except for the body pose and hand poses. More evaluation is included in the \Supp. It is shown that 1) the five benchmark datasets have varied distribution, which is expected due to their different designated purposes, and 2) collectively, the five datasets provide a wide, near-complete coverage of the entire dataset pool.

\subsection{Benchmarking on Individual Datasets}
\label{sec:single_datasets}

\begin{figure}[t!]
  \centering
  \vspace{0mm}
  \includegraphics[width=\linewidth]{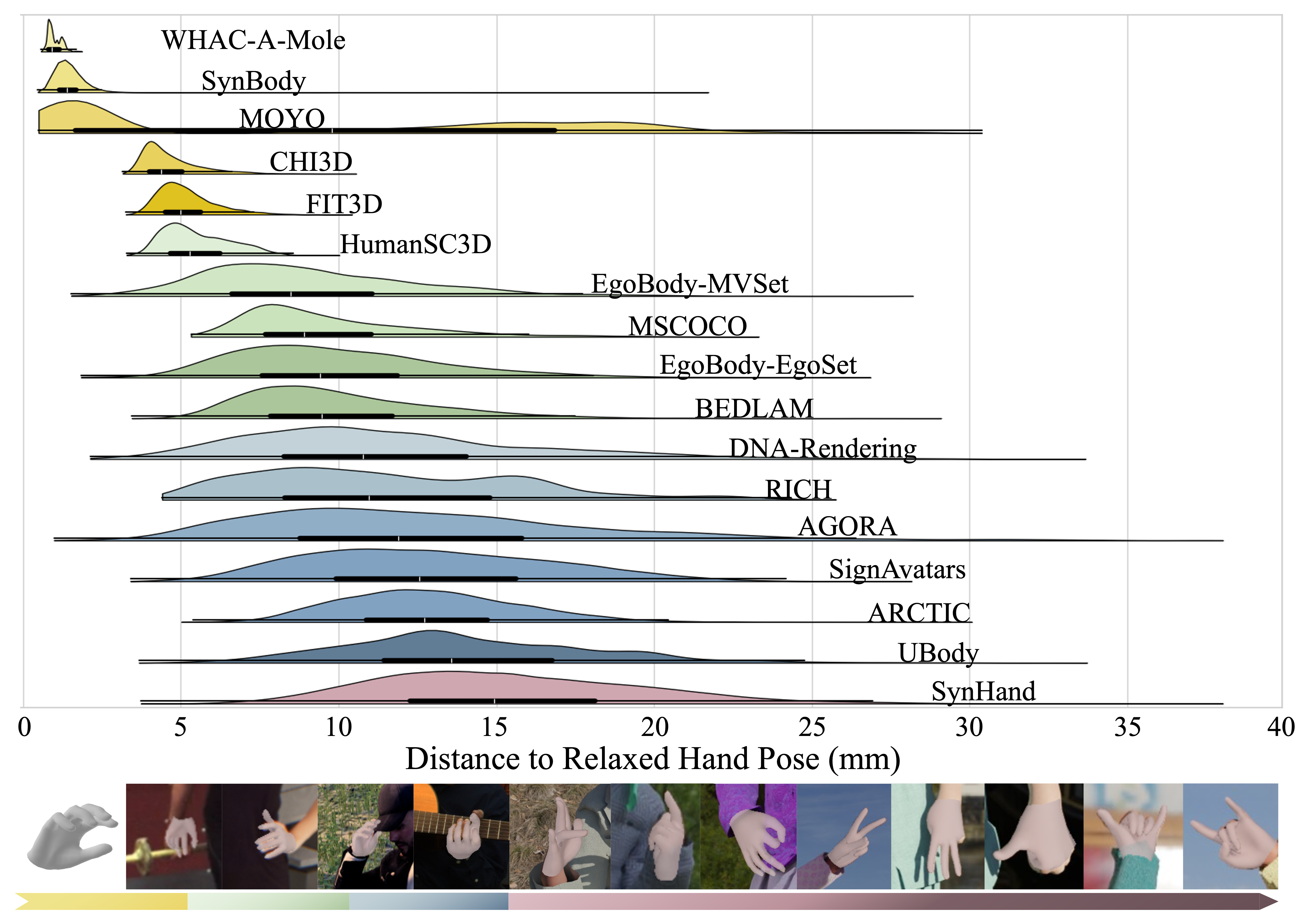}
  \caption{\textbf{Analysis of hand poses in whole-body datasets.} The distribution of the distance to the relaxed hand pose of each dataset (top) is shown along with the illustration of pose complexity at various distances (bottom). The hand pose with a lower distance is more similar to the relaxed pose.}
  \label{fig:hand_pose_analysis}
\end{figure}

\begin{figure}[t!]
  \centering
  \vspace{0mm}
  \includegraphics[width=\linewidth]{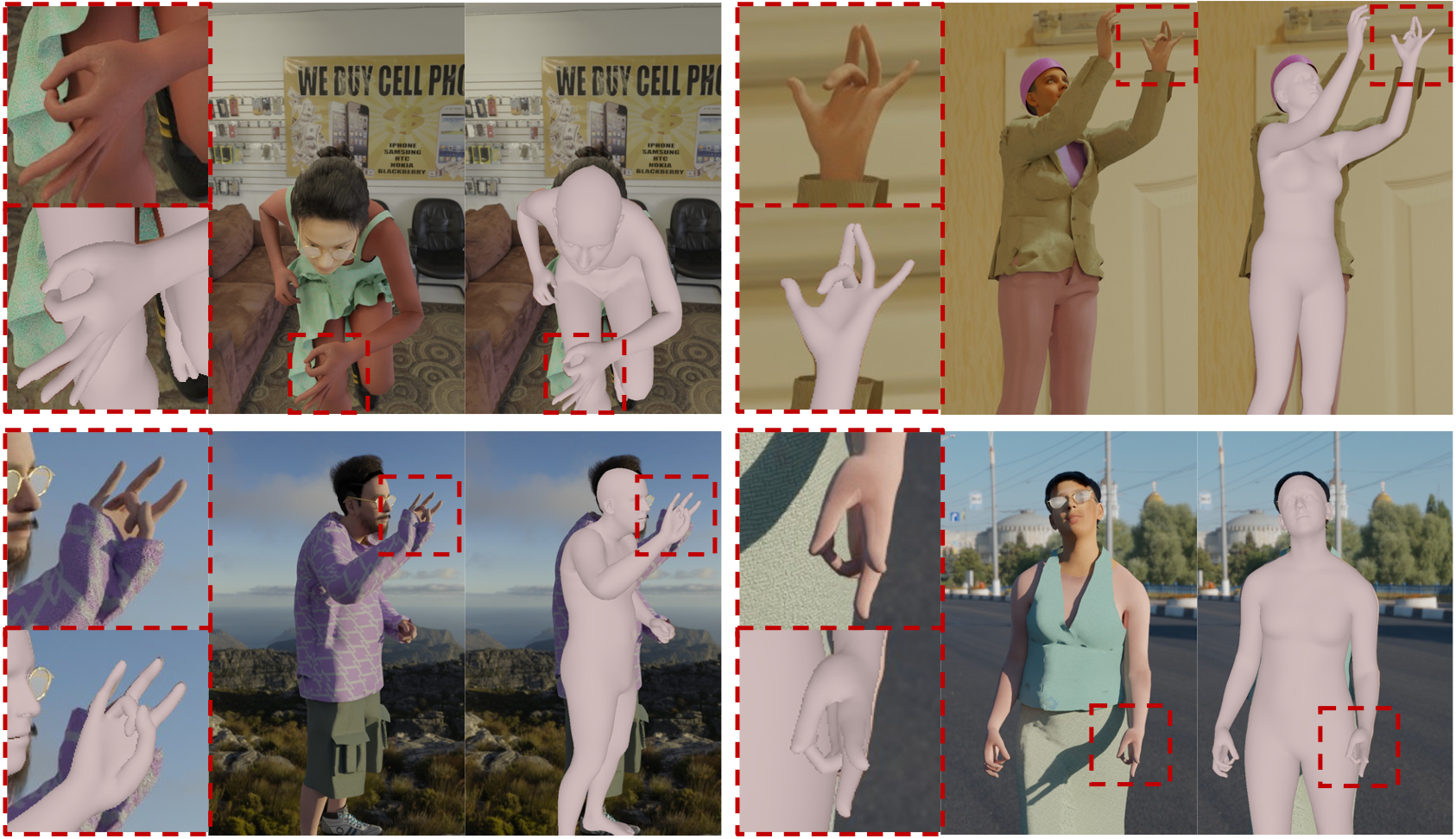}
  \caption{\textbf{Visualization of SynHand dataset} with complex hand poses and accurate annotations.}
  \label{fig:synhand}
\end{figure}


    
    
    
    

\begin{table}[t]
\centering
\caption{\textbf{Benchmarking EHPS methods on SynHand.} Unit: mm. *~denotes that the method uses integrated detection results, F1=98.0.}
\label{tab:synhand}
\vspace{0mm}
\resizebox{\linewidth}{!}{
    \begin{tabular}{lcccccc}
    \toprule
    & 
    \multicolumn{2}{c}{All} &
    \multicolumn{2}{c}{Hands} &
    \multicolumn{2}{c}{Face} \\
    
    \cmidrule(lr){2-3} \cmidrule(lr){4-5} \cmidrule(lr){6-7} 
    
    Method &
    PA-PVE$\downarrow$ & PVE$\downarrow$ &
    PA-PVE$\downarrow$ & PVE$\downarrow$ &
    PA-PVE$\downarrow$ & PVE$\downarrow$ \\
    
    \midrule
    Hand4Whole~\cite{GyeongsikMoon2020hand4whole} & 64.9 & 116.3 & 11.7 & 50.8 & 3.8 & 44.2 \\
    PyMAF-X~\cite{pymafx2023} & 63.5 & 112.6 & 11.8 & 49.2 & 3.6 & 44.9 \\
    OSX~\cite{lin2023one} & 60.6 & 107.0 & 16.7 & 55.6 & 4.0 & 41.6 \\
    HybrIK-X~\cite{li2023hybrik} & 53.6 & 96.5 & 16.7 & 50.4 & 6.0 & 43.6 \\
    AiOS~\cite{sun2024aios} & 52.2 & 94.1 & 11.8 & 43.9 & 3.3 & 33.5 \\
    Multi-HMR*~\cite{multi-hmr2024} & 38.2 & 70.3 & 11.6 & 35.2 & 3.3 & 29.7 \\
    \name-H32 & 37.7 & 64.7 & 14.0 & 42.7 & 3.1 & 28.5 \\
    \name-H40 & 29.2 & 47.6 & 10.3 & 32.0 & 2.9 & 22.7 \\
    \namepami-H32 & 37.3 & 63.9 & 12.4 & 38.6 & 3.1 & 28.1 \\
    \namepami-H40 & \textbf{21.7} & \textbf{34.4} & \textbf{9.8} & \textbf{23.7} & \textbf{2.1} & \textbf{15.4} \\
    
    \bottomrule
    \end{tabular}
}
\end{table}

In this section, we aim to benchmark datasets and find those that do well in various scenarios. To gauge the performance of each dataset, we train a model (\name with ViT-S as the backbone, trained for 5 epochs. See \Sec\ref{sec:architecture} for model architecture) with the training set of that dataset and evaluate the model on the \textit{testing} sets of five evaluation datasets: AGORA, UBody, EgoBody, 3DPW, and EHF. 

In \Tab\ref{tab:single_datasets}, we report the primary metrics (\Sec\ref{sec:preliminaries} and \textcolor{blue}{ranking of each of the \numdatasets datasets.} We also compute the mean primary error (MPE) to facilitate easy comparison between individual datasets. Note that for AGORA, UBody, EgoBody, and 3DPW, their performances on their own test set are excluded from computing MPE. This is because in-domain evaluation results are typically much better than cross-domain ones, leading to significant error drops. In addition, note that there are datasets designed for specific purposes (\eg, Talkshow~\cite{yi2023generating} for gesture generation, DNA-Rendering~\cite{renbody}, for human NeRF reconstruction, \textcolor{blue}{MoYo~\cite{tripathi2023ipman} for physical plausibility in complex human motions
, and SignAvatars~\cite{yu2025signavatars} for sign language motion studies}), being ranked lower on our benchmark, which focuses on EHPS (a perception task) does not reduce their unique values and contributions to the computer vision community. 

From the benchmark, we observe models trained on a single dataset tend to perform well on the same domain but often cannot do well on other domains. For example, the model trained on AGORA is ranked $1^{st}$ on AGORA, but $6^{th}$ on UBody, \textcolor{blue}{$10^{th}$ on EgoBody, $17^{th}$ on 3DPW, and $30^{th}$ on EHF.} This observation indicates that 1) the test scenarios are diverse, showcasing the challenging landscape of EHPS, and 2) data scaling is essential for training a robust and transferable model for EHPS due to significant gaps between different domains.

\subsection{\textcolor{blue}{Hand Evaluation Benchmarks}}
\label{sec:hand_eval}
\textcolor{blue}{
To provide a comprehensive evaluation of the hand performance in the whole-body setting, two additional datasets are selected for hand pose evaluation. First, \textbf{ARCTIC} is a lab-based hand-object interaction dataset with accurate SMPL-X annotations for both body pose and hand pose under various scenarios. There are 10 subjects interacting with 11 objects, with 210K frames of videos captured in total. Second, \textbf{SynHand} is the synthetic dataset introduced in this work (elaborated below), which features close-up human shots, diverse body shapes with garments, and a wide range of expressive hand poses. 
}

\textcolor{blue}{
We introduce \textbf{SynHand}, a synthetic dataset specifically designed for EHPS estimation, with a particular emphasis on overcoming the challenges of hand pose estimation with the following motivations.
1) Hand poses in a relaxed state are frequently observed in many whole-body datasets that do not prioritize hands. These datasets often lack specific design or selection criteria for hand poses during the data collection process.
2) The images in mainstream whole-body datasets, such as those captured or rendered with large fields of view (FOV)~\cite{patel2021agora, zhang2022egobody}, often include multiple humans within a single frame. However, this broad FOV comes at the cost of resolution, resulting in low-resolution or barely visible hands in the images.
3) Unlike studio-captured datasets where markers can facilitate in annotations~\cite{fan2023arctic}, obtaining accurate hand pose annotations remains challenging for in-the-wild datasets.}

\textcolor{blue}{
SynHand leverages the layered human model SMPL-XL~\cite{yang2023synbody} and the advanced synthetic rendering toolbox XRFeitoria~\cite{xrfeitoria} to generate photorealistic images of virtual humans with diverse body shapes, clothing, and accessories. To capture hands with clarity and visibility, camera placements are strategically optimized for close-up shots, ensuring detailed visual representations.
%
Moreover, the hand poses are extracted from InterHand~\cite{Moon_2020_ECCV_InterHand2.6M}, AGORA~\cite{patel2021agora} and GRAB~\cite{taheri2020grab} following BEDLAM~\cite{black2023bedlam}, while the body poses are sourced from AMASS~\cite{mahmood2019amass}. 
For body poses, we carefully select 100 motions from AMASS, in which hand movements can be effectively addressed, such as motions where the wrists are brought forward with fewer occlusions. These motions were chosen to ensure that, with appropriate camera placement, the hand poses can be emphasized. 80\% of the body poses are sampled from this subset while 20\% are randomly sampled from the entire AMASS dataset.
For hand poses, left and right hands are randomly sampled from the three datasets independently. We adopt the official train and test split provided for InterHand datasets.
The combined whole-body poses are subsequently used to animate the SMPL-XL model, forming the motion sequences for our dataset.
The remaining aspects of data generation, such as scene setup and rendering configurations, follow the approach used in the PDHuman dataset~\cite{wang2023zolly}.
\Fig\ref{fig:synhand} shows examples from the SynHand dataset with complicated hand poses and accurate annotations. We also provide the benchmark on SynHand as shown in \Tab\ref{tab:synhand}.
}

\textcolor{blue}{
\Fig\ref{fig:hand_pose_analysis} explains the reason why ARCTIC and SynHand are selected as key benchmarks for hand pose estimation in EHPS. \Fig\ref{fig:hand_pose_analysis} presents an analysis conducted across 17 representative whole-body datasets with hand pose annotations. It illustrates the distribution of hand poses as distance to the relaxed hand (when hand pose parameters are all zero). Note that the relaxed hands have naturally curled fingers. Specifically, the average per-vertex position distance from the relaxed hand pose is calculated for each hand pose within each dataset. Higher distance values indicate greater deviations from the relaxed hand pose, suggesting more complicated and potentially more meaningful hand poses. The 17 datasets are thus sorted according to their median distance to the relaxed hand pose. Notably, SynHand exhibits the largest median value away from the relaxed hand, and has a wide range of hand pose distribution, spanning from less than 5 to greater than 35. ARCTIC is also ranked highly on this list.
}

\section{Foundation Models}
\begin{figure*}
  \centering
  \includegraphics[width=\linewidth]{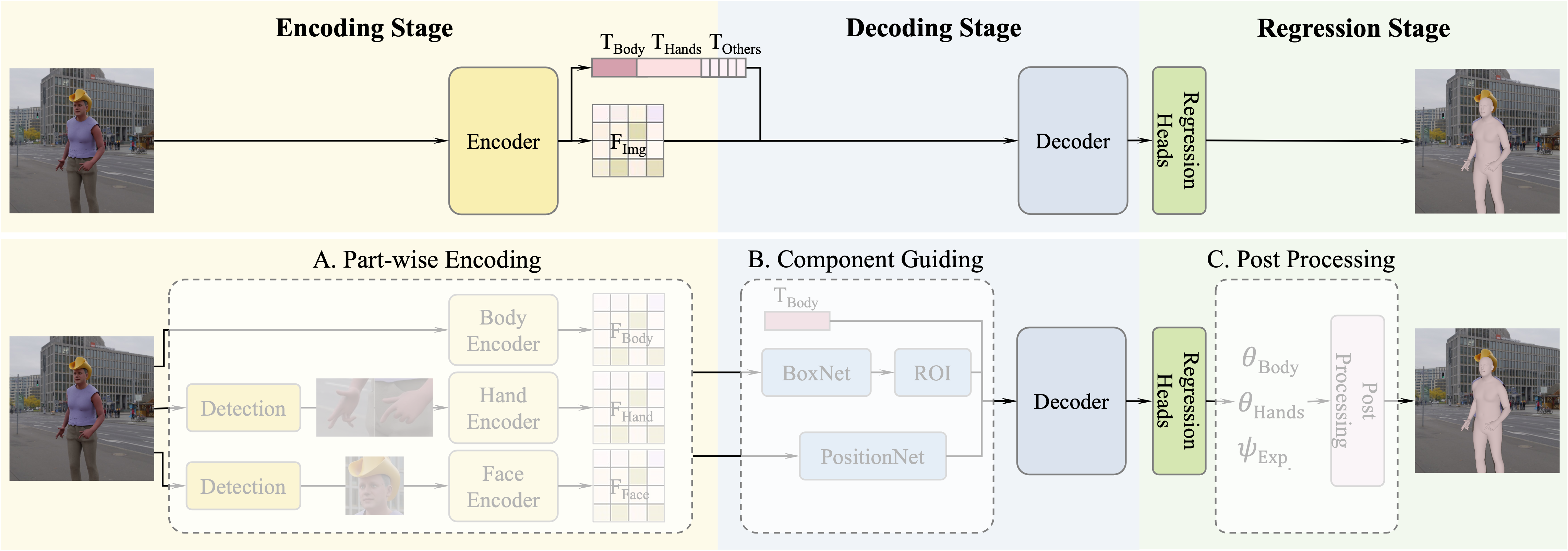}
  \caption{\textbf{Architecture of \namepami.} Compared with other frameworks with algorithmic modules in various stages (bottom), \namepami (top) has a minimalistic framework design in all three stages. Noted that \name consists B. Component Guiding module in the decoder stage.}
  \label{fig:architecture}
\end{figure*}

\vspace{-0.1cm}

\subsection{\textcolor{blue}{Architecture Design}}
\label{sec:architecture}

We highlight that training foundation models are very expensive. Hence, it is infeasible to conduct extensive architectural searches. In this light, \textcolor{blue}{we introduce our minimalistic foundation model frameworks, \name and \namepami that only retain the essential parts for two reasons.}
First, it must be scalable and efficient as we train with a large amount of data. Second, we aim to create a framework that is decoupled from specific algorithm designs, providing a clean foundation for future research.

\textcolor{blue}{As illustrated in \Fig\ref{fig:architecture}, the mainstream frameworks can be summarized into three stages: encoding, decoding, and regression stage. Innovative model designs are typically introduced at one or more of these stages. Our minimalistic framework simplifies each stage, ensuring that \name and \namepami do not require hand or face decoders following part-wise encoding \cite{Rong_2021frank, GyeongsikMoon2020hand4whole, pymafx2023} in the encoding stage. It avoids cross-part feature interaction modules \cite{PavlakosGeorgios2020expose, Feng_2021_pixie}, projections of coarse SMPL-X estimations \cite{HongwenZhang2022PyMAFXTW} in the decoding stage, and post-processings \cite{li2023hybrik} in the regression stage. }

\textcolor{blue}{In the decoding stage, \name retains only the component guiding module, which is intuitive for by-part estimation. \name consists of three parts as shown in \Fig\ref{fig:architecture}}: a backbone that extracts image features, which we employ Vision Transformer~\cite{dosovitskiy2020image} for its scalability \textcolor{blue}{in the encoding stage; the decoding stage consists of a component guiding module} that predicts bounding boxes and crops regions of interest from the feature map for hands and face; \textcolor{blue}{Multiple regression heads in the regression stage that estimate parameters for each part (e.g. body orientation, body pose, hand pose and expressions). The component guiding module in \name functions analogously to a hand and face bounding box detector in feature space. Hand joints regressed from cropped features are used to guide information extraction. However, this leads to significant information loss during cropping from the original encoded features and results in error accumulation during the decoding stage. }

 \textcolor{blue}{To further simplify the design, \namepami eliminates the component guiding module and replaces it with a transformer decoder that has no explicit by-part modeling. \namepami, as its name suggests, is a minimalistic framework that further simplifies \name. By replacing the component guiding module with a transformer decoder, we observe a significant performance boost in hand pose estimation.}

\subsection{\textcolor{blue}{Implementation Details}}
As shown in \Fig\ref{fig:architecture}, \name and \textcolor{blue}{\namepami} utilizes a minimalistic design. Before entering the backbone, the image is cropped by a whole body bounding box and resized to $I$ with (height, width) as (512, 384). The image is then tokenized into 32$\times$24 patches with patch size 16, and undergoes patch embedding, and positional encoding is added to obtain image tokens $T_{img}$. $T_{task}$ is additional learnable tokens (task tokens) that are concatenated with $T_{img}$. \textcolor{blue}{For \name, the $T_{\text{task}}$ is specifically designed for body pose and has a size of $(25, D_{\text{feat}})$, where $D_{feat}$ is the depth of the feature. \namepami extends the total size of task tokens to (80, $D_{feat}$) with additional $T_{hands}$, $T_{face}$, $T_{orients}$ and $T_{trans}$ for hand poses, face expressions, body and hands orientations and translations. Note that the hand pose token $T_{hands}$ has the dimension of (30, $D_{feat}$) in total, as the hand pose is considered as complex as the body pose.} The tokens are processed with \textit{backbone}. Thanks to the scalability of ViT~\cite{dosovitskiy2020image}, we are able to experiment with various model sizes. 
\textcolor{blue}{In the decoding stage, \textit{neck} of \name processes the image feature $F_{img}$ from the encoder and predicts face and hand bounding boxes.} The predicted bounding boxes are used in the ROI (regions of interest) module to crop features from \textcolor{blue}{$F_{img}$}, which is re-organized and undergoes transposed convolution (deconv). \textcolor{blue}{The component guiding module consists of a positional network to predict 3D keypoints of body and hands. $F_{img}$ with the keypoints guidance will be fed into regression heads along with respective task tokens. Whereas \namepami concatenates $F_{img}$ with $T_{task}$ and feeds it into a 6-layered decoder instead of the component guiding module. The decoded $T_{task}$ is then fed into hand and face heads to predict parameters.}
\textcolor{blue}{In the regression stage of \namepami, we include an additional hand orientation head alongside the heads for body pose, hand pose, and expression. The predicted hand orientation is to formulate a parameter-wise consistency loss with the wrist pose of the body, enhancing accuracy in the distal joints of the upper body and hand positioning.}

\section{Scaling Up EHPS}
\begin{table*}
  \caption{\textbf{Evaluate foundation models on whole-body benchmarks.} We study the scaling law of the amount of data and the model sizes. The metrics are MPJPE for 3DPW, and PVE for other evaluation benchmarks. MPE: mean primary error. The lower, the better for all the metrics. The best values are bolded and the second best are underlined.
  AGORA uses the validation set, and EgoBody uses the EgoSet. 40+ denotes the model trained with 10 million instances from 40 datasets. Unit: mm.
  }
  \label{tab:foundation_model}
  \centering
  \vspace{-2mm}
  \resizebox{\textwidth}{!}{
  \begin{tabular}{cccc|ccccc|c}
    \toprule
    
    \#Datasets & \#Inst. & Model & \#Param. &
    AGORA~\cite{patel2021agora} &
    EgoBody~\cite{zhang2022egobody} &
    UBody~\cite{lin2023one} & 
    3DPW~\cite{von2018recovering} &
    EHF~\cite{Pavlakos_2019smplx} &
    MPE$\downarrow$ \\
        
    \midrule
    5 & 0.75M & \name-S5 & 32M &
    119.0 & 114.2 & 110.1 & 110.2 & 100.5 & 110.8 \\
    
    10 & 1.5M & \name-S10 & 32M &
    116.0 & 88.6 & 107.7 & 97.4 & 89.9 & 99.9 \\

    20 & 3.0M & \name-S20 & 32M &
    109.2 & 84.3 & 70.7 & 87.5 & 86.6 & 87.7 \\

    32 & 4.5M & \name-S32 & 32M &
    105.2 & 82.5 & 68.1 & 83.2 & 74.1 & 82.6 \\

    40 & 7.5M & \name-S40 & 32M &
    97.9 & 78.8 & 71.4 & 80.2 & 69.1 & 79.5 \\

    40 & 10.0M & \name-S40+ & 32M &
    94.0 & 78.0 & 66.9 & 80.3 & 67.4 & 77.3 \\
    \midrule

    5 &  0.75M & \name-B5 & 103M &
    102.7 & 108.1 & 105.8 & 104.8 & 96.1 & 103.5 \\
    
    10 &  1.5M & \name-B10 & 103M &
    97.8 & 76.4 & 107.3 & 89.9 & 74.7 & 89.2 \\

    20 & 3.0M & \name-B20 & 103M &
    95.6 & 75.5 & 65.3 & 83.5 & 73.0 & 78.6 \\

    32 & 4.5M & \name-B32 & 103M &
    88.0 & 72.7 & 63.3 & 80.3 & 67.3 & 74.3 \\

    40 & 7.5M & \name-B40 & 103M &
    86.2 & 70.1 & 64.4 & 77.2 & 68.3 & 73.2 \\

    40 & 10.0M & \name-B40+ & 103M &
    81.9 & 68.2 & 60.4 & 74.9 & 64.6 & 70.0\\
    \midrule

    5 & 0.75M & \name-L5 & 327M &
    88.3 & 98.7 & 110.8 & 97.8 & 89.5 & 97.0 \\
    
    10 & 1.5M & \name-L10 & 327M &
    82.6 & 69.7 & 104.0 & 82.5 & 64.0 & 80.6 \\

    20 & 3.0M & \name-L20 & 327M &
    80.7 & 66.6 & 61.5 & 78.3 & 65.4 & 70.5 \\

    32 & 4.5M & \name-L32 & 327M &
    74.2 & 62.2 & 57.3 & 75.2 & 62.4 & 66.2 \\

    40 & 7.5M & \name-L40 & 327M &
    69.0 & 60.7 & 58.6 & 71.5 & 58.6 & 63.7 \\

    40 & 10.0M & \name-L40+ & 327M &
    69.2 & 59.1 & 54.7 & 71.1 & 64.6 & 63.7\\
    \midrule

    5 & 0.75M & \name-H5 & 662M &
    89.0 & 87.4 & 102.1 & 88.3 & 68.3 & 87.0 \\

    10 & 1.5M & \name-H10 & 662M &
    81.4 & 65.7 & 100.7 & 78.7 & 56.6 & 76.6 \\

    20 & 3.0M & \name-H20 & 662M &
    77.5 & 63.5 & 59.9 & 74.4 & 59.4 & 67.0 \\

    32 & 4.5M & \name-H32 & 662M &
    69.5 & \underline{59.5} & 54.5 & 75.0 & 56.9 & 63.1 \\

    40 & 7.5M & \name-H40 & 662M &
    68.2 & 59.9 & 56.9 & \underline{70.9} & 53.8 & 61.9 \\

    40 & 10.0M & \name-H40+ & 662M &
    65.8 & \textbf{57.0} & 53.7 & \textbf{70.5} & 52.3 & \underline{59.9} \\
    \midrule

    5 & 0.75M & \namepami-H5 & 687M &
    96.3 & 92.3 & 95.2 & 82.0 & 71.6 & 87.5 \\

    10 & 1.5M & \namepami-H10 & 687M &
    71.4 & 68.7 & 100.8 & 78.2 & 60.3 & 75.9 \\

    20 & 3.0M & \namepami-H20 & 687M &
    76.4 & 63.8 & 59.8 & \textbf{70.5} & 65.4 & 67.2 \\

    32 & 4.5M & \namepami-H32 & 687M &
    70.6 & 59.8 & 56.5 & 75.8 & 53.0 & 63.2 \\

    40 & 7.5M & \namepami-H40 & 687M &
    \textbf{56.1} & 61.5 & \textbf{51.1} & 76.0 & \textbf{49.7} & \textbf{58.9} \\

     40 & 10.0M & \namepami-H40+ & 687M &
    \underline{59.5} & 62.0 & \underline{52.8} & 74.8 & \underline{50.5} & \underline{59.9} \\
    \bottomrule
    
    \end{tabular}}
\end{table*}
\begin{table*}
  \caption{\textbf{Evaluate foundation models on hands.} We study the scaling law of the amount of data and the model sizes and evaluate the hand pose estimation. The metrics are hand PA-PVE and hand PVE for all evaluation benchmarks. (PA-)MPE: (PA-)mean primary error. Best values are bolded and second best are underlined.
  AGORA uses the validation set, and EgoBody uses the EgoSet. Unit: mm.
  }
  \label{tab:foundation_model_hand}
  \centering
  \vspace{-2mm}
  \resizebox{\textwidth}{!}{
  \begin{tabular}{c|cccccccccccc|cc}
    \toprule
    
    Model &
    \multicolumn{2}{c}{SynHand}&
    \multicolumn{2}{c}{AGORA~\cite{patel2021agora}}&
    \multicolumn{2}{c}{EgoBody~\cite{zhang2022egobody}} &
    \multicolumn{2}{c}{UBody~\cite{lin2023one}} & 
    \multicolumn{2}{c}{ARCTIC~\cite{fan2023arctic}}&
    \multicolumn{2}{c}{EHF~\cite{Pavlakos_2019smplx}} &
    \multicolumn{2}{c}{Hand} \\
    
    \cmidrule(lr){2-3} \cmidrule(lr){4-5} \cmidrule(lr){6-7} 
    \cmidrule(lr){8-9} \cmidrule(lr){10-11} \cmidrule(lr){12-13} 
    \cmidrule(lr){14-15}
    &
    PA-PVE$\downarrow$ & PVE$\downarrow$ &
    PA-PVE$\downarrow$ & PVE$\downarrow$ &
    PA-PVE$\downarrow$ & PVE$\downarrow$ &
    PA-PVE$\downarrow$ & PVE$\downarrow$ &
    PA-PVE$\downarrow$ & PVE$\downarrow$ &
    PA-PVE$\downarrow$ & PVE$\downarrow$ &
    PA-MPE$\downarrow$ & MPE$\downarrow$\\
        
    \midrule
    \name-S5 & 15.6 & 59.7 & 10.2 & 66.8 & 10.8 & 53.3 & 11.8 & 59.4 & 16.7 & 58.7 & 16.0 & 64.0 & 13.5 & 60.3 \\
    \name-S10 & 16.0 & 54.4 & 10.2 & 65.2 & 10.0 & 48.6 & 11.5 & 57.4 & 17.5 & 56.6 & 16.0 & 59.1 & 13.5 & 56.9 \\
    \name-S20 & 15.4 & 50.8 & 10.0 & 63.3 & 10.0 & 47.2 & 11.1 & 49.6 & 18.9 & 45.2 & 15.5 & 54.7 & 13.5 & 51.8 \\
    \name-S32 & 14.6 & 49.3 & 9.8 & 61.9 & 10.0 & 46.0 & 10.7 & 47.8 & 18.9 & 42.9 & 14.8 & 54.6 & 13.1 & 50.4 \\
    \name-S40 & 12.3 & 42.3 & 9.1 & 55.4 & 9.5 & 42.7 & 9.8 & 44.8 & 19.2 & 39.8 & 13.5 & 51.6 & 12.2 & 46.1 \\
    \name-S40+ & 12.7 & 44.2 & 9.0 & 54.6 & 9.4 & 42.7 & 9.6 & 43.4 & 19.3 & 41.1 & 13.7 & 50.2 & 12.3 & 46.0 \\
    \midrule
    
    \name-B5 & 14.2 & 53.5 & 9.6 & 59.0 & 10.6 & 48.0 & 11.9 & 56.9 & 16.9 & 55.6 & 15.4 & 58.4 & 13.1 & 55.2 \\
    \name-B10 & 14.9 & 48.7 & 9.5 & 57.8 & 10.1 & 45.5 & 12.0 & 57.1 & 17.9 & 53.4 & 15.7 & 55.1 & 13.3 & 52.9 \\
    \name-B20 & 14.1 & 45.6 & 9.3 & 56.5 & 9.9 & 44.6 & 11.0 & 46.9 & 18.9 & 40.9 & 15.9 & 53.7 & 13.2 & 48.0 \\
    \name-B32 & 13.9 & 44.5 & 9.2 & 54.5 & 9.9 & 43.7 & 10.8 & 43.9 & 19.0 & 40.1 & 14.5 & 52.1 & 12.9 & 46.5 \\
    \name-B40 & 11.3 & 38.3 & 8.6 & 50.6 & 9.2 & 39.0 & 9.1 & 40.8 & 19.2 & 37.9 & 13.4 & 48.2 & 11.8 & 42.5 \\
    \name-B40+ & 11.8 & 38.2 & 8.5 & 48.3 & 9.3 & 38.8 & 8.8 & 38.7 & 18.9 & 38.1 & 13.3 & 46.3 & 11.8 & 41.4 \\
    \midrule
    
    \name-L5 & 13.9 & 49.6 & 9.2 & 53.0 & 10.5 & 45.2 & 12.5 & 56.3 & 17.0 & 52.8 & 14.7 & 57.8 & 13.0 & 52.5 \\
    \name-L10 & 14.7 & 46.4 & 9.1 & 51.9 & 10.0 & 43.1 & 12.8 & 56.1 & 18.1 & 50.8 & 15.6 & 52.9 & 13.4 & 50.2 \\
    \name-L20 & 14.4 & 43.7 & 8.9 & 51.0 & 9.9 & 42.7 & 10.6 & 43.3 & 18.9 & 39.3 & 15.0 & 49.4 & 13.0 & 44.9 \\
    \name-L32 & 13.5 & 42.4 & 8.7 & 47.8 & 9.8 & 41.4 & 10.2 & 39.2 & 18.9 & 38.8 & 14.1 & 47.1 & 12.5 & 42.8 \\
    \name-L40 & 10.7 & 34.2 & 8.0 & 42.8 & 8.9 & 36.5 & 8.4 & 35.7 & 19.2 & 36.1 & 12.8 & 43.1 & 11.3 & 38.1 \\
    \name-L40+ & 11.1 & 33.0 & 8.0 & 43.2 & 9.0 & 36.5 & 8.2 & 34.2 & 19.0 & 36.6 & 13.3 & 46.3 & 11.4 & 38.3 \\
    \midrule
    
    \name-H5 & 13.4 & 47.6 & 9.1 & 52.6 & 10.5 & 43.5 & 12.1 & 53.3 & 17.4 & 49.3 & 14.3 & 55.6 & 12.8 & 50.3 \\
    \name-H10 & 14.6 & 44.7 & 9.0 & 51.4 & 10.0 & 42.6 & 12.6 & 54.8 & 18.8 & 49.3 & 15.6 & 50.2 & 13.4 & 48.8 \\
    \name-H20 & 14.8 & 41.9 & 8.8 & 49.5 & 9.8 & 41.3 & 10.3 & 41.0 & 18.9 & 38.3 & 14.4 & 47.1 & 12.8 & 43.2 \\
    \name-H32 & 14.0 & 42.7 & 8.5 & 45.6 & 9.8 & 39.6 & 9.8 & 36.4 & 18.8 & 37.0 & 14.8 & 42.2 & 12.6 & 40.6 \\
    \name-H40 & \underline{10.3} & 32.0 & 8.0 & 41.8 & \underline{8.9} & \underline{34.8} & \underline{8.1} & 34.7 & 18.9 & 35.0 & 12.1 & 39.9 & 11.0 & 36.4 \\
    \name-H40+ & 10.8 & 31.5 & \underline{7.9} & 40.1 & \textbf{8.8} & \textbf{34.7} & \textbf{7.9} & \underline{33.1} & 19.1 & 36.3 & 11.9 & 38.1 & 11.1 & 35.6 \\
    \midrule
    
    \namepami-H5 & 15.0 & 48.7 & 15.2 & 57.1 & 10.7 & 44.4 & 13.1 & 48.9 & 13.2 & 39.6 & 14.8 & 53.6 & 13.7 & 48.7 \\
    \namepami-H10 & 12.6 & 39.8 & 8.3 & 41.4 & 9.2 & 35.9 & 11.4 & 49.0 & 13.2 & 37.6 & 12.9 & 42.0 & 11.3 & 40.9 \\
    \namepami-H20 & 12.6 & 41.6 & 8.9 & 45.2 & 9.2 & 37.0 & 9.5 & 37.7 & 11.9 & 28.8 & 13.1 & 44.8 & 10.9 & 39.2 \\
    \namepami-H32 & 12.4 & 38.6 & 8.2 & 43.6 & 9.0 & 35.7 & 9.0 & 35.3 & 11.6 & 26.5 & 12.1 & 41.5 & 10.4 & 36.9 \\
    \namepami-H40 & \textbf{9.8} & \textbf{23.7} & \textbf{7.3} & \textbf{36.9} & \textbf{8.8} & 35.6 & \textbf{7.9} & \textbf{32.9} & \textbf{10.9} & \underline{24.1} & \textbf{11.4} & \textbf{36.4} & \textbf{9.3} & \textbf{31.6} \\
    \namepami-H40+ & 10.8 & \underline{25.1} & \underline{7.9} & \underline{38.4} & 9.0 & 36.3 & \textbf{7.9} & 34.0 & \underline{11.1} & \textbf{23.6} & \underline{11.7} & \underline{38.8} & \underline{9.7} & \underline{32.7} \\
    \bottomrule
    \end{tabular}}
\end{table*}
\begin{table}[t]
\centering
\caption{\textbf{Mean Primary Error (MPE) of whole-body and hand pose estimation.} We evaluate EHPS methods on multiple benchmark datasets for whole-body and hand pose estimation. Unit: mm.}
\label{tab:mpe}
\vspace{0mm}
\resizebox{\linewidth}{!}{
    \begin{tabular}{lc|ccc}
    \toprule
    
    
    Method & \#Params &
    MPE$\downarrow$ & Hand-MPE$\downarrow$ & Hand-PA-MPE$\downarrow$ \\
    
    \midrule
    PyMAF-X~\cite{pymafx2023} & 206M & 120.3 & 62.7 & 11.7 \\
    Hand4Whole~\cite{GyeongsikMoon2020hand4whole} & 78M & 115.6 & 51.4 & 11.4 \\
    HybrIK-X~\cite{li2023hybrik} & 76M & 115.0 & 49.0 & 14.0 \\
    OSX~\cite{lin2023one} & 312M & 111.6 & 55.8 & 14.0 \\
    AiOS~\cite{sun2024aios} & 70M & 70.6 & 41.9 & 11.4 \\

    \midrule
    \name-S40+ & 32M & 77.3 & 46.0 & 12.3 \\
    \name-B40+ & 103M & 70.0 & 41.4 & 11.8 \\
    \name-L40+ & 327M & 63.7 & 38.3 & 11.4 \\
    \name-H40+ & 662M & 59.9 & 35.6 & 11.1 \\
    \namepami-H40 & 687M & \textbf{58.9} & \textbf{31.6} & \textbf{9.3} \\
    
    \bottomrule
    \end{tabular}
}
\end{table}
\begin{figure}[t!]
  \centering
  \vspace{0mm}
  \includegraphics[width=\linewidth]{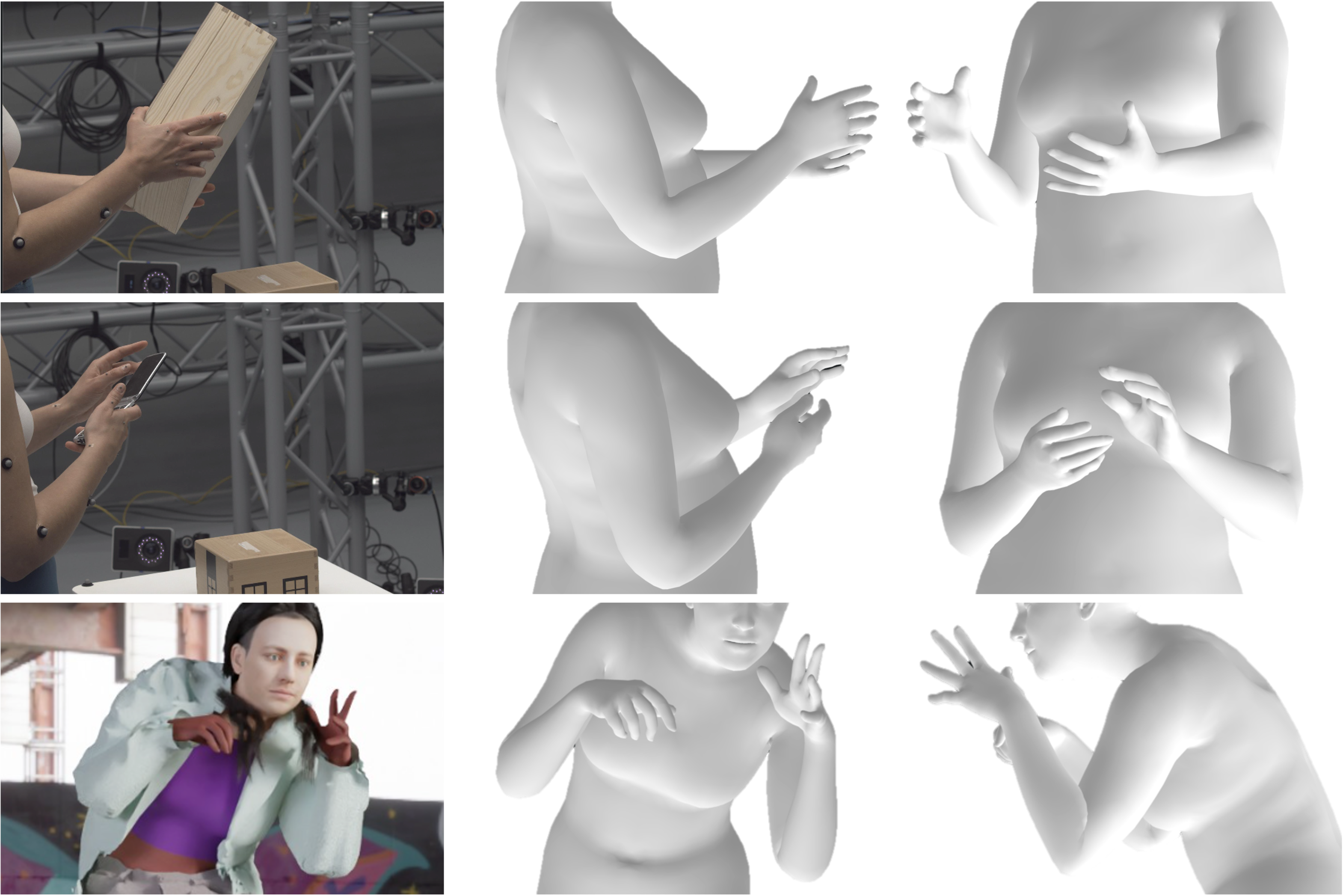}
 \caption{\textbf{Visualization of hands by \namepami.} \namepami demonstrates robust hand pose estimation in whole-body pose and shape estimation tasks across various scenarios, including occlusion (top), object interaction (middle), and challenging hand poses (bottom).}
  \label{fig:visualization_qualitative_hand}
\end{figure}

\begin{figure*}[t]
  \centering
  \includegraphics[width=\linewidth]{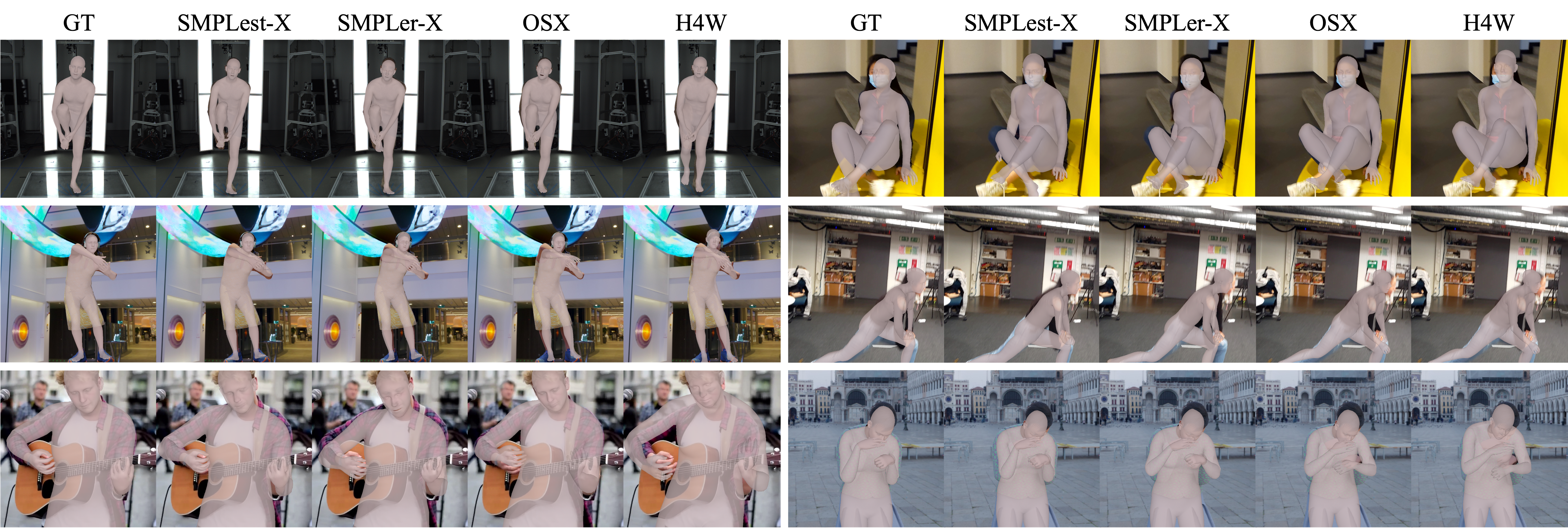}
  \caption{\textbf{Visualization.} We compare \namepami-L40 and \name-L40 with OSX~\cite{lin2023one} and Hand4Whole (H4W)~\cite{GyeongsikMoon2020hand4whole} (trained with the MSCOCO, MPII, and Human3.6M) in various scenarios.}
  \label{fig:visualization_qualitative}
\end{figure*}

\subsection{Generalist Foundation Models}
\subsubsection{Data Scaling}
\label{sec:foundation_model:training}

The SOTA methods~\cite{lin2023one, li2023hybrik} usually train with only a few (\eg, MSCOCO, MPII, and Human3.6M) datasets, whereas we investigate training with many more datasets. However, we highlight that the dataset benchmark in \Tab\ref{tab:single_datasets} cannot be used: selecting datasets based on their performance on the test sets of the evaluation benchmarks leaks information about the test sets. Hence, we construct another dataset benchmark in the \Supp, which ranks individual datasets on the \textit{training} set of the major EHPS benchmarks. \textcolor{blue}{We experiment with five dataset counts: 5, 10, 20, 32, and 40 as the training set, corresponding to six different total instance lengths: 0.75M, 1.5M, 3.0M, 4.5M, 7.5M, and 10M (denoted as 40+). For scaling up to 32 datasets, we prioritize higher-ranked datasets in the aforementioned benchmark in \Supp. For scaling up further to 40 datasets, we incorporate additional open-sourced datasets and explore the impact of further increasing instance counts within the same data domain.} To prevent larger datasets from shadowing smaller datasets, we adopt a balanced sampling strategy. Specifically, all selected datasets are uniformly upsampled or downsampled to the same length and add up to the designated total length. \textcolor{blue}{To facilitate training, we standardize all datasets into the HumanData format~\cite{mmhuman3d}.}

\subsubsection{Model Scaling}
 We also study four ViT backbones of different sizes (ViT-Small, Base, Large and Huge). The training is conducted on 16 V100 GPUs, with a total batch size of 512 (256 for ViT-Huge) for 10 epochs \textcolor{blue}{(5 epochs for ViT-Huge training with 7.5M and 10M instances from 40 datasets)}. \textcolor{blue}{Specifically, \name-L20 and \namepami-L20 take more than 400 GPU hours, \name-H32 and \namepami-H32 take more than 700 GPU hours, \name-H40+ and \namepami-H40+ take more than 2600 GPU hours to train.} We use Adam optimizer with cosine annealing for both training and fine-tuning. The learning rate for training is \num{1e-5} with the minimum learning rate set to \num{1e-6}, while the learning rate for finetuning is \num{1e-5} with the minimum learning rate set to \num{5e-7}.
 

\subsubsection{Evaluating the Generalists}
In \Tab\ref{tab:foundation_model}, \textcolor{blue}{\Tab\ref{tab:foundation_model_hand} and \Tab\ref{tab:mpe}, we show experimental results of data scaling and model scaling with proposed minimalistic framework \name and \namepami}. It is observed that:
1) More training \textcolor{blue}{dataset leads to better performance in terms of MPE, hand-MPE and hand-PA-MPE.} The model performance improves gradually as the number of training datasets increases, \textcolor{blue}{while with the same training data domain, improving the number of instances does not guarantee the improvement of performance.}
2) A larger foundation model performs better at any given amount of data. However, the marginal benefits of scaling up decrease beyond a sufficiently large model size. Specifically, ViT-H has more than twice the parameters than ViT-L, but the performance gain is not prominent. 
3) The foundation model always performs better than in-domain training on a single training set. For example, \name-B20, performs better on the test sets of AGORA-val, UBody, EgoBody, and 3DPW, than models trained specifically on the corresponding training set in \Tab\ref{tab:single_datasets}. 
4) \textcolor{blue}{By removing the component guiding module in \name and replacing it with a decoder, \namepami has significant improvement in hand pose estimation (15\% lower in hand-PA-MPE and 13\% lower in hand-MPE) as shown in \Tab\ref{tab:mpe}.}
In addition, we compare our foundation model with SOTA methods across various scenarios, as shown in \Fig\ref{fig:visualization_qualitative} for whole-body qualitative evaluation and \Fig\ref{fig:visualization_qualitative_hand} for diverse hand poses in the circumstance of severe occlusion, object interaction and challenging poses.

Moreover, we show detailed by-part results of body, hands, and face on main benchmarks such as AGORA test set (\Tab\ref{tab:agora_test}), AGORA validation set (\Tab\ref{tab:agora_val}), UBody (\Tab\ref{tab:ubody}), EgoBody-EgoSet (\Tab\ref{tab:egobody}), EHF (\Tab\ref{tab:ehf}), \textcolor{blue}{ARCTIC(\Tab\ref{tab:arctic}) and DNA-Rendering (\Tab\ref{tab:renbody})}. We highlight that the foundation models show strong and balanced performances on all benchmarks. 

\subsection{Specialist Models}
\subsubsection{Finetuning the Specialists}
\begin{table*}
  \caption{\textbf{AGORA test set.} $\dagger$ denotes the methods that are finetuned on the AGORA training set. $\ast$denotes the methods that are trained on AGORA training set only.}
  \label{tab:agora_test}
  \centering
  \vspace{-2mm}
  \resizebox{\textwidth}{!}{
  \begin{tabular}{lccccccccccccccc}
    \toprule
    & 
    \multicolumn{2}{c}{NMVE$\downarrow$ (mm)} &
    \multicolumn{2}{c}{NMJE$\downarrow$ (mm)} &
    \multicolumn{5}{c}{MVE$\downarrow$ (mm)} &
    \multicolumn{5}{c}{MPJPE$\downarrow$ (mm)} \\
    
    \cmidrule(lr){2-3} \cmidrule(lr){4-5} \cmidrule(lr){6-10} \cmidrule(lr){11-15}  
    Method &
    All & Body &
    All & Body &
    All & Body & Face & LHand & RHand &
    All & Body & Face & LHand & RHhand 
    \\
    \midrule
    BEDLAM~\cite{black2023bedlam}& 
    179.5 & 132.2 & 177.5 & 131.4 & 131.0 & 96.5 & {25.8} & {38.8} & {39.0} & 129.6 & 95.9 & {27.8} & {36.6} & {36.7} \\
    
    Hand4Whole~\cite{GyeongsikMoon2020hand4whole}$\dagger$& 
    144.1 & 96.0 & 141.1 & 92.7 & 135.5 & 90.2 & 41.6 & 46.3 & 48.1 & 132.6 & 87.1 & 46.1 & 44.3 & 46.2 \\
    
    BEDLAM~\cite{black2023bedlam}$\dagger$ & 
    142.2 & 102.1 & 
    141.0 & 101.8 & 
    {103.8} & 74.5 & \textbf{23.1} & \underline{31.7} & 
    \underline{33.2} & {102.9} & 74.3 & \textbf{24.7} & \underline{29.9} & \underline{31.3} \\
    
    PyMaF-X~\cite{HongwenZhang2022PyMAFXTW}$\dagger$ &
    141.2 & 94.4 & 
    140.0 & 93.5 & 
    125.7 & 84.0 & 35.0 & 44.6 & 
    45.6 & 124.6 & 83.2 & 37.9 & 42.5 & 43.7 \\
    
    OSX~\cite{lin2023one} $\ast$&
    130.6 & 85.3 & 
    127.6 & 83.3 & 
    122.8 & 80.2 & 36.2 & 45.4 & 
    46.1 & 119.9 & 78.3 & 37.9 & 43.0 & 43.9 \\

    HybrIK-X~\cite{li2023hybrik} & 
    {120.5} & {73.7} & 
    {115.7} & {72.3} & 
    112.1 & {68.5} & 37.0 & 46.7 & 
    47.0 & 107.6 & {67.2} & 38.5 & 41.2 & 41.4 \\


    Multi-HMR~\cite{multi-hmr2024} & 
    102.0 & 63.4 & 101.8 & 64.1 & 95.9 & 59.6 & 27.7 & 40.2 & 40.9 & 95.7 & 60.3 & 29.2 & 38.1 & 39.0 \\

    NLF-L~\cite{sarandi24nlf} & 98.6 & \underline{62.1}	& 96.6	& \underline{61.9}	& 92.7	& 58.4	& 27.0	& 37.9 & 38.1 &	90.8 &	58.2 & 28.5 &	34.4 & 34.9 \\

    AiOS~\cite{sun2024aios} & \underline{97.8}	& \textbf{61.3} & \underline{96.0} & \textbf{60.7} & \underline{91.9} &	\underline{57.6} & \underline{24.6} &	38.7 & 39.6 &	\underline{90.2} &	\underline{57.1} &	\underline{25.7} & 	36.4 & 37.3 \\
    
    \name-L20$\dagger$& 
    107.2 & {68.3} & 
    {104.1} & {66.3} & 
    {99.7} & {63.5} & {29.9} & {39.1} &
    {39.5} & {96.8} & {61.7} & {31.4} & {36.7} & 37.2 \\

    \namepami-H40$\dagger$& 
    \textbf{96.2} & {63.8} & 
    \textbf{93.9} & {62.7} & 
    \textbf{86.6} & \textbf{57.4} & {25.4} & \textbf{30.8} &
    \textbf{31.5} & \textbf{84.5} & \textbf{56.4} & \underline{25.7} & \textbf{29.0} & \textbf{29.7} \\

    \bottomrule
  \end{tabular}}
\end{table*}
    
    
    
    
    
    
    

\begin{table*}[t]
  \begin{minipage}{0.48\linewidth}
    \centering
    \caption{\textbf{AGORA Val set}. $\dagger$ and $\ast$ are finetuned on the AGORA training set,  and trained on the AGORA training set only, respectively.}
    \label{tab:agora_val}
    \vspace{-2mm}
    \resizebox{\textwidth}{!}{
    \begin{tabular}{lcccccc}
    \toprule
    & 
    \multicolumn{3}{c}{PA-PVE$\downarrow$ (mm)} &
    \multicolumn{3}{c}{PVE$\downarrow$ (mm)} \\
    
    \cmidrule(lr){2-4} \cmidrule(lr){5-7} 
    
    Method &
    All & Hands & Face &
    All & Hands & Face \\
    
    \midrule

    Hand4Whole~\cite{GyeongsikMoon2020hand4whole}$\dagger$ & 
    73.2 & 9.7 & 4.7 & 183.9  & 72.8 & 81.6 \\

    PyMAF-X~\cite{pymafx2023} &
    75.2 & 12.5 & 3.8 & 175.0 & 119.0 & 76.5 \\

    HybrIK-X~\cite{li2023hybrik} &
    44.6 & 9.6 & 4.7 & 69.0 & 44.2 & 35.2 \\

    OSX~\cite{lin2023one} & 
    69.4 & 11.5 & 4.8 & 168.6 & 70.6 & 77.2 \\
    
    OSX~\cite{lin2023one}$\ast$ & 
    {45.0} & {8.5} & {3.9} & 79.6 & 48.2 & 37.9 \\
    
    AiOS~\cite{sun2024aios} & 43.0 & 8.5 & 3.1 & 63.4 & 46.1 & 31.7  \\
    \name-B1$\ast$ &
    48.9 & {8.6} & 4.0 & 86.1 & 51.5 & 41.2 \\
    
    \name-L20 & 
    48.6 & 8.9 & 4.0 & 80.7 & 51.0 & 41.3 \\
    
    
    \name-L20$\dagger$ & 
    {39.1} & 9.3 & {3.8} & {62.5} & {42.3} & {32.8} \\

    \namepami-H40 & 
    \underline{39.2} & \textbf{7.3} & \textbf{2.7} & \textbf{56.1} & \textbf{36.9} & \textbf{26.5} \\

    \namepami-H40$\dagger$ & 
    \textbf{37.2} & \underline{7.8} & \underline{2.8} & \underline{57.4} & \underline{37.7} & \underline{30.6} \\
    \bottomrule
    \end{tabular}}
  \end{minipage}\hfill
  \begin{minipage}{0.48\linewidth}
    \centering
    \caption{\textbf{EHF}. 
    As EHF does not have a training set to benchmark datasets, we finetune the model with 5 selected datasets. As EHF is not seen in our training and can be used to validate our foundation models' transferability.
    }
    \label{tab:ehf}
    \vspace{-2mm}
    \resizebox{\textwidth}{!}{
    \begin{tabular}{lcccccc}
    \toprule
    & 
    \multicolumn{3}{c}{PA-PVE$\downarrow$ (mm)} &
    \multicolumn{3}{c}{PVE$\downarrow$ (mm)} \\
    
    \cmidrule(lr){2-4} \cmidrule(lr){5-7} 
    
    Method &
    All & Hands & Face &
    All & Hands & Face \\
    
    \midrule
    Hand4Whole~\cite{GyeongsikMoon2020hand4whole} & 
    50.3 & \underline{10.8} & 5.8 & 76.8 & {39.8} & 26.1\\

    PyMAF-X~\cite{pymafx2023} &
    50.2 & \textbf{10.2} & 5.5 & 64.9 & \textbf{29.7} & 19.7 \\

    HybrIK-X~\cite{li2023hybrik} &
    58.6 & 16.9 & 7.3 & 121.5 & 53.3 & 41.4 \\

    OSX~\cite{lin2023one} & 
    48.7 & 15.9 & 6.0 & 70.8 & 53.7 & 26.4\\
    
    AiOS~\cite{sun2024aios} & \underline{34.0} & 12.8 & {3.8} & {45.4} & 44.1 & {16.9} \\

    NLF-L~\cite{sarandi24nlf} & \textbf{26.0} & 20.7 & 6.3 & \textbf{36.4} & 43.0 & \textbf{13.9} \\
    
    \name-L20 & 
    {37.8} & {15.0} & {5.1}& {65.4} & {49.4} & {17.4}\\
    
    

    \namepami-H40 & 
    {34.7} & {11.4} & \textbf{3.4}& {49.7} & {36.4} & \underline{16.2} \\

    \namepami-H40$\dagger$ & 
    \underline{34.0} & {11.2} & \underline{3.7}& \underline{40.9} & \underline{33.3} &{20.2} \\

    \bottomrule
    \end{tabular}}
  \end{minipage}
 \vspace{-0.3cm} 
\end{table*}

\begin{table*}[t]
  \begin{minipage}{0.48\linewidth}
    \centering
    \caption{\textbf{UBody.} $\dagger$ denotes the methods that are finetuned on the UBody training set. $\ast$ denotes the methods that are trained on UBody training set only. }
    \label{tab:ubody}
    \vspace{-2mm}
    \resizebox{\textwidth}{!}{
    \begin{tabular}{lcccccc}
    \toprule
    & 
    \multicolumn{3}{c}{PA-PVE$\downarrow$ (mm)} &
    \multicolumn{3}{c}{PVE$\downarrow$ (mm)} \\
    
    \cmidrule(lr){2-4} \cmidrule(lr){5-7} 
    
    Method &
    All & Hands & Face &
    All & Hands & Face \\
    
    \midrule
    PIXIE~\cite{Feng_2021_pixie} & 
    61.7 & 12.2 & 4.2 & 168.4 & 55.6 & 45.2 \\
    
    Hand4Whole~\cite{GyeongsikMoon2020hand4whole} & 
    44.8 & {8.9} & 2.8 & 104.1 & 45.7 & 27.0 \\

    PyMAF-X~\cite{pymafx2023} &
    78.6 & 11.2 & 6.2 & 146.7 & 56.0 & 37.6 \\

    HybrIK-X~\cite{li2023hybrik} &
    87.0 & 14.2 & 11.2 & 201.6 & 57.2 & 70.7 \\

    OSX~\cite{lin2023one} & 
    42.4 & 10.8 & \underline{2.4} & 92.4 & 47.7 & 24.9 \\

    OSX~\cite{lin2023one}$\dagger$ & 
    42.2 & {8.6} & \textbf{2.0} & 81.9 & 41.5 & \underline{21.2} \\

    AiOS~\cite{sun2024aios} & 32.5 & \textbf{7.3} & 2.8 & 58.6 & 39.0 & 19.6 \\
    
    \name-B1$\ast$ &
    38.5 & 10.8 & 3.0 & 64.8 & 45.4 & 22.3 \\
    
    \name-L20 & 
    33.2 & 10.6 & 2.8 & 61.5 & 43.3 & 23.1 \\
    
    
    \name-L20$\dagger$ & 
    {31.9} & 10.3 & 2.8 & {57.4} & {40.2} & {21.6} \\
    
    \namepami-H40& 
    \underline{27.8} & \underline{7.9} & 2.5 & \underline{51.1} & \underline{32.9} & {21.4} \\
    
    \namepami-H40$\dagger$ & 
    \textbf{26.9} & 8.4 & \underline{2.4} & \textbf{50.2} & \textbf{31.8} & \textbf{18.9} \\
    \bottomrule
    \end{tabular}}
  \end{minipage}\hfill
  \begin{minipage}{0.48\linewidth}
    \centering
    \caption{\textbf{EgoBody-EgoSet.} $\dagger$ denotes the methods that are finetuned on the EgoBody-EgoSet training set. $\ast$ denotes the methods that are trained on EgoBody-EgoSet training set only.}
    \label{tab:egobody}
    \vspace{-2mm}
    \resizebox{\textwidth}{!}{
    \begin{tabular}{lcccccc}
    \toprule
    & 
    \multicolumn{3}{c}{PA-PVE$\downarrow$ (mm)} &
    \multicolumn{3}{c}{PVE$\downarrow$ (mm)} \\
    
    \cmidrule(lr){2-4} \cmidrule(lr){5-7} 
    
    Method &
    All & Hands & Face &
    All & Hands & Face \\
    
    \midrule
    Hand4Whole~\cite{GyeongsikMoon2020hand4whole} & 
    58.8 & {9.7} & 3.7 & 121.9 & 50.0 & 42.5 \\

    PyMAF-X~\cite{pymafx2023} &
     66.3 & 9.7 & 3.5 & 121.8 & 53.4 & 45.9 \\

    HybrIK-X~\cite{li2023hybrik} &
    52.7 & 11.2 & 5.9 & 111.4 & 44.6 & 44.5 \\

    OSX~\cite{lin2023one} & 
    54.6 & 11.6 & 3.7 & 115.7 & 50.6 & 41.1 \\

    OSX~\cite{lin2023one}$\dagger$ & 
    45.3 & 10.0 & {3.0} & 82.3 & 46.8 & 35.2 \\

    AiOS~\cite{sun2024aios} & 38.0 & \underline{9.0} & \underline{2.9} & 61.6 & 40.0 & 26.7 \\
    
    \name-B1$\ast$ & 
    56.1 & 10.7 & 3.5 & 87.2 & 49.4 & 34.9 \\
    
    \name-L20 & 
    38.9 & 9.9 & {3.0} & 66.6 & 42.7 & 31.8\\
    
    
    \name-L20$\dagger$ & 
    {37.8} & 9.9 & \underline{2.9} & {63.6} & {42.5} & {30.8} \\

    \namepami-H40 &  \underline{33.6} & \textbf{8.8} &	\textbf{2.5} &	\underline{61.5} &	\underline{35.6} & \underline{28.5} \\

    \namepami-H40$\dagger$ & \textbf{33.5} & \textbf{8.8} & \textbf{2.5} & \textbf{59.3} & \textbf{35.3} & \textbf{28.0} \\
    \bottomrule
    \end{tabular}}
  \end{minipage}
\end{table*}

\begin{table*}[htbp]
  \begin{minipage}{0.48\linewidth}
    \centering
    \caption{\textbf{ARCTIC.} $\dagger$ and $\ast$ denote the methods that are finetuned on the ARCTIC training set and trained on the ARCTIC training set only, respectively.}
    \label{tab:arctic}
    \vspace{-2mm}
    \resizebox{\textwidth}{!}{
    \begin{tabular}{lcccccc}
    \toprule
    & 
    \multicolumn{3}{c}{PA-PVE$\downarrow$ (mm)} &
    \multicolumn{3}{c}{PVE$\downarrow$ (mm)} \\
    
    \cmidrule(lr){2-4} \cmidrule(lr){5-7} 
    
    Method &
    All & Hands & Face &
    All & Hands & Face \\
    
    \midrule
    Hand4Whole~\cite{GyeongsikMoon2020hand4whole} & 
    63.4 & 18.1 & 4.0 & 136.8 & 54.8 & 59.2 \\

    PyMAF-X~\cite{pymafx2023} &
    69.4 & 14.0 & 3.1 & 137.9 & 58.8 & 53.6 \\

    HybrIK-X~\cite{li2023hybrik} &
    63.4 & 15.4 & 5.6 & 147.4 & 44.5 & 53.9 \\

    OSX~\cite{lin2023one} & 
    56.9 & 17.5 & 3.9 & 102.6 & 56.5 & 44.6 \\
    
    OSX~\cite{lin2023one}$\dagger$ & 
    33.0 & 18.8 & 3.3 & 58.4 & 39.4 & 30.4 \\
    
    AiOS~\cite{sun2024aios} & 30.2 & 19.2 &  \textbf{2.1} & 47.1 & 38.3 & \textbf{26.1} \\
    
    \name-B1$\ast$ & 
    45.2 & {18.9} & 3.4 & 66.6 & 42.5 & 34.0 \\

    \name-L10 & 
    46.9 & {18.1} & \underline{2.3} & 76.9 & 50.8 & 33.2 \\
    
    
    \name-L10$\dagger$ & 
    {33.1} & 19.0 & {2.7} & {54.9} & {40.1} & {27.3} \\

    \namepami-H40 & \underline{24.2} & \underline{10.9} & {3.3} & \underline{40.3} & \underline{24.1} & 28.0 \\
    
    \namepami-H40$\dagger$ & \textbf{23.2} & \textbf{10.7} & {3.2} & \textbf{38.1} & \textbf{22.6} & \underline{27.1}\\
    
    \bottomrule
    \end{tabular}}
  \end{minipage} \hfill
  \begin{minipage}{0.48\linewidth}
    \centering
    \caption{\textbf{DNA-Rendering-HiRes}. $\dagger$ and $\ast$ are finetuned on the DNA-Rendering-HiRes training set and trained on the DNA-Rendering-HiRes training set only, respectively.}
    \label{tab:renbody}
    \vspace{-2mm}
    \resizebox{\textwidth}{!}{
    \begin{tabular}{lcccccc}
    \toprule
    & 
    \multicolumn{3}{c}{PA-PVE$\downarrow$ (mm)} &
    \multicolumn{3}{c}{PVE$\downarrow$ (mm)} \\
    
    \cmidrule(lr){2-4} \cmidrule(lr){5-7} 
    
    Method &
    All & Hands & Face &
    All & Hands & Face \\
    
    \midrule
    
    Hand4Whole~\cite{GyeongsikMoon2020hand4whole} & 
    62.8 & 11.0 & 4.2 & 111.4 & 56.4 & 52.6 \\

    PyMAF-X~\cite{pymafx2023} &
    54.6 & 11.0 & 4.3 & 102.8 & 55.5 & 52.6 \\

    HybrIK-X~\cite{li2023hybrik} &
    50.7 & 10.9 & 5.7 & 92.0 & 47.8 & 48.2\\

    OSX~\cite{lin2023one} & 
    59.9 & 10.6 & 4.3 & 105.7 & 55.0 & 52.5 \\

    OSX~\cite{lin2023one}$\dagger$ & 
    43.5 & 7.5 & 3.5 & 67.1 & 43.3 & 38.2 \\

    \name-B1$\ast$ & 
    45.6 & 7.5 & 3.4 & 63.2 & 40.7 & 34.2 \\

    \name-L20 & 
    44.4 & 11.1 & 4.5 & 77.7 & 47.5 & 43.2 \\
    
    
    \name-L20$\dagger$ & 
    {37.9} & \underline{7.3} & {3.4} & {56.5} & {38.4} & {34.9} \\

    \namepami-H40 & 
    \underline{33.6} & 7.4 & \textbf{3.1} & \underline{56.2} & \underline{34.6} & \underline{32.0} \\

    \name-H40$\dagger$ & 
    \textbf{32.0} & \textbf{7.1} & \underline{3.2} & \textbf{51.7} & \textbf{32.8} & \textbf{31.9} \\
    
    \bottomrule
    \end{tabular}}
  \end{minipage}
\end{table*}

Training the foundation model with a large number of data is expensive. For example, \textcolor{blue}{\name-H40+ and \namepami-H40+ takes more than 2600 GPU hours to train.} Hence, it is critical to investigate finetuning strategies that allow for low-cost adaptation of the foundation model to specific scenarios. We reiterate that in real-life applications, the \textit{test set} is inaccessible. \textcolor{blue}{Hence, besides the respective training set of the target dataset, we use our dataset selection strategy based on the benchmarking in the \Supp and select four high-ranking datasets} on the target train set to finetune the model for five epochs. 

\subsubsection{Evaluating the Specialists}
We finetune the generalist model with ViT-L backbone to match the current SOTA~\cite{lin2023one} \textcolor{blue}{and the generalist models with ViT-H backbone to explore the limit of \name and \namepami specialists. The evaluation of specialists are shown in \Tab\ref{tab:agora_test}, \Tab\ref{tab:agora_val}, \Tab\ref{tab:ehf}, \Tab\ref{tab:ubody}, \Tab\ref{tab:egobody}, \Tab\ref{tab:arctic} and \Tab\ref{tab:renbody}.} 
Moreover, we evaluate the transferability of our foundation models on \textcolor{blue}{EHF (\Tab\ref{tab:ehf}),} ARCTIC (\Tab\ref{tab:arctic}) and DNA-Rendering (\Tab\ref{tab:arctic}). \textcolor{blue}{EHF is a test-only benchmark without training set provided, }ARCTIC features complicated hand-object interaction with whole-body annotations, and DNA-Rendering includes diverse subjects, motions, and garments. Note that ARCTIC is not seen by foundation models trained on top 10 datasets, and DNA-Rendering is not seen by foundation models trained on top 20 datasets. The foundation models, however, achieve much better performance than SOTAs with conventional data sampling strategies.

\subsection{Training Domains}
\begin{table*}
  \caption{\textbf{Impact of in-domain training.} We investigate the impact of seeing the train split of a benchmark dataset during training and how this may affect the generalizability of a model. The highlighted yellow shaded numbers denote that the corresponding train split is used in training. Except for 3DPW using MPJPE as the metric, other datasets are evaluated via PVE. The lower the better for all the metrics. Top-1 values are bolded, and the second best values are underlined. EgoBody: EgoBody-EgoSet. AGORA: AGORA-Val. \#Row: Row number. \#Data.: number of datasets. \#Seen: number of evaluation benchmarks' train splits used in the training. \#Inst.: number of training instances. Unit: mm.
  }
  \label{tab:dataset_seen}
  \centering
  \resizebox{\textwidth}{!}{
  \begin{tabular}{c|cccc|c|ccccc|cc}
    \toprule
    \#Row & \#Data. & \#Seen & \#Inst. & Model&
    MPE$\downarrow$ &
    AGORA~\cite{patel2021agora} &
    EgoBody~\cite{zhang2022egobody} &
    UBody~\cite{lin2023one} & 
    3DPW~\cite{von2018recovering} &
    EHF~\cite{Pavlakos_2019smplx} &
    ARCTIC~\cite{fan2023arctic} &
    DNA-R-HiRes~\cite{renbody}\\
    \midrule

    R1 & 4 & 4 & 0.75M & \name-L & 83.8 & \seen{98.1} &  \seen{80.0} &  \seen{65.0} &  \seen{96.2} & 79.9 & 101.5 & 101.3 \\
    R2 & 5 & 1 & 0.75M & \name-L5 & 97.0 &  \seen{88.3} & 98.7 & 110.8 & 97.8 & 89.5 & 95.1 & 85.9 \\
    R3 & 5 & 4 & 0.75M & \name-L & 85.2 &  \seen{96.7} &  \seen{81.9} &  \seen{68.1} &  \seen{95.5} & 83.6 & 103.6 & 98.2 \\
    R4 & 10 & 2 & 0.75M & \name-L10 & 80.6 &  \seen{82.6} &  \seen{69.7} & 104.0 & 82.5 & 64.0 & 76.9 & 76.2 \\
    R5 & 10 & 4 & 1.5M & \name-L & 72.7 &  \seen{84.0} &  \seen{71.6} &  \seen{62.8} &  \seen{81.7} & {63.4} & 80.8 & {75.8} \\
    R6 & 20 & 5 & 3M & \name-L20 & {70.5} &  \seen{{80.7}} &  \seen{{66.6}} &  \seen{{61.5}} &  \seen{{78.3}} & 65.4 &  \seen{{52.2}} & 77.7 \\
    R7 & 32 & 6 & 4.5M & \name-L32 & {66.2} &  \seen{{74.2}} &  \seen{{62.2}} &  \seen{{57.3}} &  \seen{\underline{75.2}} & {62.3} &  \seen{{48.6}} &  \seen{\textbf{54.4}} \\
    R8 & 40 & 6 & 7.5M & \namepami-H40 & \textbf{58.9} &  \seen{\textbf{56.1}} &  \seen{\textbf{61.5}} &  \seen{\textbf{51.1}} &  \seen{{76.0}} & \textbf{49.7} &  \seen{\textbf{40.3}} &  \seen{\underline{56.2}} \\
    R9 & 40 & 6 & 10.0M & \namepami-H40+ & \underline{59.9} &  \seen{\underline{59.5}} &  \seen{\underline{62.0}} &  \seen{\underline{52.8}} &  \seen{\textbf{74.8}} & \underline{50.5} &  \seen{\underline{41.2}} &  \seen{{56.3}} \\
    \bottomrule
    \end{tabular}}
\end{table*}
In \Tab\ref{tab:dataset_seen}, we study the impact of using the training split of an evaluation benchmark on its corresponding test split. We have two key observations. First, in-domain training (including the training split of a dataset in training and testing on the test split of the same dataset) is highly effective, as ``seeing" the dataset always brings significant performance improvement \textcolor{blue}{(R2 vs. R3)}. Second, having out-of-domain training sets helps if the training data is sufficiently diverse. For example, \name-L with four seen datasets benefitted little from having one additional dataset (5 datasets in total with 85.2 mm MPE) but tremendously from having 10 datasets in training (with 72.7 mm MPE). It is worth noting that training with a lot of datasets also benefits out-of-domain (``unseen" benchmarks) performance as errors on EHF (by 20.2 mm PVE reduction), ARCTIC (by 22.8 mm PVE reduction), and DNA-Rendering (by 22.4 mm PVE reduction) decrease with more datasets in the training set and four seen datasets fixed \textcolor{blue}{(R3 vs. R5). With the amount of training data limited and fixed, training with out-of-domain data with more varieties brings more benefits than in-domain data (R3 vs. R4)}. \textcolor{blue}{Finally, training on 40 datasets with \namepami-H achieves the best performance with a 58.9 mm MPE, demonstrating it as a robust and effective SMPL-X estimator. However, further increasing the number of instances from 7.5M to 10.0M within the same training domain (comprising the 40 datasets) does not promise a performance improvement (R8 vs. R9).}

\subsection{Data Sampling Strategies}
\begin{table}[t]
  \caption{\textbf{Data sampling strategies}. We trained \name-H32 models with different data sampling strategies. Unit: mm.}
  \label{tab:sampling_strategies}
  \centering
  \resizebox{0.9\linewidth}{!}{
  \begin{tabular}{lccc}
    \toprule

    Method & Strategy & \#Instance & MPE$\downarrow$\\
    
    \midrule
    \name-H32 & balanced & 4.5M & 63.08 \\
    
    \name-H32 & weighted & 4.5M & 62.12 \\
    
    \name-H32 & concatenated & 5.6M & 63.32 \\
    
    \bottomrule
  \end{tabular}}
\end{table}

As for the sampling strategy, we perform the ablation study on three different strategies, including 1) Balanced: we set all the datasets to have the same length; 2) Weighted: we set the dataset length according to the individual dataset benchmark rankings. Specifically, we sort the datasets based on their rankings and then assign weights to each dataset. As a result of this weighting, the datasets are upsampled or downsampled so that the lengths of the datasets are adjusted to an arithmetic sequence. The length of the dataset with the highest ranking is 4 times that of the dataset with the lowest ranking, and the sum of the total lengths of all datasets is fixed; 3) Concatenated: we simply concatenate all the datasets with their original length.

The performance of the foundation model is not sensitive to the sampling strategy as shown in \Tab\ref{tab:sampling_strategies}, while the balanced strategy is more intuitive, easy to implement, and efficient, the weighted strategy may have more potential to improve performance but may require more effort in weight tuning.

\subsection{Training Schemes}
\begin{table}
  \caption{\textbf{Training schemes}. We study the different training schemes by comparing the model trained with the Top 5 / Top 10 datasets with the Bottom 5 / Bottom 10 datasets according to our individual dataset benchmark rankings. Unit: mm.}
  \label{tab:training_schemes}
  \centering
  \resizebox{0.9\linewidth}{!}{
  \begin{tabular}{lccc}
    \toprule

    Method & Dataset & \#Instance & MPEMPE$\downarrow$\\
    
    \midrule
    \name-B & Top5 & 0.75M & 103.47 \\
    
    \name-B & Bottom5 & 0.75M & 115.61 \\
    
    \name-B & Top10 & 1.5M & 89.20 \\
    
    \name-B & Bottom10 & 1.5M & 115.10\\
    
    \bottomrule
  \end{tabular}}
\end{table}

As shown in \Tab\ref{tab:training_schemes}, we perform the ablation study for the training scheme. We investigate the effect of dataset selection. \textcolor{blue}{We select the bottom five and bottom 10 datasets according to the rankings of our individual dataset benchmark on training sets in \Supp} and trained the \name-B model with the same number of instances as used in training with the top 5 and top 10 datasets.

It is proved that our training scheme is efficient. Selecting the top 5 or top 10 datasets according to the single dataset benchmark leads to a much better performance compared to selecting the bottom five or bottom 10 datasets. The foundation model can benefit from adding higher-ranked (i.e., Top 5/10) data into training, while lower-ranked data (i.e., Bottom 5/10) is not as effective in improving the model's performance. 

\subsection{\textcolor{blue}{Effectiveness of Tokens}}
\begin{figure}[t!]
  \centering
  \vspace{0mm}
  \includegraphics[width=\linewidth]{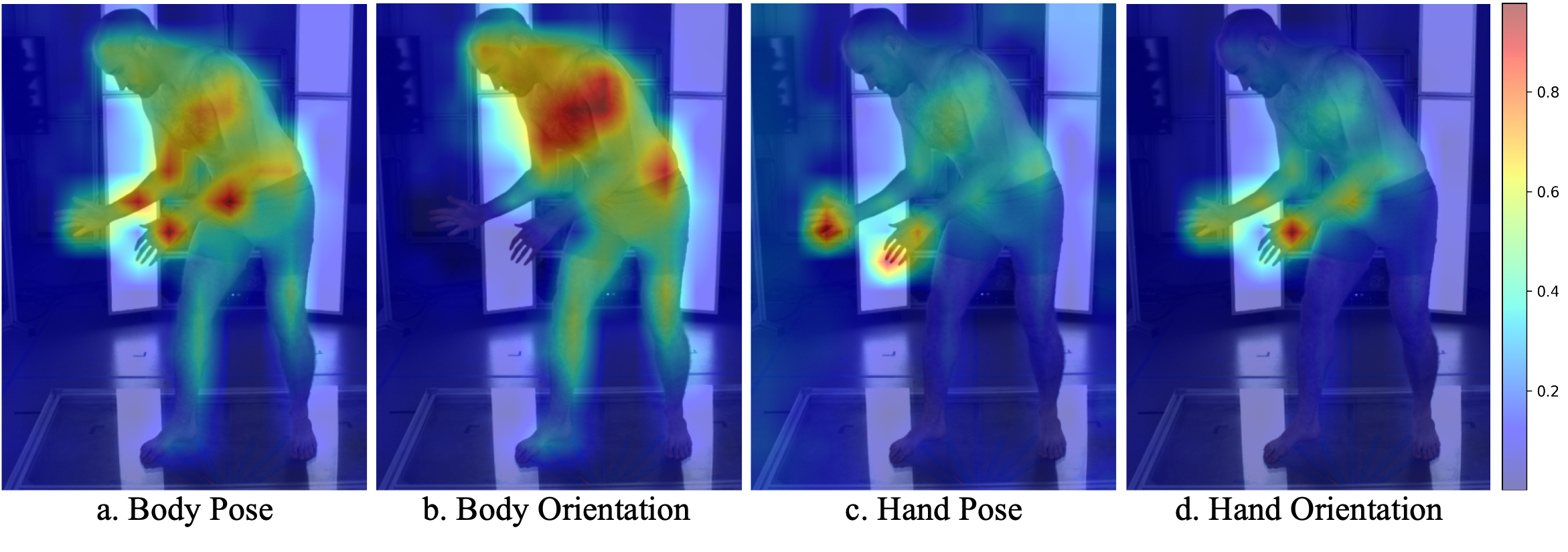}
  \caption{\textbf{Attentions of tokens.} The extended tokens in \namepami attends to the respective information in the image feature without additional component guidance.}
  \label{fig:attention_map}
\end{figure}

\textcolor{blue}{In \namepami, the component guiding module from \name is replaced with a decoder, and the body pose token is extended to include multiple specific task tokens. As illustrated in \Fig\ref{fig:attention_map}, these tokens effectively attend to their respective information in the image feature without requiring additional guidance. For instance, the body pose token activates in the torso region, body orientation tokens focus on the shoulders and pelvis, hand pose tokens concentrate on the fingers of both hands, and hand orientation tokens primarily attend to the upper body, with a particular emphasis on the wrists. The hand pose token has proved effective in capturing local hand pose (evaluated by hand-PA-MPE) while the hand orientation token shows that the upper body is essential for accurate hand orientation (evaluated by hand-MPE).}

\begin{table}[t]
  \caption{\textbf{Scaling up with CNN backbone.} We study effect of data scaling on the method with CNN backbones. Unit: mm.}
  \label{tab:scale_cnn}
  \vspace{0mm}
  \centering
  \resizebox{0.9\linewidth}{!}{
  \begin{tabular}{lccc}
    \toprule    
    Model & Sampling strategy & \#Inst. & MPE$\downarrow$ \\
    \midrule
    
    H4W~\cite{GyeongsikMoon2020hand4whole} & - & 0.65M & 116.59 \\
    H4W-32~\cite{GyeongsikMoon2020hand4whole} & balanced & 4.5M & 98.33 \\
    H4W-32~\cite{GyeongsikMoon2020hand4whole} & concatenated & 5.6M & 96.90 \\

    \bottomrule
    \end{tabular}
    }
\end{table}
\subsection{\textcolor{blue}{Scaling Up with CNN Backbones}}
\textcolor{blue}{We choose ViT architecture as our backbone since it has been a popular choice for building large models, and this is the first comprehensive investigation in scaling up both training data and model size of EHPS models. Though the transformer-based architecture cannot be trivially changed to a CNN-based counterpart (\name and \namepami requires task tokens to estimate body pose parameters), we have also trained a Hand4Whole~\cite{GyeongsikMoon2020hand4whole}, as the CNN-based foundation model (ResNet50 backbone for body and hand, ResNet18 for face with a total of 77M trainable parameters) as shown in Table~\ref{tab:scale_cnn} to verify that the conclusion still holds true for CNN-based architectures. We highlight that similar to \name and \namepami, scaling up the training data for a CNN-based foundation model also shows a substantial performance boost.}

\section{\textcolor{blue}{Discussions}}
\textcolor{blue}{
Previous algorithmic studies may have been premature, as the impact of massive data had not been fully explored or ruled out. Our study investigates the non-algorithmic aspects of EHPS through a focus on data- and model-scaling, and systematically apply extensive datasets to their fullest extent on various model sizes. In this section, we discuss the key limitations and take-aways.
}

\textcolor{blue}{
By consolidating nearly all available datasets, we observe a clear diminishing return across all model sizes: within the same data domain (40 datasets), increasing the training instances from 7.5M to 10M raises the computational cost by 43\% (from 23 hours per epoch to 33 hours per epoch), yet results in only marginal performance improvement. Hence, our results suggest that further data scaling has reached a saturation point, indicating the limitations of scaling alone in driving further advancements. With this saturation established, the field is \textit{better} poised for a necessary shift toward genuine algorithmic advancements upon sufficient pretraining with diverse data. 
}

\textcolor{blue}{
Another primary limitation of this study is the representativeness of the datasets used. While we employ five comprehensive benchmark datasets to assess whole-body generalization capabilities (with two additional datasets for hand evaluation), these may not fully capture the complexity and diversity of real-world distributions. Nonetheless, to the best of our knowledge, this work constitutes the most systematic generalization study of EHPS to date. We are confident that our observations are broadly applicable to in-the-wild scenarios and provide valuable insights for future research.
}


\vspace{-3mm}
\section{Conclusion}
\label{sec:conclusion}
\vspace{-2mm}

In this work, we benchmark datasets for EHPS that provide us insights to train and finetune a \textcolor{blue}{family of foundation models with minimalistic architectures. Our extensive experiments on data- and model-scaling have systematically explored the impact of leveraging all available datasets to their fullest extent, allowing the foundation models to outperform existing state-of-the-art approaches in accuracy and generalizability}. 

\textcolor{blue}{
Our study is thus useful in three ways. 
First, our systematic study on EHPS datasets, generalization metrics (MPE), and proposed benchmarks serve to gauge the performances of future investigations in building high-performance and robust models for in-the-wild EHPS. 
Second, \name and \namepami are intentionally designed with simplicity and modularity in mind, incorporating highly reusable components and trained on massive datasets, thereby relieving future researchers of the need to replicate this resource-intensive process. Hence, the pretrained models or their components can be a plug-and-play part of any system for EHPS or downstream tasks. 
Third, our benchmark-finetuning paradigm can be useful for the rapid adaptation of any foundation model to specific scenarios. Specifically, users may collect a training set, evaluate pretrained models of various other datasets on it, and select the most relevant datasets to finetune a foundation model. 
}



\appendices
\section{Additional Distribution Comparisons}
\label{sec:supp:add_distribution}

\begin{figure*}[t]
  \centering
  \includegraphics[width=\linewidth]{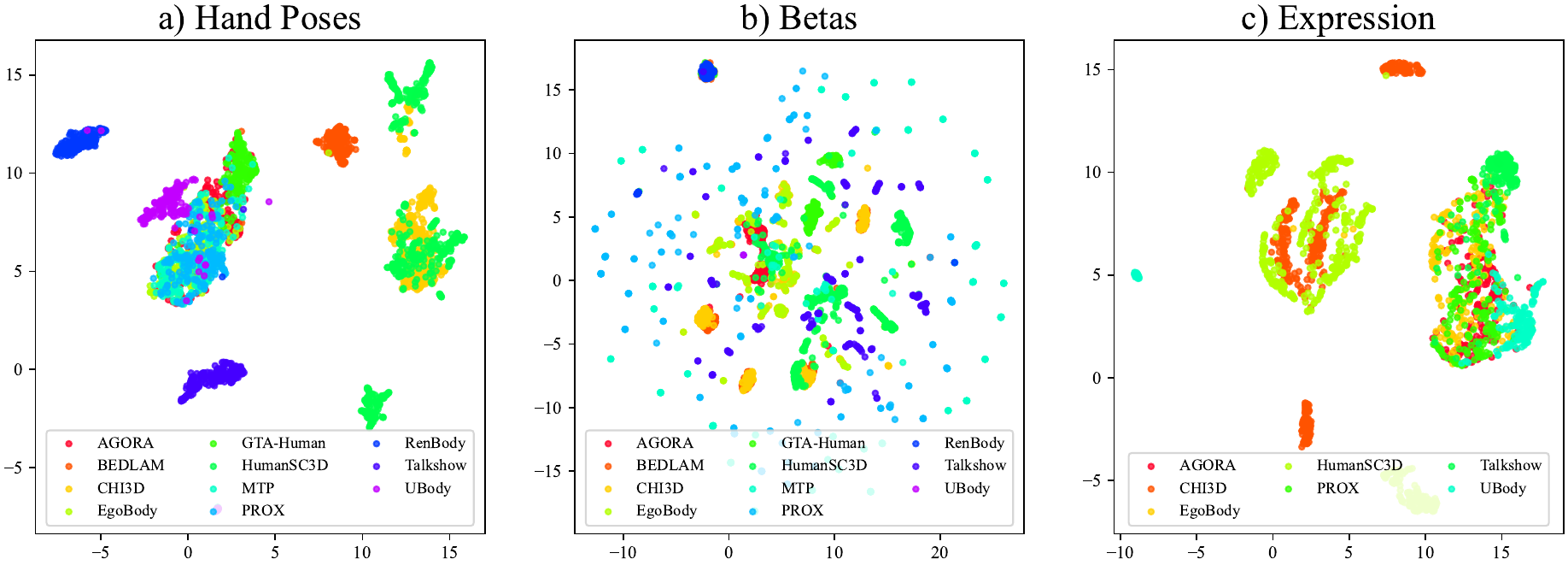}
  \caption{\textbf{Comparisons of hand pose and shape beta parameters distributions among different datasets}. We illustrate these distributions with UMAP~\cite{mcinnes2018umap}. The two axes are the two dimensions of the embedded space and have no unit.}
  \label{fig:supp_distributions}
\end{figure*}

In \Fig\ref{fig:supp_distributions}, we plot more distributions of additional parameters: a) hand poses, b) betas (body shape), and c) facial expression, all via UMAP dimension reduction. Datasets without proper SMPL-X parameters (\eg, SMPL annotation only, or pseudo-annotated that typically have invalid hand poses) are not included in the study. For hand poses, we concatenate both left and right-hand parameters in rotation matrix representation. For betas and expression, we directly use their first 10 components. It is observed that datasets such as DNA-Rendering, CHI3D, HumanSC3D, and Talkshow form distinct clusters for hand poses and betas, and it is difficult to find any dataset to provide a well-spread coverage. For expression, there is still a lack of diverse datasets. this observation holds with another analysis of hand poses in the Main Paper \Fig 3. Thus, we present the \synhand dataset.

\section{Adaption of SMPL/SMPL-X Annotations}
\label{sec:supp:add_exp}


While we strive to utilize as many datasets as possible in our study, we find that there are only a few datasets with neutral SMPL-X annotations and many datasets with female/male (gendered) SMPL-X annotations or SMPL annotations. An intuitive solution is to use the official fitting tool~\cite{Pavlakos_2019smplx}, however, this optimization-based tool is relatively slow to convert a large number of annotations (fitting takes 241$\pm$126 seconds per frame). Hence, we experiment with a new approach. 

For gendered SMPL-X annotations, we train a small adapter network $A$ (consisting of three layers of fully-connected layers) that takes in gendered body shape parameters $\beta_{f/m}$ and converts it neutral body shape parameters such that the following loss is minimized:
\begin{equation}
    \mathcal{L} = || M_{f/m}(\theta, \beta_{f/m}) - M_{n}(\theta, A(\beta_{f/m}) ||_{2}
\end{equation}
\noindent where $M_{f/m}$ are gendered SMPL-X body model, and $M_{n}$ is the neutral SMPL-X body model, $\theta$ is body pose is obtained by random sampling in the embedding space of VPoser~\cite{Pavlakos_2019smplx}. We test our adapter on AGORA~\cite{patel2021agora} and find that the vertex-to-vertex error between ground truth gendered SMPL-X mesh and neutral SMPL-X with adapted neutral $\beta$ is 8.4 mm, which we consider to be sufficiently small. This approach is very fast (0.09 seconds per frame). Hence, we apply our adapter on AGORA, EgoBody, RenBody, and RICH. 

However, we empirically find that the adapter does not work well across significantly different topologies (\ie SMPL and SMPL-X), training similar adapters results in a 27.1 mm vertex-to-vertex error. Hence, for datasets with SMPL annotations, we only supervise ground truth global orientation and body pose. Although this is a slight abuse of the parameters (SMPL and SMPL-X parameters are not directly transferable), we find in our experiments that such a strategy leads to performance gains.


\section{Complete Results}
\label{sec:supp:complete_results}

\begin{table*}[ht]
  \begin{minipage}{0.48\linewidth}
    \centering
    \caption{\textbf{Full results on AGORA Val set}. 
    }
    \label{tab:full_agora_val}
    \vspace{-2mm}
    \resizebox{\textwidth}{!}{
    \begin{tabular}{lcccccc}
    \toprule
    & 
    \multicolumn{3}{c}{PA-PVE$\downarrow$ (mm)} &
    \multicolumn{3}{c}{PVE$\downarrow$ (mm)} \\
    
    \cmidrule(lr){2-4} \cmidrule(lr){5-7} 
    
    Method &
    All & Hands & Face &
    All & Hands & Face \\
    
    \midrule

    \name-S5 & 72.1 & 10.2 & 5.1 & 119.0 & 66.8 & 58.9 \\
    \name-S10 & 67.5 & 10.2 & 4.8 & 116.0 & 65.2 & 57.5 \\
    \name-S20 & 62.1 & 10.0 & 4.4 & 109.2 & 63.3 & 55.2 \\
    \name-S32 & 58.7 & 9.8 & 4.2 & 105.2 & 61.9 & 53.9 \\
    \name-S40 & 56.7 & 9.1 & 3.6 & 97.9 & 55.4 & 48.1 \\
    \name-S40+ & 55.8 & 9.0 & 3.6 & 94.0 & 54.6 & 46.3 \\   
    \midrule
    
    \name-B5 & 63.8 & 9.6 & 4.8 & 102.6 & 59.0 & 50.8 \\
    \name-B10 & 58.4 & 9.5 & 4.6 & 97.8 & 57.8 & 49.1 \\
    \name-B20 & 56.9 & 9.3 & 4.3 & 95.6 & 56.5 & 47.9 \\
    \name-B32 & 52.0 & 9.2 & 4.1 & 88.0 & 54.5 & 45.9 \\
    \name-B40 & 51.3 & 8.6 & 3.6 & 86.2 & 50.6 & 41.8 \\
    \name-B40+ & 50.1 & 8.5 & 3.5 & 81.9 & 48.3 & 38.8 \\
    \midrule
    
    \name-L5 & 56.1 & 9.2 & 4.3 & 88.3 & 53.0 & 43.3 \\
    \name-L10 & 50.6 & 9.1 & 4.1 & 82.6 & 51.9 & 42.3 \\
    \name-L20 & 48.6 & 8.9 & 4.0 & 80.7 & 51.0 & 41.3 \\
    \name-L32 & 45.1 & 8.7 & {3.8} & 74.2 & 47.8 & 38.7 \\
    \name-L40 & 44.7 & 8.0 & 3.2 & 69.0 & 42.8 & 32.2 \\
    \name-L40+ & 44.8 & 8.0 & 3.2 & 69.2 & 43.2 & 32.1 \\
    \midrule
    
    \name-H5 & 57.8 & 9.1 & 4.2 & 89.0 & 52.6 & 42.6 \\
    \name-H10 & 51.0 & 9.0 & 4.0 & 81.4 & 51.4 & 40.5 \\
    \name-H20 & 47.1 & 8.8 & 3.9 & 77.5 & 49.5 & 39.4 \\
    \name-H32 & {42.9} & {8.5} & {3.7} & {69.5} & {45.6} & {35.9} \\
    \name-H40 & 44.1 & 8.0 & 3.1 & 68.2 & 41.8 & 32.9 \\
    \name-H40+ & 42.0 & \underline{7.9} & 3.0 & 65.8 & 40.1 & \underline{31.2} \\
    \midrule
    
    \namepami-H5 & 52.2 & 15.2 & 3.5 & 96.3 & 57.1 & 41.1 \\
    \namepami-H10 & 49.2 & 8.3 & 3.0 & 71.4 & 41.4 & 31.3 \\
    \namepami-H20 & 48.3 & 8.9 & 3.2 & 76.4 & 45.2 & 34.7 \\
    \namepami-H32 & 45.2 & 8.2 & 3.1 & 70.6 & 43.6 & 33.8 \\
    \namepami-H40 & \underline{39.2} & \textbf{7.3} & \textbf{2.7} & \textbf{56.1} & \textbf{36.9} & \textbf{26.5} \\
    \namepami-H40+ & \textbf{37.9} & \underline{7.9} & \underline{2.8} & \underline{59.5} & \underline{38.4} & {31.3} \\
    
    \bottomrule
    \end{tabular}}
  \end{minipage}\hfill
  \begin{minipage}{0.48\linewidth}
    \centering
    \caption{\textbf{Full results on EHF}. 
    }
    \label{tab:full_ehf}
    \vspace{-2mm}
    \resizebox{\textwidth}{!}{
    \begin{tabular}{lcccccc}
    \toprule
    & 
    \multicolumn{3}{c}{PA-PVE$\downarrow$ (mm)} &
    \multicolumn{3}{c}{PVE$\downarrow$ (mm)} \\
    
    \cmidrule(lr){2-4} \cmidrule(lr){5-7} 
    
    Method &
    All & Hands & Face &
    All & Hands & Face \\
    
    \midrule

    \name-S5 & 70.7 & 16.0 & 5.9 & 100.5 & 64.0 & 27.1 \\
    \name-S10 & 60.5 & 16.0 & 5.7 & 89.9 & 59.1 & 22.3 \\
    \name-S20 & 51.0 & 15.5 & 5.5 & 86.6 & 54.7 & 22.1 \\
    \name-S32 & 50.5 & 14.8 & 5.2 & 74.1 & 54.6 & 20.0 \\
    \name-S40 & 45.8 & 13.5 & 3.7 & 69.1 & 51.6 & 23.7 \\
    \name-S40+ & 43.7 & 13.7 & 3.5 & 67.4 & 50.2 & 22.1 \\
    \midrule
    
    \name-B5 & 61.4 & 15.4 & 5.8 & 96.1 & 58.4 & 27.1 \\
    \name-B10 & 46.7 & 15.7 & 5.6 & 74.7 & 55.1 & 21.3 \\
    \name-B20 & 41.9 & 15.9 & 5.3 & 73.0 & 53.7 & 20.8 \\
    \name-B32 & 40.7 & 14.5 & 5.2 & 67.3 & 52.1 & 20.6 \\
    \name-B40 & 38.1 & 13.4 & 3.5 & 68.3 & 48.2 & 24.4 \\
    \name-B40+ & 36.3 & 13.3 & 3.3 & 64.6 & 46.3 & 22.6 \\
    \midrule
    
    \name-L5 & 53.9 & 14.7 & 5.9 & 89.5 & 57.8 & 29.9 \\
    \name-L10 & 40.7 & 15.6 & 5.3 & 64.0 & 52.9 & 18.1 \\
    \name-L20 & {37.8} & 15.0 & {5.1} & 65.4 & 49.4 & {17.4} \\
    \name-L32 & {37.1} & {14.1} & {5.0} & 62.3 & 47.1 & {17.0} \\
    \name-L40 & 32.9 & 12.8 & 3.3 & 58.6 & 43.1 & 18.5 \\
    \name-L40+ & 36.3 & 13.3 & 3.3 & 64.6 & 46.3 & 22.6 \\
    \midrule
    
    \name-H5 & 47.0 & {14.3} & 5.9 & 68.3 & 55.6 & 25.0 \\
    \name-H10 & 40.1 & 15.6 & 5.2 & {56.6} & 50.2 & 18.9 \\
    \name-H20 & {39.0} & 14.4 & {5.0} & {59.4} & 47.1 & 17.8 \\
    \name-H32 & {39.0} & 14.8 & {5.0} & {56.9} & {42.2} & 19.0 \\
    \name-H40 & 37.6 & 12.1 & 3.5 & 53.8 & 39.9 & 18.0 \\
    \name-H40+ & {37.2} & 11.9 & 3.5 & 52.3 & \underline{38.1} & 19.4 \\
    \midrule

    \namepami-H5 & 45.0 & 14.8 & 4.3 & 71.6 & 53.6 & 29.6 \\
    \namepami-H10 & 39.6 & 12.9 & \textbf{3.3} & 60.3 & 42.0 & 19.2 \\
    \namepami-H20 & 37.3 & 13.1 & 3.6 & 65.4 & 44.8 & 18.9 \\
    \namepami-H32 & \textbf{31.8} & 12.1 & \underline{3.4} & 53.0 & 41.5 & 17.9 \\
    \namepami-H40 & \underline{34.7} & \textbf{11.4} & \underline{3.4} & \textbf{49.7} & \textbf{36.4} & \textbf{16.2} \\
    \namepami-H40+ & {37.2} & \underline{11.7} & \underline{3.4} & \underline{50.5} & 38.8 & \underline{16.4} \\
        
    \bottomrule
    \end{tabular}}
  \end{minipage}
 \vspace{-0.3cm} 
\end{table*}
\begin{table*}[ht]
  \begin{minipage}{0.48\linewidth}
    \centering
    \caption{\textbf{Full results on UBody.} 
    }
    \label{tab:full_ubody}
    \resizebox{\textwidth}{!}{
    \begin{tabular}{lcccccc}
    \toprule
    & 
    \multicolumn{3}{c}{PA-PVE$\downarrow$ (mm)} &
    \multicolumn{3}{c}{PVE$\downarrow$ (mm)} \\
    
    \cmidrule(lr){2-4} \cmidrule(lr){5-7} 
    
    Method &
    All & Hands & Face &
    All & Hands & Face \\
    
    \midrule
    
    \name-S5 & 53.9 & 11.8 & 3.9 & 110.1 & 59.4 & 34.5 \\
    \name-S10 & 50.4 & 11.5 & 3.7 & 107.6 & 57.4 & 32.8 \\
    \name-S20 & 37.5 & 11.1 & 3.2 & 70.7 & 49.6 & 26.1 \\
    \name-S32 & 36.4 & 10.7 & 3.0 & 68.1 & 47.8 & 25.0 \\
    \name-S40 & 37.7 & 9.8 & 4.8 & 71.4 & 44.8 & 27.0 \\
    \name-S40+ & 36.2 & 9.6 & 4.4 & 66.9 & 43.4 & 25.4 \\
    \midrule
    
    \name-B5 & 52.3 & 11.9 & 3.8 & 105.8 & 56.9 & 32.6 \\
    \name-B10 & 49.7 & 12.0 & 3.6 & 107.3 & 57.1 & 31.7 \\
    \name-B20 & 35.5 & 11.0 & 3.0 & 65.3 & 46.9 & 23.4 \\
    \name-B32 & 33.7 & 10.8 & 2.8 & 63.3 & 43.9 & 22.7 \\
    \name-B40 & 34.7 & 9.1 & 5.2 & 64.4 & 40.8 & 24.0 \\
    \name-B40+ & 32.3 & 8.8 & 3.6 & 60.4 & 38.7 & 22.9 \\
    \midrule
    
    \name-L5 & 51.8 & 12.5 & 3.6 & 110.8 & 56.3 & 37.5 \\
    \name-L10 & 48.0 & 12.8 & 3.5 & 104.0 & 56.1 & 32.0 \\
    \name-L20 & 33.2 & 10.6 & 2.8 & 61.5 & 43.3 & 23.1 \\
    \name-L32 & 30.9 & 10.2 & 2.7 & 57.3 & 39.2 & 21.6 \\
    \name-L40 & 31.2 & 8.4 & 4.2 & 58.6 & 35.7 & 22.6 \\
    \name-L40+ & 29.5 & 8.2 & 3.0 & 54.7 & 34.2 & 20.8 \\
    \midrule
    
    \name-H5 & 48.1 & 12.1 & 3.7 & 102.1 & 53.3 & 33.4 \\
    \name-H10 & 48.5 & 12.6 & 3.5 & 100.7 & 54.8 & 30.9 \\
    \name-H20 & 32.8 & 10.3 & 2.8 & 59.9 & 41.0 & 22.7 \\
    \name-H32 & {29.9} & {9.8} & \underline{2.6} & {54.5} & {36.4} & {20.6} \\
    \name-H40 & 31.1 & \underline{8.1} & 3.7 & 56.9 & 34.7 & 24.1 \\
    \name-H40+ & 29.1 & \textbf{7.9}& 2.7 & 53.7 & \underline{33.1} & \textbf{21.0} \\
    \midrule

    \namepami-H5 & 45.6 & 13.1 & 5.7 & 95.2 & 48.9 & 32.9 \\
    \namepami-H10 & 55.1 & 11.4 & 5.8 & 100.8 & 49.0 & 31.6 \\
    \namepami-H20 & 31.6 & 9.5 & 2.8 & 59.8 & 37.7 & 23.0 \\
    \namepami-H32 & 30.1 & 9.0 & 2.7 & 56.5 & 35.3 & 22.0 \\
    \namepami-H40 & \textbf{27.8} & \textbf{7.9} & \textbf{2.5} & \textbf{51.1} & \textbf{32.9} & \underline{21.4} \\
    \namepami-H40+ & \underline{28.7} & \textbf{7.9} & \underline{2.6} & \underline{52.8} & 34.0 & 22.6 \\
    \bottomrule
    \end{tabular}}
  \end{minipage}\hfill
  \begin{minipage}{0.48\linewidth}
    \centering
    \caption{\textbf{Full results on EgoBody-EgoSet.} 
    }
    \label{tab:full_egobody}
    \resizebox{\textwidth}{!}{
    \begin{tabular}{lcccccc}
    \toprule
    & 
    \multicolumn{3}{c}{PA-PVE$\downarrow$ (mm)} &
    \multicolumn{3}{c}{PVE$\downarrow$ (mm)} \\
    
    \cmidrule(lr){2-4} \cmidrule(lr){5-7} 
    
    Method &
    All & Hands & Face &
    All & Hands & Face \\
    
    \midrule
    
    \name-S5 & 62.8 & 10.8 & 4.1 & 114.2 & 53.3 & 44.3 \\
    \name-S10 & 52.2 & 10.0 & 3.4 & 88.6 & 48.6 & 37.6 \\
    \name-S20 & 48.1 & 10.0 & 3.3 & 84.3 & 47.2 & 37.8 \\
    \name-S32 & 46.0 & 10.0 & 3.1 & 82.5 & 46.0 & 36.2 \\
    \name-S40 & 43.9 & 9.5 & 3.1 & 78.8 & 42.7 & 35.5 \\
    \name-S40+ & 43.2 & 9.4 & 3.1 & 78.0 & 42.7 & 34.8 \\
    \midrule
    
    \name-B5 & 59.4 & 10.6 & 4.0 & 108.1 & 48.0 & 40.0 \\
    \name-B10 & 45.3 & 10.1 & 3.2 & 76.4 & 45.5 & 32.4 \\
    \name-B20 & 43.8 & 9.9 & 3.2 & 75.5 & 44.6 & 32.7 \\
    \name-B32 & 40.7 & 9.9 & 3.1 & 72.7 & 43.7 & 32.4 \\
    \name-B40 & 38.6 & 9.2 & 3.0 & 70.1 & 39.0 & 30.4 \\
    \name-B40+ & 38.4 & 9.3 & 2.9 & 68.2 & 38.8 & 30.9 \\
    \midrule
    
    \name-L5 & 52.9 & 10.5 & 3.8 & 98.7 & 45.2 & 39.1 \\
    \name-L10 & 40.5 & 10.0 & 3.0 & 69.7 & 43.1 & 32.0 \\
    \name-L20 & 38.9 & 9.9 & 3.0 & 66.6 & 42.7 & 31.8 \\
    \name-L32 & 36.3 & {9.8} & 2.9 & 62.2 & 41.4 & 30.7 \\
    \name-L40 & 35.9 & 8.9 & 2.8 & 60.7 & 36.5 & 29.8 \\
    \name-L40+ & 35.2 & 9.0 & 2.8 & 59.1 & 36.5 & 28.8 \\
    \midrule
    
    \name-H5 & 48.0 & 10.5 & 3.4 & 87.4 & 43.5 & 37.5 \\
    \name-H10 & 38.8 & 10.0 & 2.9 & 65.7 & 42.6 & 31.1 \\
    \name-H20 & 36.7 & {9.8} & 2.9 & 63.5 & 41.3 & 30.8 \\
    \name-H32 & {34.3} & {9.8} & {2.7} & \underline{59.5} & {39.6} & {28.7} \\
    \name-H40 & 34.1 & \underline{8.9}& \underline{2.6} & 59.9 & \underline{34.8} & 30.4 \\
    \name-H40+ & \textbf{33.1} & \textbf{8.8} & \underline{2.6} & \textbf{57.0} & \textbf{34.7} & \underline{28.3} \\
    \midrule

    \namepami-H5 & 46.4 & 10.7 & 3.7 & 92.3 & 44.4 & 38.1 \\
    \namepami-H10 & 43.3 & 9.2 & 2.7 & 68.7 & 35.9 & 29.9 \\
    \namepami-H20 & 36.7 & 9.2 & 2.8 & 63.8 & 37.0 & 30.8 \\
    \namepami-H32 & 34.8 & 9.0 & \underline{2.6} & \textbf{59.8} & 35.7 & 28.8 \\
    \namepami-H40 & \underline{33.6} & \textbf{8.8} & \textbf{2.5} & {61.5} & 35.6 & 28.5 \\
    \namepami-H40+ & 34.5 & 9.0 & \underline{2.6} & {62.0} & 36.3 & \textbf{28.1} \\
    
    \bottomrule
    \end{tabular}}
  \end{minipage}
\end{table*}
\begin{table*}[ht]
  \begin{minipage}{0.48\linewidth}
    \centering
    \caption{\textbf{Full results on ARCTIC.} 
    }
    \label{tab:full_arctic}
    \vspace{-2mm}
    \resizebox{\textwidth}{!}{
    \begin{tabular}{lcccccc}
    \toprule
    & 
    \multicolumn{3}{c}{PA-PVE$\downarrow$ (mm)} &
    \multicolumn{3}{c}{PVE$\downarrow$ (mm)} \\
    
    \cmidrule(lr){2-4} \cmidrule(lr){5-7} 
    
    Method &
    All & Hands & Face &
    All & Hands & Face \\
    
    \midrule

    \name-S5 & 66.1 & \textbf{16.7} & 4.0 & 117.3 & 58.7 & 46.5 \\
    \name-S10 & 58.8 & 17.5 & 3.2 & 104.6 & 56.6 & 41.1 \\
    \name-S20 & 37.6 & 18.9 & 2.7 & 58.7 & 45.2 & 30.5 \\
    \name-S32 & 34.5 & 18.9 & 2.7 & 55.3 & 42.9 & 28.9 \\
    \name-S40 & 28.8 & 19.2 & 3.1 & 47.3 & 39.8 & 28.4 \\
    \name-S40+ & 29.7 & 19.3 & 3.1 & 50.4 & 41.1 & 31.7 \\
    \midrule
    
    \name-B5 & 66.3 & \underline{16.9} & 3.4 & 105.4 & 55.6 & 41.4 \\
    \name-B10 & 54.0 & 17.9 & 2.5 & 85.2 & 53.4 & 35.0 \\
    \name-B20 & 34.9 & 18.9 & 2.7 & 56.3 & 40.9 & 29.6 \\
    \name-B32 & 31.9 & 19.0 & 2.8 & 52.6 & 40.1 & 27.4 \\
    \name-B40 & 26.9 & 19.2 & 3.1 & 46.3 & 37.9 & 28.7 \\
    \name-B40+ & 27.5 & 18.9 & 3.1 & 47.5 & 38.1 & 31.3 \\
    \midrule
    
    \name-L5 & 57.2 & 17.0 & 2.9 & 95.1 & 52.8 & 37.7 \\
    \name-L10 & 46.9 & 18.1 & \underline{2.3} & 76.9 & 50.8 & 33.2 \\
    \name-L20 & 31.9 & 18.9 & 2.5 & 52.2 & 39.3 & 27.0 \\
    \name-L32 & 29.4 & 18.9 & 2.7 & 48.6 & 38.8 & 26.8 \\
    \name-L40 & 24.9 & 19.2 & 3.0 & 42.6 & 36.1 & 27.2 \\
    \name-L40+ & 25.8 & 19.0 & 3.2 & 45.5 & 36.6 & 30.3 \\
    \midrule
    
    \name-H5 & 49.3 & 17.4 & 2.5 & 79.9 & 49.3 & 33.9 \\
    \name-H10 & 41.4 & 18.8 & \textbf{2.1} & 71.6 & 49.3 & 30.9 \\
    \name-H20 & {29.3} & 18.9 & 2.5 & {48.5} & {38.3} & {26.3} \\
    \name-H32 &  27.7 & 18.8 & 2.6 & 44.7 & 37.0 & \underline{24.7} \\
    \name-H40 & \underline{24.0} & 18.9 & 2.9 & 40.5 & 35.0 & 25.1 \\
    \name-H40+ & 25.2 & 19.1 & 3.0 & 42.1 & 36.3 & 28.0 \\
    \midrule

    \namepami-H5 & 37.2 & 13.2 & 4.1 & 73.9 & 39.6 & 37.9 \\
    \namepami-H10 & 43.1 & 13.2 & 4.2 & 84.8 & 37.6 & 38.0 \\
    \namepami-H20 & 26.3 & 11.9 & 3.0 & 43.5 & 28.8 & \textbf{24.1} \\
    \namepami-H32 & 26.3 & 11.6 & 3.3 & 42.6 & 26.5 & 27.3 \\
    \namepami-H40 & 24.2 & \textbf{10.9} & 3.3 & \underline{40.3} & \underline{24.1} & 28.0 \\
    \namepami-H40+ & \textbf{23.9} & \underline{11.1} & 3.3 & \textbf{39.3} & \textbf{23.6} & 26.9 \\
    
    \bottomrule
    \end{tabular}}
  \end{minipage} \hfill
  \begin{minipage}{0.48\linewidth}
    \centering
    \caption{\textbf{Full results on DNA-Rendering-HiRes}. 
    }
    \label{tab:full_renbody}
    \vspace{-2mm}
    \resizebox{\textwidth}{!}{
    \begin{tabular}{lcccccc}
    \toprule
    & 
    \multicolumn{3}{c}{PA-PVE$\downarrow$ (mm)} &
    \multicolumn{3}{c}{PVE$\downarrow$ (mm)} \\
    
    \cmidrule(lr){2-4} \cmidrule(lr){5-7} 
    
    Method &
    All & Hands & Face &
    All & Hands & Face \\
    
    \midrule
    


    \name-S5 & 70.9 & 10.4 & 4.7 & 104.9 & 57.6 & 49.7 \\
    \name-S10 & 63.9 & 11.0 & 4.4 & 98.4 & 57.0 & 47.3 \\
    \name-S20 & 55.6 & 10.2 & 4.4 & 87.3 & 53.3 & 46.2 \\
    \name-S32 & 47.1 & 7.7 & 3.5 & 70.1 & 46.9 & 39.0 \\
    \name-S40 & 46.5 & 7.9 & 3.5 & 71.8 & 45.5 & 38.4 \\
    \name-S40+ & 45.4 & 7.8 & 3.5 & 71.1 & 44.7 & 37.6 \\
    \midrule
    
    \name-B5 & 59.9 & 10.5 & 4.3 & 91.1 & 50.5 & 44.6 \\
    \name-B10 & 53.3 & 11.5 & 4.1 & 83.7 & 50.9 & 42.4 \\
    \name-B20 & 50.7 & 11.7 & 4.2 & 83.3 & 50.9 & 43.5 \\
    \name-B32 & 40.9 & 7.4 & 3.4 & 61.9 & 40.5 & 36.6 \\
    \name-B40 & 41.0 & 7.6 & 3.5 & 64.7 & 40.0 & 36.2 \\
    \name-B40+ & 39.1 & 7.5 & 3.4 & 62.8 & 38.8 & 35.5 \\
    \midrule
    
    \name-L5 & 52.4 & 10.3 & 4.0 & 85.9 & 47.6 & 44.5 \\
    \name-L10 & 47.0 & 11.2 & 3.8 & 76.2 & 47.8 & 41.7 \\
    \name-L20 & 44.4 & 11.1 & 4.5 & 77.7 & 47.5 & 43.2 \\
    \name-L32 & 35.8 & {7.2} & {3.2} & 54.4 & 36.7 & 34.0 \\
    \name-L40 & 35.8 & 7.3 & 3.2 & 57.3 & 35.8 & 33.5 \\
    \name-L40+ & 34.7 & 7.3 & 3.2 & 56.5 & 34.9 & 32.6 \\
    \midrule
    
    \name-H5 & 53.9 & 10.3 & 3.9 & 81.9 & 46.3 & 40.7 \\
    \name-H10 & 47.4 & 10.9 & 3.7 & 76.2 & 47.0 & 39.0 \\
    \name-H20 & 43.0 & 11.2 & 3.8 & 72.8 & 45.6 & 40.5 \\
    \name-H32 & {34.0} & \textbf{7.1} & \textbf{3.1} & \textbf{51.4} & \underline{34.5} & {32.0} \\
    \name-H40 & 35.0 & 7.3 & \underline{3.2} & 58.2 & 35.3 & 33.4 \\
    \name-H40+ & \underline{33.8} & \underline{7.2} & \underline{3.2} & \underline{55.3} & \textbf{33.8} & \underline{31.8} \\
    \midrule

    \namepami-H5 & 46.0 & 10.4 & 3.7 & 79.0 & 46.8 & 41.7 \\
    \namepami-H10 & 50.3 & 9.8 & 3.6 & 78.4 & 44.4 & 37.8 \\
    \namepami-H20 & 43.6 & 10.5 & 3.8 & 69.7 & 47.2 & 39.1 \\
    \namepami-H32 & 35.1 & 7.4 & \underline{3.2} & 55.4 & 35.5 & 32.1 \\
    \namepami-H40 & \textbf{33.6} & 7.4 & \textbf{3.1} & 56.2 & 34.6 & 32.0 \\
    \namepami-H40+ & \underline{33.8} & 7.5 & \textbf{3.1} & 56.4 & 35.2 & \textbf{31.5} \\

    \bottomrule
    \end{tabular}}
  \end{minipage}
\end{table*}

\begin{table}[ht]
  \centering
  \caption{\textbf{Full results on 3DPW.} 
  Only whole-body (SMPL-X) methods are listed.}
  \label{tab:full_3dpw}
  \vspace{0mm}
  \centering
  \resizebox{0.9\linewidth}{!}{
  \begin{tabular}{lcc}
    \toprule    
    Method & MPJPE$\downarrow$(mm) & PA-MPJPE$\downarrow$(mm)  \\

    \midrule

    \name-S5 & 110.2 & 79.1 \\
    \name-S10 & 97.4 & 69.0 \\
    \name-S20 & 87.4 & 60.0 \\
    \name-S32 & 83.2 & 57.1 \\
    \name-S40 & 80.2 & 54.4 \\
    \name-S40+ & 80.3 & 53.4 \\
    \midrule
    
    \name-B5 & 104.8 & 72.0 \\
    \name-B10 & 89.9 & 62.7 \\
    \name-B20 & 83.5 & 57.6 \\
    \name-B32 & 80.3 & 53.4 \\
    \name-B40 & 77.2 & 49.0 \\
    \name-B40+ & 74.9 & 48.3 \\
    \midrule
    
    \name-L5 & 97.8 & 62.6 \\
    \name-L10 & 82.5 & 56.0 \\
    \name-L20 & 78.3 & 52.1 \\
    \name-L32 & 75.2 & {50.5}\\
    \name-L40 & 71.5 & 45.8 \\
    \name-L40+ & 71.1 & 46.1 \\
    \midrule
    
    \name-H5 & 88.3 & 60.3 \\
    \name-H10 & 78.7 & 54.8 \\
    \name-H20 & {74.4} & 50.9 \\
    \name-H32 & 75.0 & 50.6 \\
    \name-H40 & \underline{70.9} & \underline{45.0} \\
    \name-H40+ & \textbf{70.5} & 45.4 \\
    \midrule

    \namepami-H5 & 82.0 & 49.7 \\
    \namepami-H10 & 78.2 & 47.6 \\
    \namepami-H20 & \textbf{70.5} & \textbf{43.2} \\
    \namepami-H32 & 75.8 & 45.8 \\
    \namepami-H40 & 76.0 & 46.5 \\
    \namepami-H40+ & 74.8 & 45.8 \\
    
    \bottomrule
  \end{tabular}
  }
\end{table}
\begin{table*}[ht]
  \vspace{-2mm}
  \caption{\textbf{Selection of training datasets by ranking on the training set of key benchmarks.} For each dataset, we evaluate a model trained on the training set and on the \textit{training} sets of four major benchmarks: AGORA, UBody, EgoBody (EgoSet), and 3DPW. Datasets are then ranked by MPE. $\bigstar$: ranking on MPE. Top 1 values on each benchmark are bolded, and the rest of Top-5 are underlined. Unit: mm.}
  \label{tab:single_datasets_eval_on_train}
  \centering
  \vspace{-2mm}
  \resizebox{0.8\textwidth}{!}{
  \begin{tabular}{l|c|ccccc}
    \toprule
    
    Dataset & 
    \textbf{MPE$\downarrow$} & 
    AGORA~\cite{patel2021agora}$\downarrow$ & 
    UBody~\cite{lin2023one}$\downarrow$ & 
    EgoBody~\cite{zhang2022egobody}$\downarrow$ & 
    3DPW~\cite{von2018recovering}$\downarrow$\\
    \midrule
    
    BEDLAM~\cite{black2023bedlam} & \textbf{124.7} & \underline{167.8} & \underline{126.7} & \underline{106.3} & \textbf{98.1} \\
    AGORA~\cite{patel2021agora} & \underline{129.9} & \textbf{131.7} & \underline{124.4} & 134.2 & 131.2 \\
    GTA-Human~\cite{cai2024playing} & \underline{135.1} & \underline{164.2} & 137.6 & 135.2 & \underline{103.5} \\
    SynBody~\cite{yang2023synbody} & \underline{138.6} & \underline{172.3} & 146.0 & \underline{129.7} & \underline{106.3} \\
    InstaVariety~\cite{kanazawa2019learning} & \underline{139.6} & 198.2 & 128.4 & 131.6 & \underline{100.6} \\
    
    \midrule
    MSCOCO~\cite{lin2014microsoft} & 139.7 & 196.8 & \underline{110.4} & \underline{130.5} & 121.1 \\
    SPEC~\cite{kocabas2021spec} & 150.0 & \underline{166.2} & 138.8 & 155.4 & 139.7 \\
    EgoBody-MVSet~\cite{zhang2022egobody} & 151.8 & 193.3 & 194.7 & \underline{119.7} & \underline{99.3} \\
    MPII~\cite{andriluka14cvpr} & 152.0 & 205.5 & 127.3 & 143.3 & 131.9 \\
    RICH~\cite{huang2022capturing} & 155.7 & 198.9 & 171.8 & 136.9 & 115.2 \\
    
    \midrule
    Egobody-EgoSet~\cite{zhang2022egobody} & 157.1 & 213.6 & \underline{123.5} & \textbf{63.6} & 134.1 \\
    CrowdPose~\cite{li2019crowdpose} & 162.3 & 213.0 & 133.7 & 146.2 & 156.3 \\
    MuCo-3DHP~\cite{Mehta2018SingleShotM3} & 163.4 & 193.2 & 189.7 & 151.1 & 119.7 \\
    UBody~\cite{lin2023one} & 166.6 & 212.9 & \textbf{61.5} & 137.6 & 149.2 \\
    PROX~\cite{hassan2019resolving} & 167.3 & 205.1 & 186.8 & 145.2 & 132.1 \\
    MPI-INF-3DHP~\cite{mehta2017monocular} & 167.5 & 221.3 & 167.4 & 150.0 & 131.4 \\
    PoseTrack~\cite{andriluka2018posetrack} & 177.0 & 219.2 & 165.4 & 173.2 & 150.2 \\
    BEHAVE~\cite{bhatnagar2022behave} & 179.0 & 204.8 & 212.3 & 167.2 & 131.8 \\
    HumanSC3D~\cite{fieraru2021learning} & 184.8 & 213.8 & 237.7 & 174.8 & 112.9 \\
    CHI3D~\cite{fieraru2020three} & 192.3 & 209.2 & 256.7 & 180.7 & 122.5 \\
    
    \midrule
    Human3.6M~\cite{ionescu2013human3} & 207.4 & 224.5 & 282.4 & 210.7 & 112.1 \\
    DNA-R-HiRes~\cite{renbody} & 207.5 & 231.1 & 275.4 & 189.4 & 134.0 \\
    ARCTIC~\cite{fan2023arctic} & 222.5 & 303.6 & 205.9 & 177.3 & 203.2 \\
    Talkshow~\cite{yi2023generating} & 225.3 & 290.0 & 132.2 & 188.1 & 290.8 \\
    UP3D~\cite{lassner2017unite} & 226.0 & 257.4 & 226.8 & 208.4 & 211.6 \\
    3DPW~\cite{von2018recovering} & 230.6 & 231.3 & 266.0 & 194.5 & 140.6 \\
    DNA-R~\cite{renbody} & 253.2 & 288.7 & 342.5 & 234.4 & 147.2 \\
    MTP~\cite{muller2021self} & 270.5 & 272.8 & 284.8 & 259.2 & 265.4 \\
    FIT3D~\cite{fieraru2021aifit} & 272.9 & 323.5 & 392.8 & 242.7 & 132.5 \\
    OCHuman~\cite{zhang2019pose2seg} & 282.3 & 307.7 & 266.7 & 261.5 & 293.4 \\
    LSPET~\cite{Johnson2011LearningEH} & 330.2 & 361.6 & 301.8 & 317.3 & 340.2 \\
    SSP3D~\cite{Sengupta2020SyntheticTF} & 512.0 & 545.9 & 533.4 & 529.7 & 439.1 \\
            
    \bottomrule
  \end{tabular}}
   \vspace{-5mm}
\end{table*}

\subsection{Benchmarking EHPS Datasets on Training Sets}
In the main paper, we benchmark individual datasets on the testing sets of the key EHPS evaluation benchmarks. However, this dataset benchmark is unsuitable for selecting top datasets for training EHPS, as the ranking leaks information about the testing sets to some extent. Hence, we construct a new benchmark that ranks EHPS datasets on the training set of AGORA, UBody, EgoBody, and 3DPW (EHF is omitted as it does not have a training set) in \Tab\ref{tab:single_datasets_eval_on_train}.

\subsection{Complete Results of Foundation Models on Evaluation Benchmarks}
We show complete results including our strongest foundation model \name-H32 on AGORA validation set (\Tab\ref{tab:full_agora_val}), UBody (\Tab\ref{tab:full_ubody}), EgoBody-EgoSet (\Tab\ref{tab:full_egobody}), EHF (\Tab\ref{tab:full_ehf}), ARCTIC (\Tab\ref{tab:full_arctic}) and RenBody-HiRes (\Tab\ref{tab:full_renbody}). 


\section{Additional Details of Datasets}
\label{sec:supp:datasets}

\subsection{Dataset Descriptions}

This section describes the \numdatasets datasets we use in our study. Note that all these are public academic datasets, each holding a license. We follow the common practice of using them in our non-commercial research and refer readers to their homepages or papers for more details regarding \emph{licenses} and their policies to ensure personal information protection.

\begin{figure}[t]
  \centering
  \includegraphics[width=\linewidth]{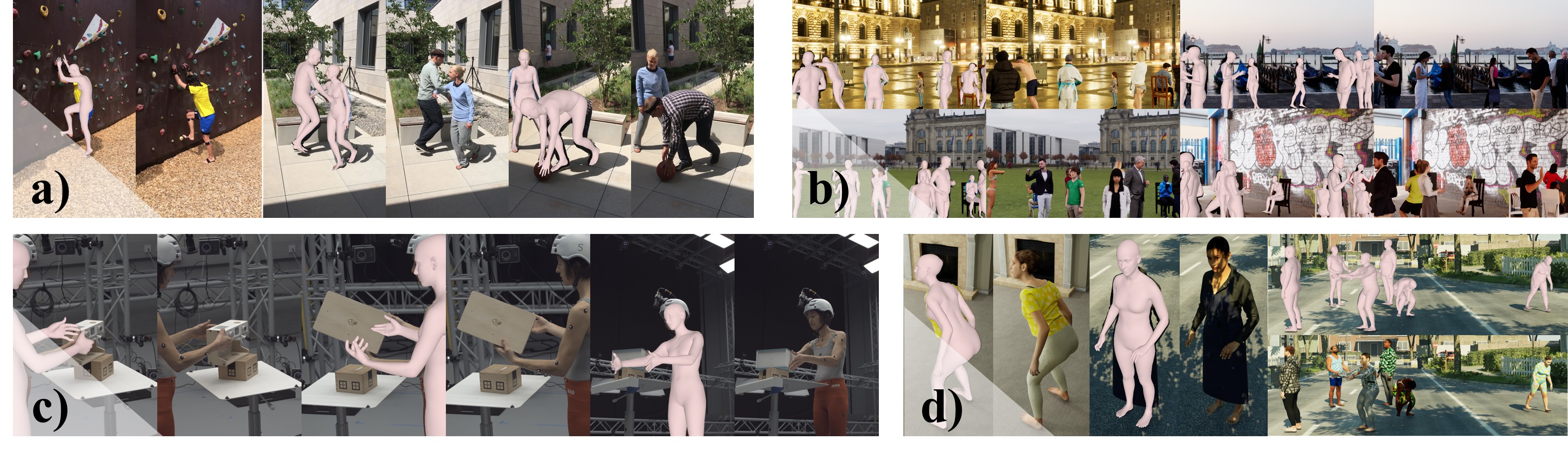}
  \caption{\textbf{Visualization} of dataset images and ground truth annotation. a) 3DPW. b) AGORA. c) ARCTIC. d) BEDLAM.}
  \label{fig:data_3dpw_agora_arctic_bedlam}
\end{figure}

\textbf{3DPW}~\cite{von2018recovering} (\Fig\ref{fig:data_3dpw_agora_arctic_bedlam}a) is the first in-the-wild dataset with considerable data, captured with a moving phone camera and IMU sensors. It features accurate SMPL annotations and 60 video sequences captured in diverse environments. We follow the official definition of train, val, and test splits. Homepage: \url{https://virtualhumans.mpi-inf.mpg.de/3DPW/}.

\textbf{AGORA}~\cite{patel2021agora} (\Fig\ref{fig:data_3dpw_agora_arctic_bedlam}b) is a synthetic dataset, rendered with high-quality human scans and realistic 3D scenes. It consists of 4240 textured human scans with diverse poses and appearances, each fitted with accurate SMPL-X annotations. There are a total of 14K training images and 3K test images, and 173K instances. Homepage and license: \url{https://agora.is.tue.mpg.de/index.html}.

\textbf{ARCTIC}~\cite{fan2023arctic} (\Fig\ref{fig:data_3dpw_agora_arctic_bedlam}c) is a lab-based hand-object interaction dataset. It features 10 subjects manipulating 11 objects. There are 210K frames of video sequences captured from 8 static cameras and 1 egocentric camera. Each frame is fitted with accurate SMPL-X annotations. We exclude the egocentric frames in our training as they only capture hands, and use 153.9K images in training. Homepage and licenses: \url{https://arctic.is.tue.mpg.de/}.

\textbf{BEDLAM}~\cite{black2023bedlam} (\Fig\ref{fig:data_3dpw_agora_arctic_bedlam}d) is a synthetic dataset that includes a wide range of variations in terms of body shapes, motions, skin tones, hair, and clothing. It is created by combining 271 different body models, 27 hairstyles, and 111 types of clothing. The dataset includes 1691 clothing textures and 2311 human motions set in 95 HDRI and 8 3D scenes. Each scene typically consists of 1 to 10 people and offers diverse camera poses. To effectively train our model, we apply a downsampling technique to reduce the size of the original data from its initial scale to  \textasciitilde190.2K instances.
Homepage and license:
\url{https://bedlam.is.tue.mpg.de/index.html}.

\begin{figure}[t]
  \centering
  \includegraphics[width=\linewidth]{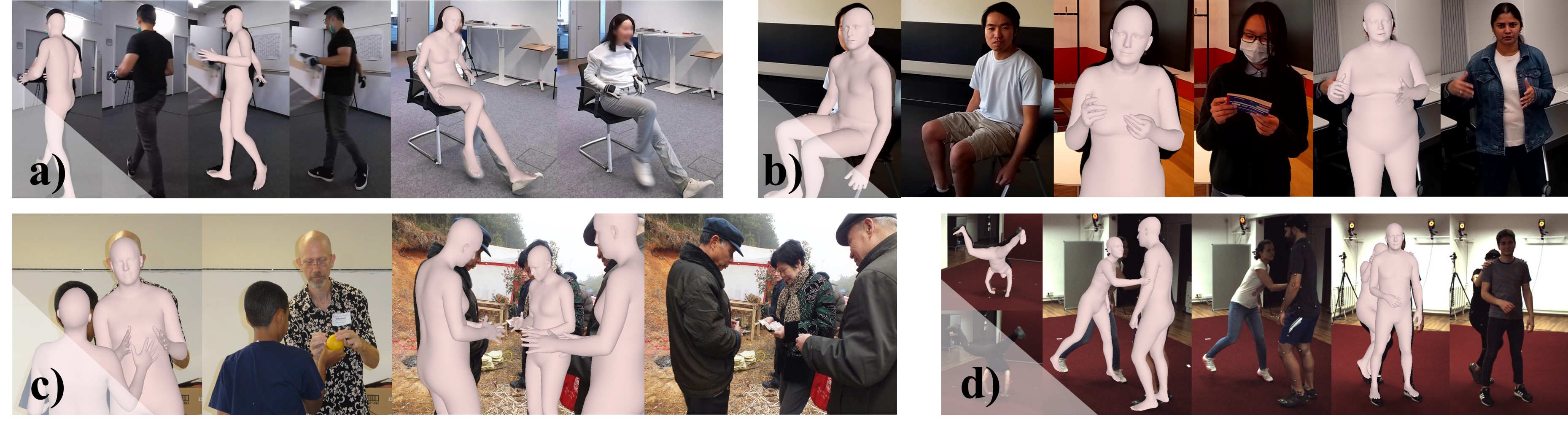}
  \caption{\textbf{Visualization} of dataset images and ground truth annotation. a) BEHAVE. b) EgoBody (EgoSet). c) CrowdPose. d) CHI3D.}
  \label{fig:data_behave_egobody_crowdpose_chi3d}
\end{figure}

\textbf{BEHAVE}~\cite{bhatnagar22behave} (\Fig\ref{fig:data_behave_egobody_crowdpose_chi3d}a) is a body human-object interaction dataset with multi-view RGBD frames and corresponding 3D SMPL-H and object fits along with the annotated contacts information. BEHAVE includes about 15k frames at 5 locations with 8 subjects performing a range of interactions with 20 common objects. Homepage: \url{https://github.com/xiexh20/behave-dataset}.

\textbf{CHI3D}~\cite{fieraru2020three} (\Fig\ref{fig:data_behave_egobody_crowdpose_chi3d}d)  is a studio-based 3d motion capture dataset (Vicon) under multiple interaction scenarios, which includes 631 multi-view sequences with 2,525 contact events and 728,664 ground truth instances of 3d poses annotated with SMPL-X parameters. We use the open-source train set. Homepage and license: \url{https://ci3d.imar.ro}.

\textbf{CrowdPose}~\cite{li2019crowdpose} (\Fig\ref{fig:data_behave_egobody_crowdpose_chi3d}c) is an in-the-wild dataset focused on crowded cases. It contains 20K images in total and 80K human instances. In this paper, we use the annotations generated by NeuralAnnot~\cite{moon2022neuralannot}, which fits the SMPL to the GT 2D joints and encompasses a total of  \textasciitilde35.7K annotated data.
Homepage: \url{https://github.com/Jeff-sjtu/CrowdPose}.

\begin{figure}[t]
  \centering
  \includegraphics[width=\linewidth]{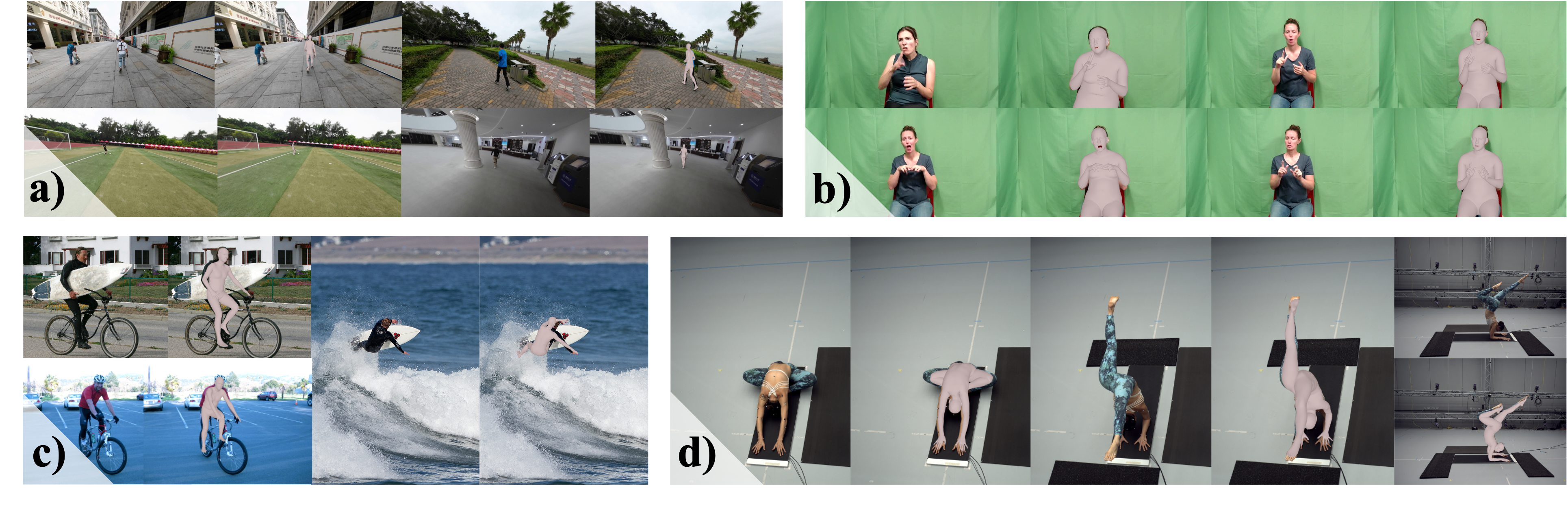}
  \caption{\textbf{Visualization} of dataset images and ground truth annotation. a) SLOPER4D. b) SignAvatars. c) DAMON. d) MoYo.}
  \label{fig:data_sloper_signavatars_damon_moyo}
\end{figure}
\begin{figure}[t]
  \centering
  \includegraphics[width=\linewidth]{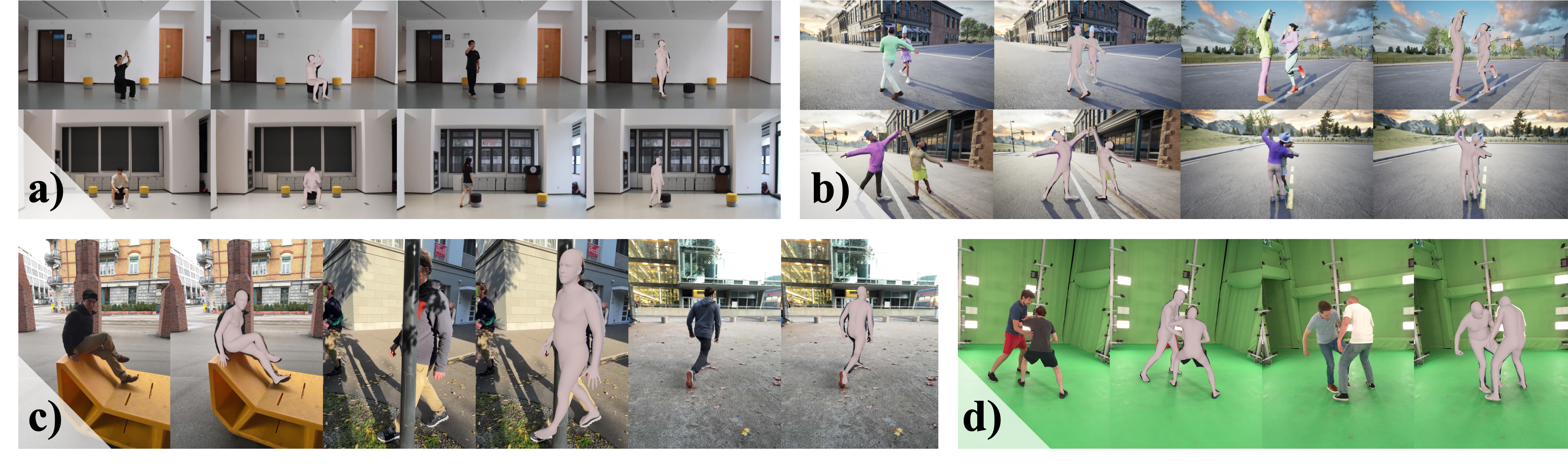}
  \caption{\textbf{Visualization} of dataset images and ground truth annotation. a) IDEA400. b) WHAC-A-Mole. c) EMDB. d) Hi4D.}
  \label{fig:data_idea_whac_emdb_hi4d}
\end{figure}

\textcolor{blue}{
\textbf{DAMON~\cite{tripathi2023deco}}
(\Fig\ref{fig:data_sloper_signavatars_damon_moyo}c) is a dataset offering SMPL-based vertex-level 3D contact annotations for 5,522 in-the-wild RGB images. Annotations include contact between 84 objects and 24 body parts. Homepage:\url{https://deco.is.tue.mpg.de/}.
}

\begin{figure}[t]
  \centering
  \includegraphics[width=\linewidth]{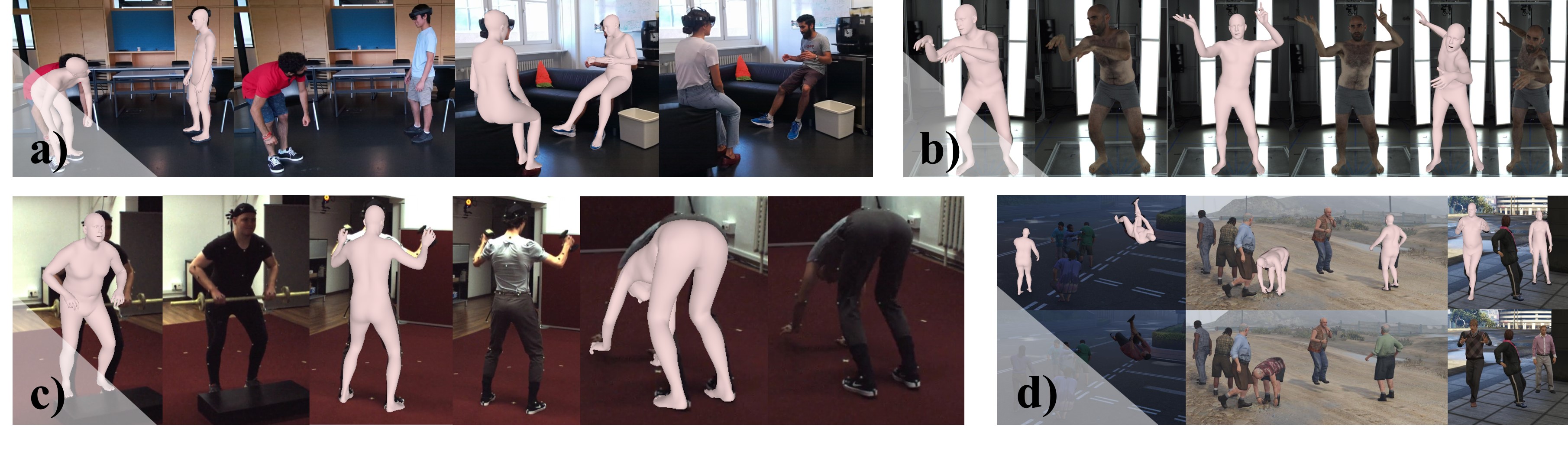}
  \caption{\textbf{Visualization} of dataset images and ground truth annotation. a) EgoBody (MVSet). b) EHF. c) FIT3D. d) GTA-Human.}
  \label{fig:data_egobodykinect_ehf_fit3d_gtahuman}
\end{figure}

\textbf{EgoBody}~\cite{zhang2022egobody} (\Fig\ref{fig:data_behave_egobody_crowdpose_chi3d}b and \Fig\ref{fig:data_egobodykinect_ehf_fit3d_gtahuman}a) is a large-scale dataset that features 3D human motions and interaction with scenes. The data is captured by a multi-view rig for third-person view in (MVSet, \Fig\ref{fig:data_egobodykinect_ehf_fit3d_gtahuman}a and a head-mounted device for egocentric view (EgoSet, in \Fig\ref{fig:data_behave_egobody_crowdpose_chi3d}b). The dataset consists of 125 sequences, 36 subjects, and 15 indoor scenes. We follow the official splits of training and test sets. Homepage: \url{https://sanweiliti.github.io/egobody/egobody.html}.

\textbf{EHF}~\cite{Pavlakos_2019smplx} (\Fig\ref{fig:data_egobodykinect_ehf_fit3d_gtahuman}b) contains 100 curated frames of one subject in an indoor studio setup. It provides SMPL-X aligned 3D mesh as the ground truth that accurately reflects the subject' diverse body, hand, and face articulations. It is usually used as a test set. The images are captured from a single camera. It is published along with SMPL-X. Homepage: \url{https://smpl-x.is.tue.mpg.de/index.html}.

\textcolor{blue}{
\textbf{EMDB}~\cite{kaufmann2023emdb}
(\Fig\ref{fig:data_idea_whac_emdb_hi4d}c) provides world-grounded SMPL annotations, body trajectories, and camera parameters for in-the-wild videos. The dataset captures 58 minutes of motion data across 81 scenes with 10 participants, using wireless electromagnetic (EM) sensors and iPhones. Homepage: \url{https://eth-ait.github.io/emdb/}.
}

\textbf{FIT3D}~\cite{fieraru2021aifit} (\Fig\ref{fig:data_egobodykinect_ehf_fit3d_gtahuman}c) is a studio-based 3d motion capture dataset including 611 multi-view sequences with 2,964,236 images and corresponding ground truth instances of 3d shapes and poses annotated with SMPL-X parameters. Motion clips include 37 repeated exercises. We use the open-source train set. Homepage and license: \url{https://fit3d.imar.ro/}.

\textbf{GTA-Human II} (\Fig\ref{fig:data_egobodykinect_ehf_fit3d_gtahuman}d) is an extended version of GTA-Human~\cite{GTAHuman}, a large-scale synthetic 3D single-human dataset generated with the GTA-V game engine, which features diversity. GTA-Human provides more than 1.4M of SMPL annotations in single-person scenes. In comparison, GTA-Human II includes multi-human scenarios with SMPL-X ground truth, obtained though SMPLify-X\cite{pavlakos2019expressive}, which estimates SMPL-X parameters from ground truth keypoints collected in-game. The toolchain is provided by MMHuman3D\cite{mmhuman3d}. The extended version contains 1.8M SMPL-X instances. Images are captured in 4K multi-person sequences, with about 600 subjects in different shapes and clothing, performing 20K daily human activity motion clips in six distinct categories of backgrounds, captured by camera angles in realistic distributions. Homepage: \url{https://caizhongang.github.io/projects/GTA-Human/}.

\begin{figure}[t]
  \centering
  \includegraphics[width=\linewidth]{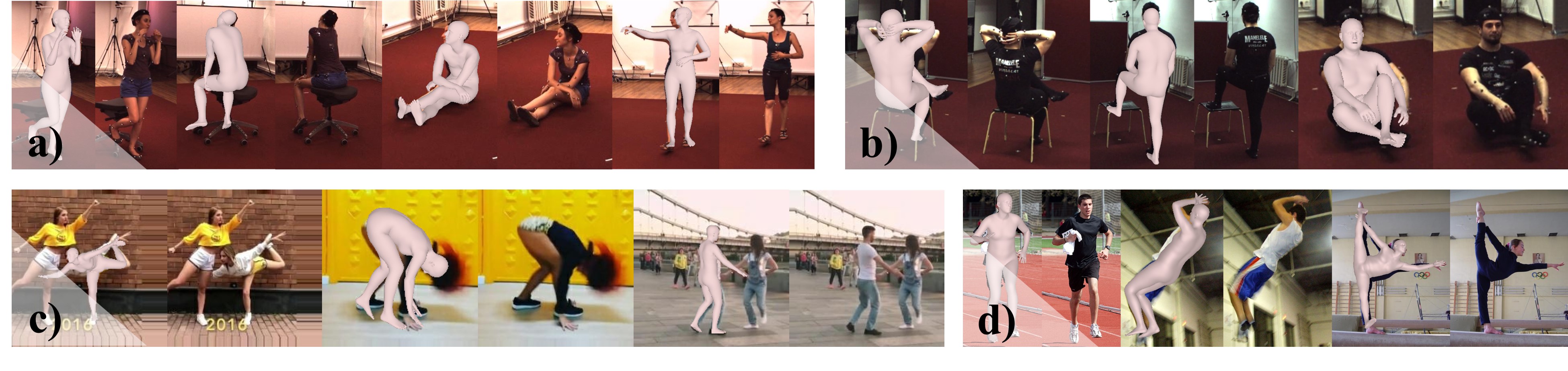}
  \caption{\textbf{Visualization} of dataset images and ground truth annotation. a) Human3.6M. b) HumanSC3D. c) InstaVariety. d) LSPET.}
  \label{fig:data_h36m_humansc3d_instavariety_lspet}
\end{figure}

\textcolor{blue}{
\textbf{Hi4D}~\cite{yin2023hi4d}(\Fig\ref{fig:data_idea_whac_emdb_hi4d}d) is a dataset capturing human interactions with physical contact. It includes 4D textured scans, SMPL annotations, meshes with vertex-level contact annotations, and segmentation masks. The dataset features 20 subject pairs across 100 sequences, totaling over 11,000 frames. Homepage: \url{https://yifeiyin04.github.io/Hi4D/}.
}

\textbf{Human3.6M}~\cite{ionescu2013human3} (\Fig\ref{fig:data_h36m_humansc3d_instavariety_lspet}a) is a studio-based 3D motion capture dataset including 3.6M human poses and corresponding images captured by a high-speed motion capture system. In this paper, we use the annotation generated by NeuralAnnot~\cite{moon2022neuralannot}, which fits the SMPL-X to the GT 2D joints and encompasses a total of  \textasciitilde312.2K annotated data.
Homepage and license: 
\url{http://vision.imar.ro/human3.6m/description.php}.

\textbf{HumanSC3D}~\cite{fieraru2021learning} (\Fig\ref{fig:data_h36m_humansc3d_instavariety_lspet}b) is a studio-based 3d motion capture dataset including 1,032 multiple-view sequences featuring 5K contact events and 1.2M ground truth instances of 3D poses annotated with SMPL-X parameters. We use the open-source train set. Homepage and license: \url{https://sc3d.imar.ro/}.

\textcolor{blue}{
\textbf{IDEA400}~\cite{lin2023motionx}(\Fig\ref{fig:data_idea_whac_emdb_hi4d}a) is an indoor motion dataset comprising 13,000 motion sequences and a total of 2.6 million frames, spanning 400 diverse actions. It includes pseudo SMPL-X annotations and semantic motion labels, annotated through an automated annotation pipeline. Homepage: \url{https://motion-x-dataset.github.io/}.
}

\textbf{InstaVariety}~\cite{kanazawa2019learning} (\Fig\ref{fig:data_h36m_humansc3d_instavariety_lspet}c) is an in-the-wild dataset, and it contains 2.1M images collected from Instagram using 84 hashtags. We use the annotation generated by NeuralAnnot~\cite{moon2022neuralannot}, which fits the SMPL to the GT 2D joints and encompasses a total of \textasciitilde218.5K annotated data.
Homepage:
\url{https://github.com/akanazawa/human_dynamics/blob/master/doc/insta_variety.md}

\textbf{LSPET}~\cite{johnson2010clustered} (\Fig\ref{fig:data_h36m_humansc3d_instavariety_lspet}d) is an in-the-wild dataset, and it contains 10K images. 
In this paper, we use the annotation generated by EFT~\cite{HanbyulJoo2022eft}, which fits the SMPL to the GT 2D joints and encompasses a total of 2,946 annotated data.
Homepage:
\url{http://sam.johnson.io/research/lspet.html}.

\begin{figure}[t]
  \centering
  \includegraphics[width=\linewidth]{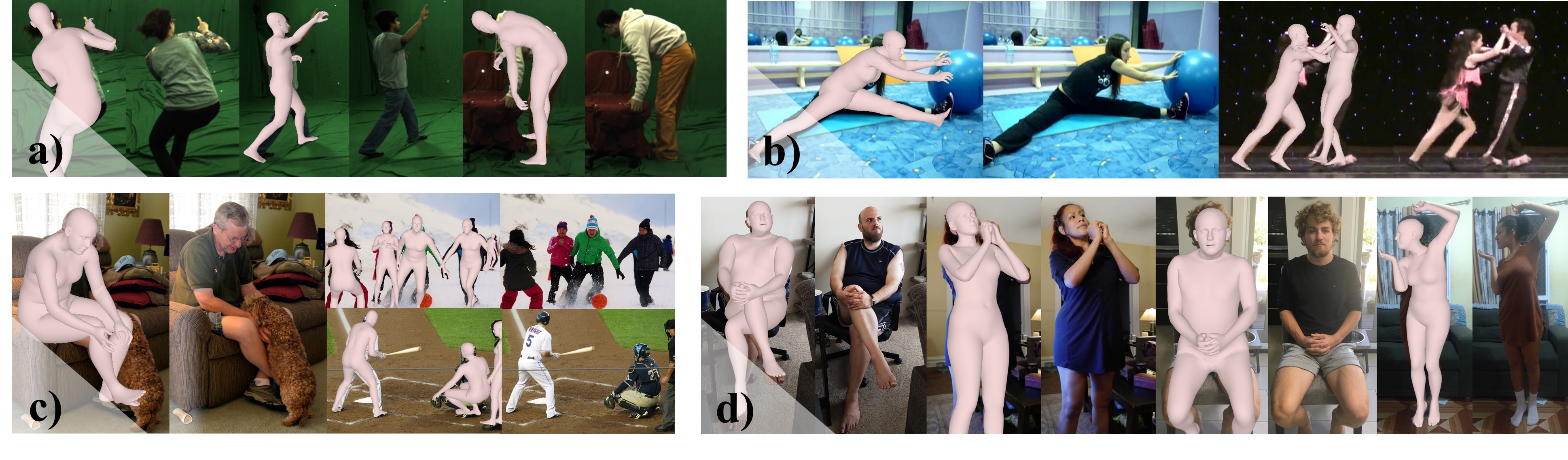}
  \caption{\textbf{Visualization} of dataset images and ground truth annotation. a) MPI-INF-3DHP. b) MPII. c) MSCOCO. d) MTP.}
  \label{fig:data_mpiinf3dhp_mpii_mscoco_mtp}
\end{figure}

\textcolor{blue}{
\textbf{MoYo}~\cite{tripathi2023ipman}(\Fig\ref{fig:data_sloper_signavatars_damon_moyo}d) is a dataset featuring 200 complex yoga poses captured in a motion capture studio with 8 synchronized cameras. It includes SMPL-X annotations derived from MoCap data, along with body-floor contact information, center of mass (CoM), and pressure annotations obtained using a pressure measurement mat. Homepage: \url{https://ipman.is.tue.mpg.de/}.
}

\textbf{MPI-INF-3DHP}~\cite{mehta2017monocular} ((\Fig\ref{fig:data_mpiinf3dhp_mpii_mscoco_mtp}a) is captured with a multi-camera markerless motion capture system in constrained indoor and complex outdoor scenes. It records 8 actors performing 8 activities from 14 camera views. We use the annotations generated by NeuralAnnot~\cite{moon2022neuralannot}, which fits the SMPL-X to the GT 2D joints and encompasses a total of 939,847 annotated data. To efficiently train our model, we downsample the NeuralAnnot data to \textasciitilde180.0K.
Homepage and license:
\url{https://vcai.mpi-inf.mpg.de/3dhp-dataset/}.

\textbf{MPII}~\cite{andriluka14cvpr} ((\Fig\ref{fig:data_mpiinf3dhp_mpii_mscoco_mtp}b) is a widely used in-the-wild dataset that offers a diverse collection of approximately 25K images. 
Each image within the dataset contains one or more instances, resulting in a total of over 40K annotated people instances. 
Among the 40k samples, \textasciitilde28k samples are used for training, while the remaining samples are reserved for testing. We use the annotations generated by NeuralAnnot~\cite{moon2022neuralannot}, which fits the SMPL-X to the GT 2D joints and encompasses a total of  \textasciitilde28.9K annotated data.
Homepage and license:
\url{http://human-pose.mpi-inf.mpg.de/}.

\textbf{MSCOCO}~\cite{lin2014microsoft} (\Fig\ref{fig:data_mpiinf3dhp_mpii_mscoco_mtp}c) is a large-scale object detection, segmentation, keypoint detection, and captioning dataset. The subset for the keypoint detection contains more than 200K images and 250K person instances.
We use the annotations generated by NeuralAnnot~\cite{moon2022neuralannot}, which fits the SMPL-X to the GT 2D joints and encompasses a total of \textasciitilde149.8K annotated data.
Homepage and license:
\url{https://cocodataset.org/#home}.

\textbf{MTP}~\cite{muller2021self} (\Fig\ref{fig:data_mpiinf3dhp_mpii_mscoco_mtp}d) is an in-door dataset containing images of actors mimicking different hard SMPL-X poses with self-contact. There are 3.7K images from 148 subjects with pseudo ground-truth SMPL-X parameters and 2D keypoints. We use 3.2K instances in training. Homepage and licenses:
\url{https://tuch.is.tue.mpg.de/}.

\begin{figure}[t]
  \centering
  \includegraphics[width=\linewidth]{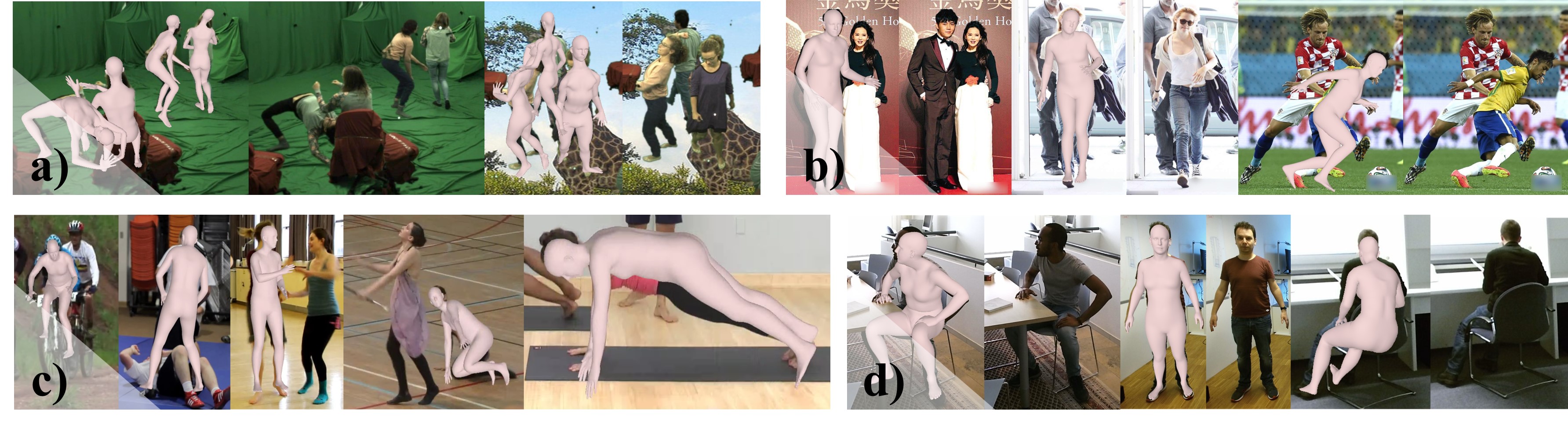}
  \caption{\textbf{Visualization} of dataset images and ground truth annotation. a) MuCo-3DHP. b) OCHuman. c) PoseTrack. d) PROX.}
  \label{fig:data_muco_ochuman_posetrack_prox}
\end{figure}

\textbf{MuCo-3DHP}~\cite{Mehta2018SingleShotM3} (\Fig\ref{fig:data_muco_ochuman_posetrack_prox}a) is an in-door multi-person dataset composited by cropping and overlaying person in MPI-INF-3DHP\cite{mehta2017monocular} with segmentation masks. It has 400K frames and contains 8 subjects with 2 different clothing for each subject. It is shot with 12 different camera positions. It has ground truth 3D keypoints and fitted SMPL parameters. We use 465.3K annotated data in training.  Homepage and licenses: \url{https://vcai.mpi-inf.mpg.de/projects/SingleShotMultiPerson/}.

\textbf{OCHuman}~\cite{zhang2019pose2seg} (\Fig\ref{fig:data_muco_ochuman_posetrack_prox}b) is an in-the-wild datset, and it focuses on heavily occluded human. This dataset contains 8110 detailed annotated human instances within 4731 images. 
We use the annotations generated by EFT~\cite{HanbyulJoo2022eft}, which fits the SMPL to the GT 2D joints and encompasses a total of 2,495 annotated data.
Homepage:
\url{https://github.com/liruilong940607/OCHumanApi}.

\textbf{PoseTrack}~\cite{andriluka2018posetrack} (\Fig\ref{fig:data_muco_ochuman_posetrack_prox}c) is a large-scale benchmark for multi-person pose estimation and tracking in videos. It contains 514 videos and includes 66,374 frames. 
Ee use the annotations generated by EFT~\cite{HanbyulJoo2022eft}, which fits the SMPL to the GT 2D joints and encompasses a total of \textasciitilde28.5K annotated data.
Homepage:
\url{https://posetrack.net}.

\textbf{PROX}~\cite{hassan2019resolving} (\Fig\ref{fig:data_muco_ochuman_posetrack_prox}d) qualitative dataset is a human-scene interaction dataset that showcases 12 indoor scenes and 20 subjects engaging with these scenes. It comprises 100K RGBD frames with pseudo ground-truth SMPL-X fittings. During training, only the RGB images are utilized, and they are horizontally flipped to align with the SMPL-X annotations. We use 88.1K instances for training.
Homepage and license: \url{https://prox.is.tue.mpg.de/}.

\begin{figure}[t]
  \centering
  \includegraphics[width=\linewidth]{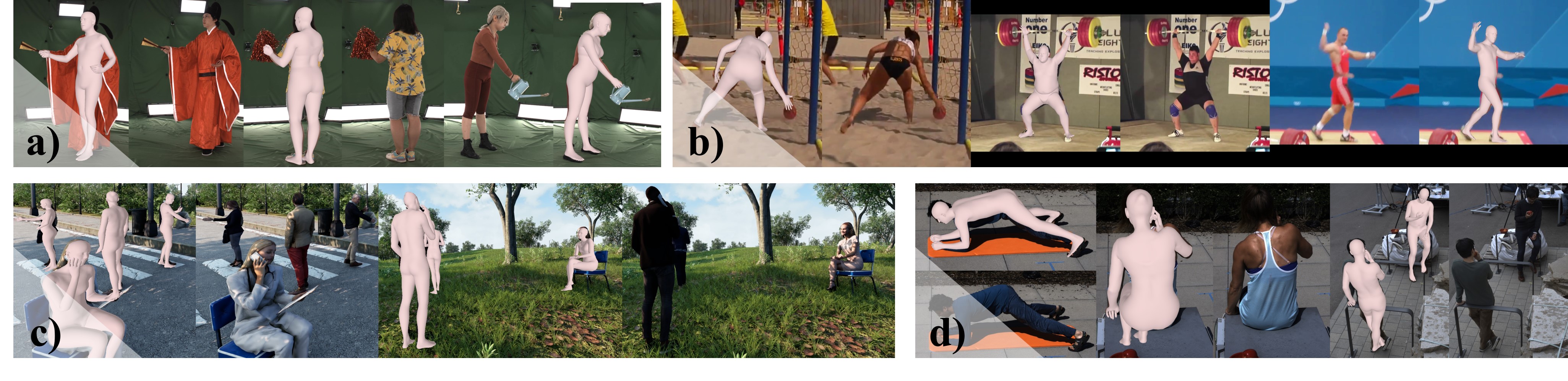}
  \caption{\textbf{Visualization} of dataset images and ground truth annotation. a) RenBody. b) SSP3D. c) SPEC. d) RICH.}
  \label{fig:data_renbody_ssp3d_rich_spec}
\end{figure}
\begin{figure}[t]
  \centering
  \includegraphics[width=\linewidth]{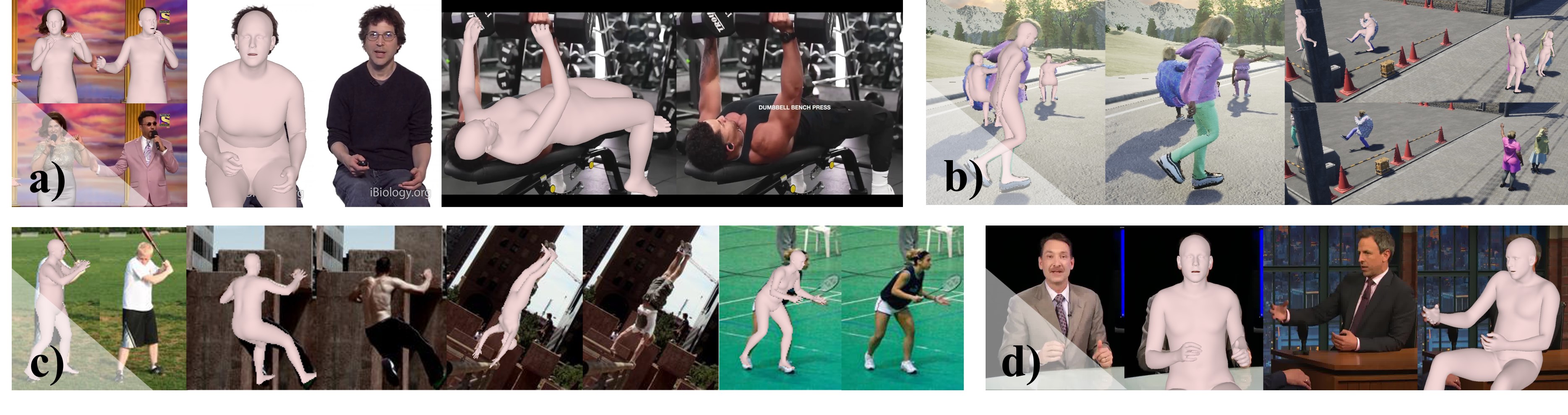}
  \caption{\textbf{Visualization} of dataset images and ground truth annotation. a) UBody. b) SynBody. c) UP3D. d) Talkshow.}
  \label{fig:data_ubody_synbody_up3d_talkshow}
\end{figure}

\textbf{DNA-Rendering}~\cite{renbody} (\Fig\ref{fig:data_renbody_ssp3d_rich_spec}a) is a large-scale multi-view studio-based dataset with ultra high resolution that features diversity in motion, clothing, and object interactions. RenBody has more than 1.5K human instances and 5K motion sequences with up to 60 RGB views and 4 Kinect views. Corresponding SMPL-X annotation is based on SMPLify-X\cite{pavlakos2019expressive}, and multi-view optimization. We separate the 60 RGB views into 48 and 12 views based on different camera distributions and captured resolutions. Homepage: \url{https://magichub.com/datasets/openxd-renbody/}.

\textbf{SPEC}~\cite{kocabas2021spec} (\Fig\ref{fig:data_renbody_ssp3d_rich_spec}c) is a synthetic dataset featuring diverse and unique camera viewpoints. It has 22191 images with 71982 ground truth instances with SMPL parameters as a train set and 3783 images with 12071 ground truth instances as a test set. Homepage and licenses: \url{https://spec.is.tue.mpg.de/index.html}. 

\textbf{RICH}~\cite{huang2022capturing} (\Fig\ref{fig:data_renbody_ssp3d_rich_spec}d) is a human-scene contact dataset. It encompasses a comprehensive collection of 142 single or multi-person multiview videos capturing 22 subjects in 5 static indoor or outdoor scenes with 6-8 static cameras. RICH comprises a rich set of resources, including a total of 90K posed 3D body meshes, each associated with dense full-body contact labels in both SMPL-X and SMPL mesh topology. We convert the original image from png and bmp to jpg and train the model with the train set, including \textasciitilde243.4K instances. 
Homepage and license:
\url{https://rich.is.tue.mpg.de/index.html}.

\textcolor{blue}{
\textbf{SignAvatars~\cite{yu2025signavatars}}
(\Fig\ref{fig:data_sloper_signavatars_damon_moyo}b) is a large-scale, multi-prompt 3D sign language motion dataset comprising 70K motion sequences and 8.34M annotations in SMPL-X format for whole-body and MANO format for hand poses. 70,000 videos from 153 signers are collected and annotated to advance digital communication for the hard-of-hearing community. Homepage: \url{https://signavatars.github.io/}.
}

\textcolor{blue}{
\textbf{SLOPER4D~\cite{dai2023sloper4d}}
(\Fig\ref{fig:data_sloper_signavatars_damon_moyo}a)is a scene-aware, human-scene interactions dataset captured in large urban environments and annotated in world space. It features 12 human subjects recorded across 10 diverse urban scenes from an egocentric perspective using LiDAR, IMUs, and a camera. The dataset comprises 15 sequences, over 100K LiDAR frames, 300K video frames, and 500K motion frames derived from IMU data. Homepage: \url{http://www.lidarhumanmotion.net/sloper4d/}.
}

\textbf{SSP3D}~\cite{STRAPS2018BMVC} SSP-3D (\Fig\ref{fig:data_renbody_ssp3d_rich_spec}b) is a small-scale dataset consisting of 311 images of persons in tight-fitted clothes in sports, with a variety of body shapes and poses. Pseudo-ground-truth SMPL body model parameters obtained via multi-frame optimization with shape consistency. Homepage: \url{https://github.com/akashsengupta1997/SSP-3D}.

\textbf{SynBody}~\cite{yang2023synbody} (\Fig\ref{fig:data_ubody_synbody_up3d_talkshow}b)
is a large-scale synthetic dataset featuring a massive number of diverse subjects and high-accuracy annotations which includes 1.7M multi-person image instances with 3D pose and shape annotations. SynBody covers 10K human body models, 1K actions, and many viewpoints. Annotations include both accurate SMPL and SMPL-X parameters. Synbody also features layered human body models and clothes. Homepage: \url{https://maoxie.github.io/SynBody/}.

\textbf{Talkshow}~\cite{yi2023generating} (\Fig\ref{fig:data_ubody_synbody_up3d_talkshow}d) is a large-scale dataset featuring talking videos of 4 subjects in 4 different scenarios. It contains 26.9 hours of video clips at 30 fps and has synchronized audio and fitted SMPL-X annotations. We obtain the video clips from the author and convert them to images, including of 332.7K instances. Homepage and licenses: \url{https://talkshow.is.tue.mpg.de/}.

\textbf{UBody}~\cite{lin2023one} (\Fig\ref{fig:data_ubody_synbody_up3d_talkshow}b) is a large-scale dataset that features a diverse range of real-life scenarios that cater to various downstream tasks, such as fitness videos, VLOGs, movies, online classes, video conferences, talk shows, and sign languages. In these scenarios, typically only the subject's upper body is visible. Heavy truncation and a focus on expressive
gestures and facial expressions make UBody especially challenging. We use the intra-scene protocol. UBody is under CC-BY-NC-SA 4.0 licenses. Homepage: \url{https://github.com/IDEA-Research/OSX}.

\textbf{UP3D}~\cite{lassner2017unite} (\Fig\ref{fig:data_ubody_synbody_up3d_talkshow}c) is an in-the-wild dataset containing 7,126 images. To obtain 3D high-quality annotations, it extends the SMPLify~\cite{bogo2016keep} and fits a pseudo label (SMPL) for each image.
Homepage and license:
\url{https://files.is.tuebingen.mpg.de/classner/up/}.

\textcolor{blue}{
\textbf{WHAC-A-Mole}~\cite{yin2025whac}
(\Fig\ref{fig:data_idea_whac_emdb_hi4d}b) is a synthetic dataset that provides annotated, world-grounded humans and cameras. It features diverse interactive multi-human motions and realistic camera trajectories. The dataset includes 2,434 sequences and 1.46 million human instances, all annotated in the SMPL-X format, capturing detailed motions and human-human interactions. Homepage: \url{https://wqyin.github.io/projects/WHAC/}.
}

\ifCLASSOPTIONcaptionsoff
  \newpage
\fi




{
\bibliographystyle{IEEEtran}
\bibliography{references}
}
\vspace{-12mm}

\begin{IEEEbiography}[{\includegraphics[width=1in,height=1.25in,clip,keepaspectratio]{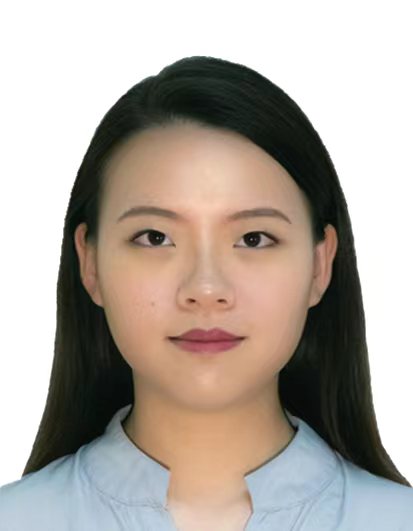}}]{Wanqi Yin} 
 is currently a Ph.D. candidate at the University of Tokyo advised by Prof. Atsushi Yamashita, and an Algorithm Researcher at SenseTime Research. She
 received her Master's degree from the University of Tokyo in 2020 and her Bachelor's degree from Nanyang Technological University before that. Her research interests include computer vision focusing on human motion capture and multi-camera systems. She has published papers at top-tier conferences including CVPR, NeurIPS, ICCV, and ECCV. 
\end{IEEEbiography}

\vspace{-12mm}

\begin{IEEEbiography}[{\includegraphics[width=1in,height=1.25in,clip,keepaspectratio]{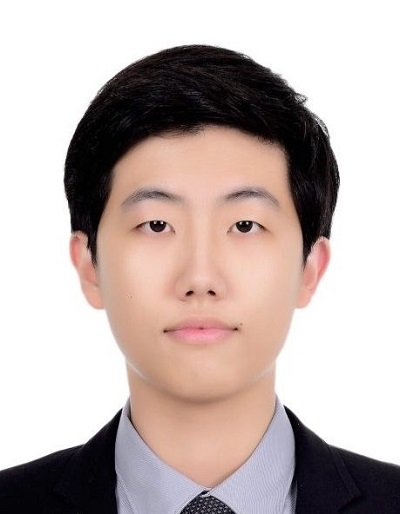}}]{Zhongang Cai}
  is a Staff Research Scientist at SenseTime Research. He attained his Ph.D. at Nanyang Technological University (NTU), advised by Prof. Ziwei Liu and Prof. Chen Change Loy. Before that, he attained his bachelor’s degree at NTU and was awarded the Lee Kuan Yew Gold Medal as the top student. His research focuses on human motion capture and generation. To date, he has published more than 30 papers on top venues such as NeurIPS, ICLR, CVPR, ICCV, and ECCV. He is actively involved in academic services and was awarded the Outstanding/Highlighted Reviewer for ICLR'22 and ICCV'21.
\end{IEEEbiography}

\vspace{-12mm}

\begin{IEEEbiography}[{\includegraphics[width=1in,height=1.25in,clip,keepaspectratio]{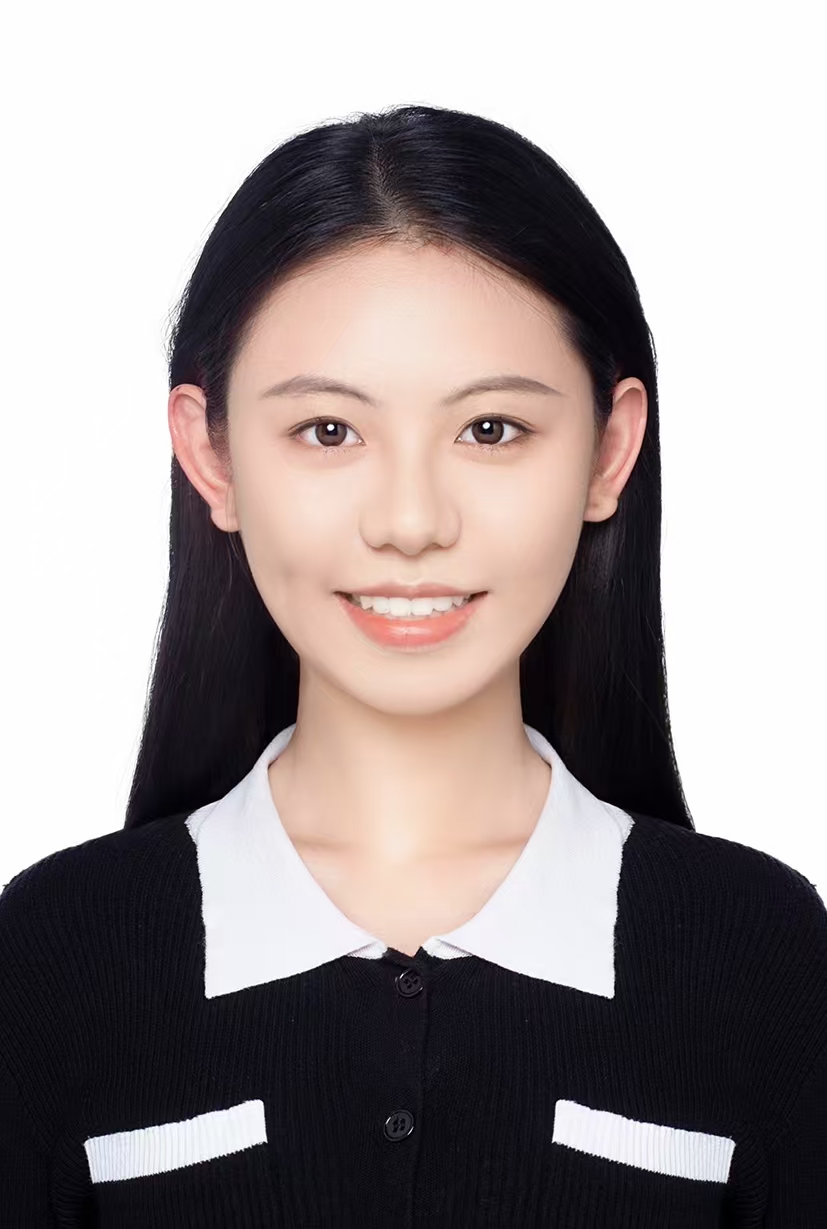}}]{Ruisi Wang}
 is a researcher at SenseTime Research. She received her bachelor's degree from the College of Computing and Data Science, Nanyang Technological University. During her university years, she actively engaged in international competitions, attaining second place in the HPC AI Innovation Challenge 2024, first place in the HPC ISC2023 Student Cluster Competition, etc. She is now interested in computer vision, deep learning, and human motion capture. She has published several conference papers at CVPR, ECCV.
\end{IEEEbiography}

\vspace{-12mm}

\begin{IEEEbiography}[{\includegraphics[width=1in,height=1.25in,clip,keepaspectratio]{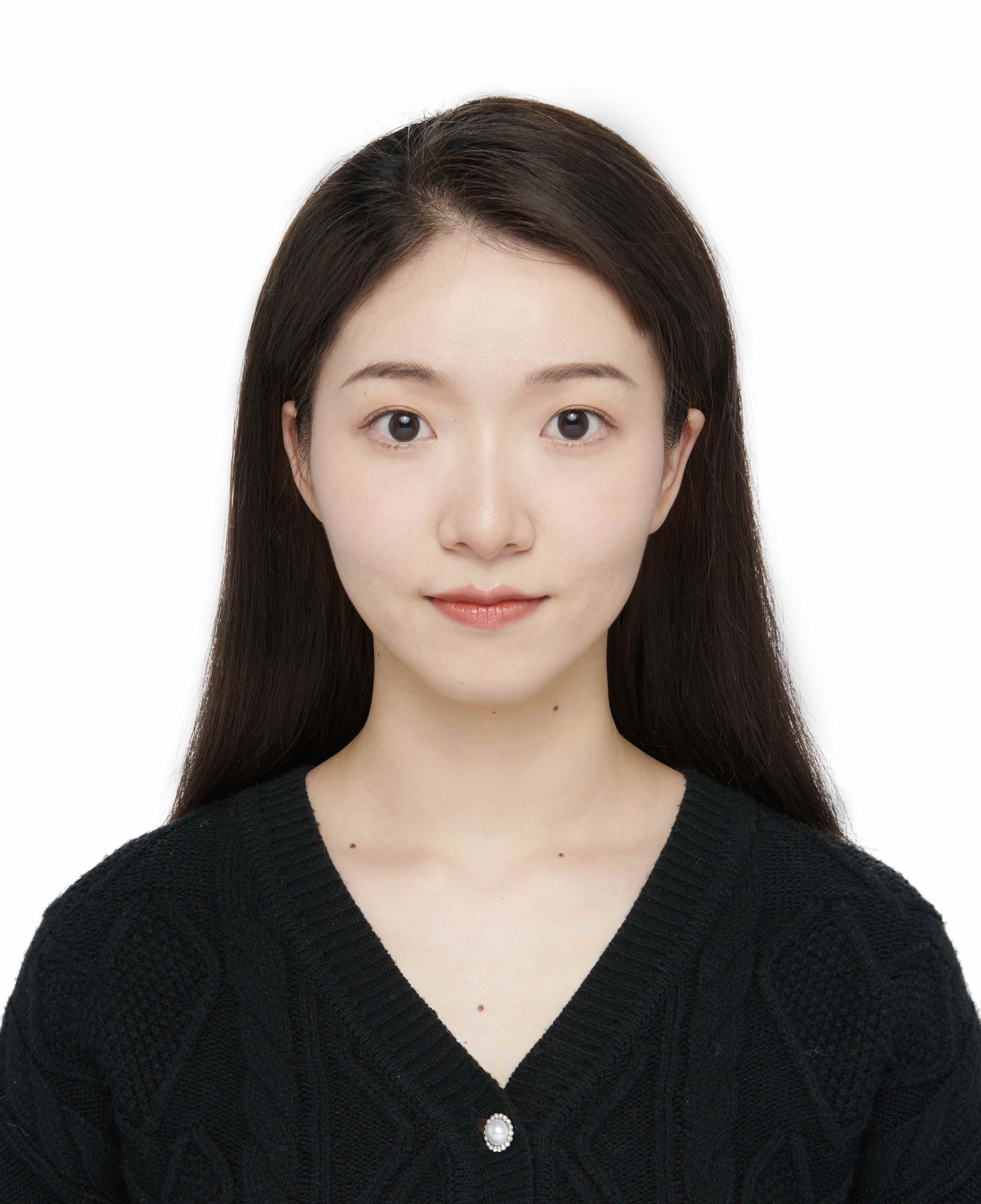}}]{Ailing Zeng}
 (Member, IEEE) is a senior researcher at Tencent AI Lab. Previously, she was a researcher at IDEA research. She has obtained the Ph.D. degree from the Department of Computer Science and Engineering, the Chinese University of Hong Kong. Her research areas include computer vision, deep learning, and multi-modal generation, targeting to build multi-modal human-like intelligent agents on scalable big data. She has published over forty top-tier conference papers at CVPR, ICCV, ECCV, ICLR, ICML, Neurips,etc, and serve as the reviewers, area chairs, or associate editors of many top conferences and journals.
\end{IEEEbiography}

\vspace{-10mm}

\begin{IEEEbiography}[{\includegraphics[width=1in,height=1.25in,clip,keepaspectratio]{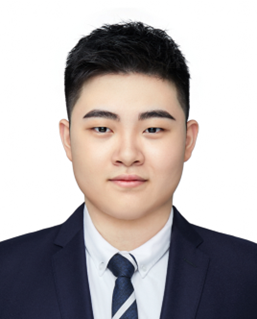}}]{Chen Wei}
  graduated from Nanyang Technological University with a bachelor's degree in Electric and Electronic Engineering. He was an algorithm research intern at S-Lab @ NTU in 2021 and joined SenseTime International Pte. Ltd in 2022. His work focuses on the generation of synthetic human data from game engines and the standardization of human-centric computer vision datasets. He has published several papers on top-tier conferences, including ICCV, NeurIPS, and CVPR. 
\end{IEEEbiography}
\vspace{-8mm}

\begin{IEEEbiography}[{\includegraphics[width=1in,height=1.25in,clip,keepaspectratio]{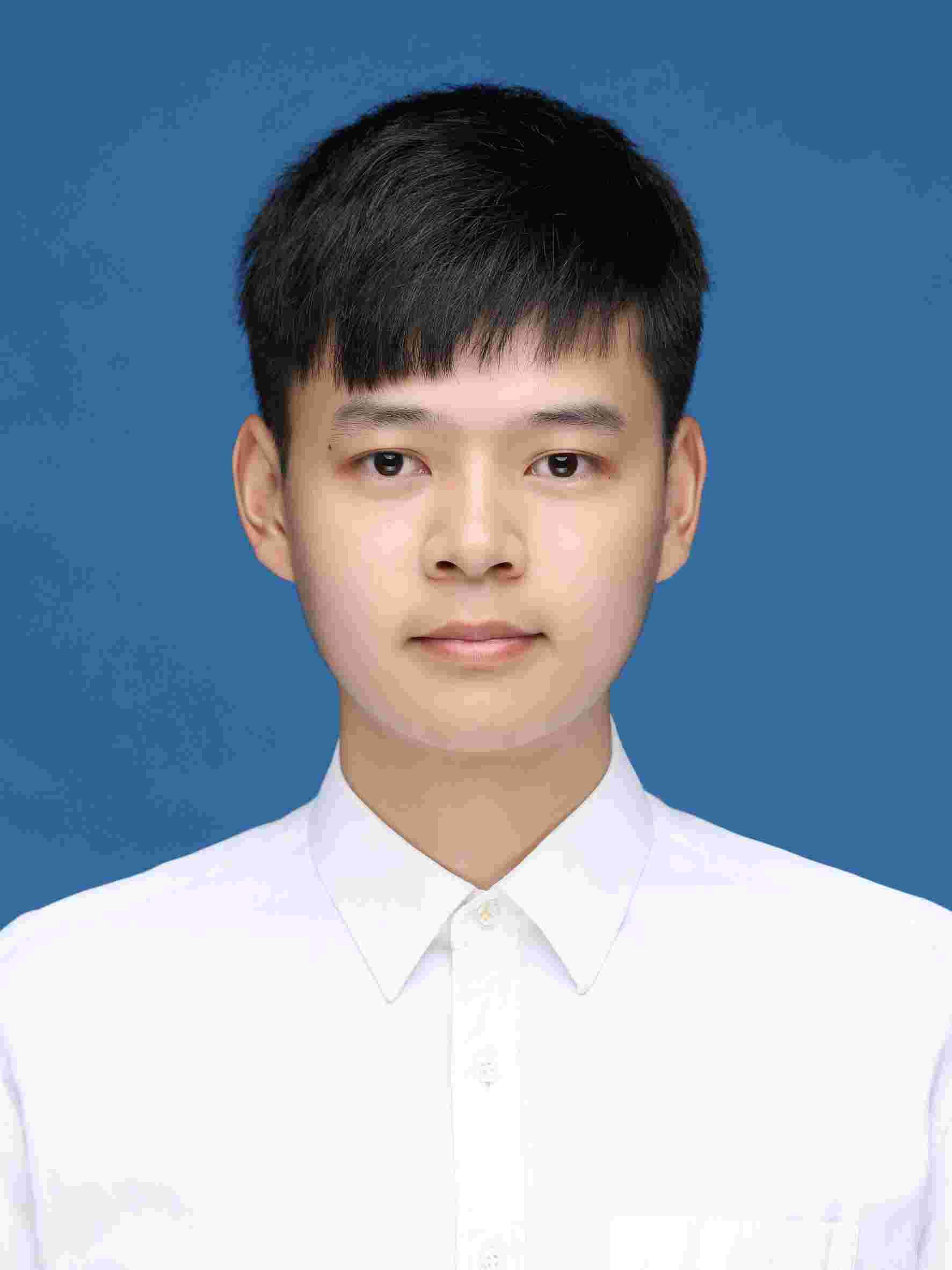}}]{Qingping Sun}
  Qingping Sun received a bachelor's degree from Hainan University and a master's degree from Hunan University. He is pursuing a PhD degree at the City University of Hong Kong and working as a research intern at SenseTime Research. His research interests include computer vision and computer graphics.
\end{IEEEbiography}

\vspace{-12mm}

\begin{IEEEbiography}[{\includegraphics[width=1in,height=1.25in,clip,keepaspectratio]{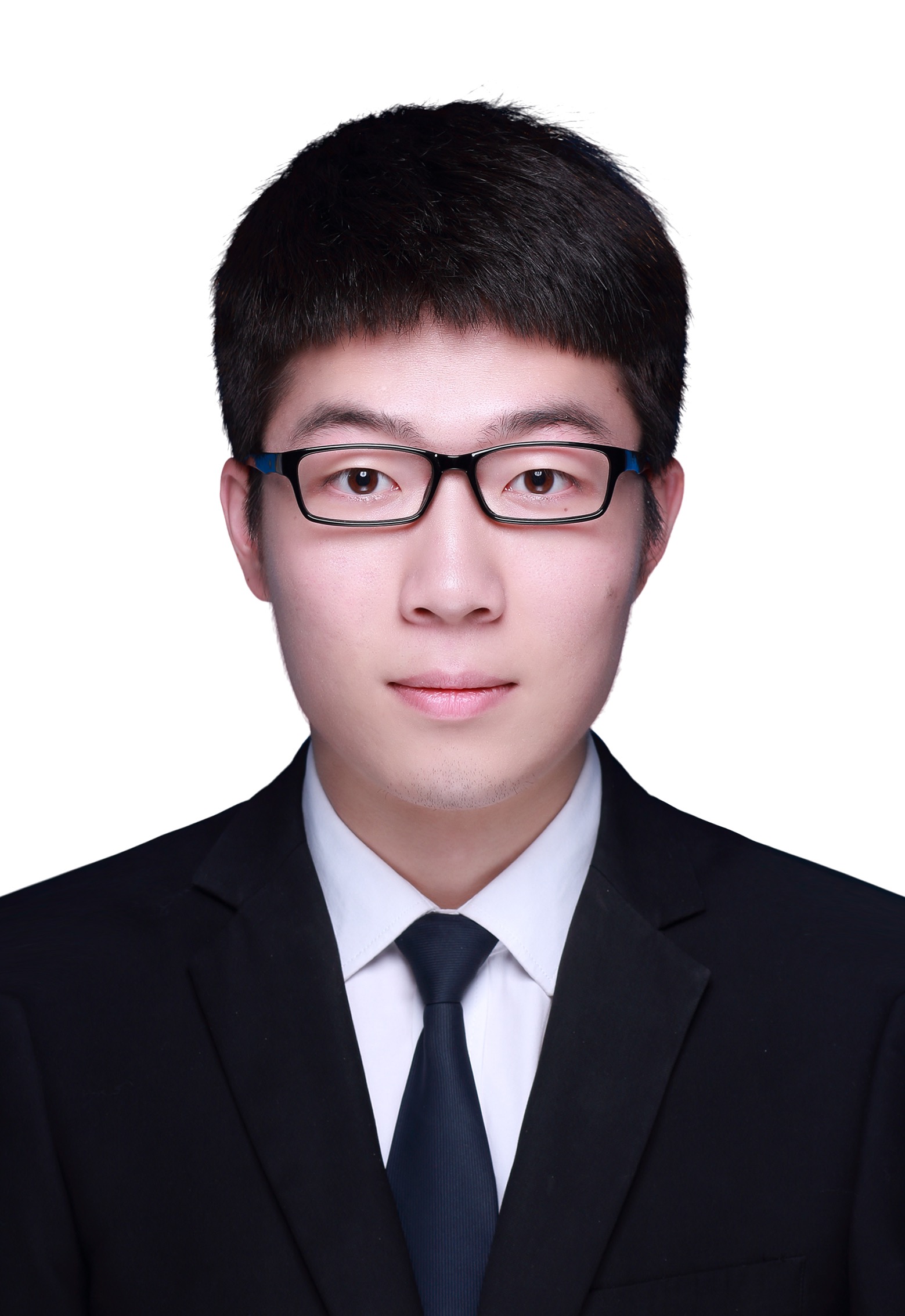}}]{Haiyi Mei}
received a B.S. degree in Automation from Nanjing University of Science and Technology and a Master's degree in Biomedical Engineering from Shandong University. He is currently a Research Fellow at SenseTime, focusing on text/image-to-video generation and synthetic data for computer vision tasks. He has published papers in top-tier conferences such as CVPR, ICCV, and NeurIPS.
\end{IEEEbiography}
\vspace{-12mm}

\begin{IEEEbiography}[{\includegraphics[width=1in,height=1.25in,clip,keepaspectratio]{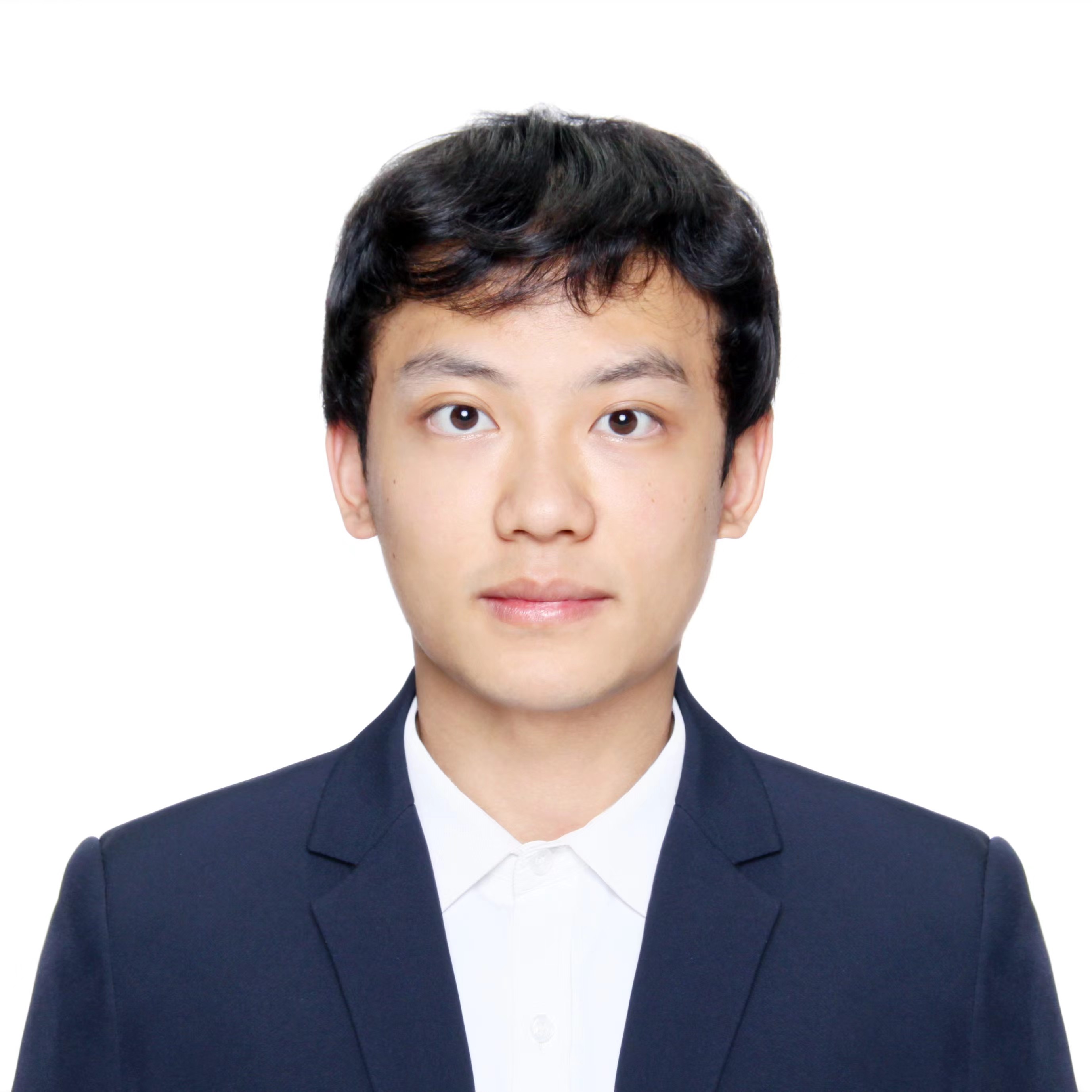}}]{Yanjun Wang}
 is currently a master’s student in Computer Science at UCLA. Previously, he worked as a research intern at SenseTime Group Inc. under the supervision of Dr. Lei Yang. He earned his bachelor’s degree from Shanghai Jiao Tong University in 2022, under the supervision of Prof. Li Song. He has published multiple papers in relevant fields at international conferences, including CVPR and NeurIPS.
\end{IEEEbiography}

\vspace{-12mm}

\begin{IEEEbiography}[{\includegraphics[width=1in,height=1.25in,clip,keepaspectratio]{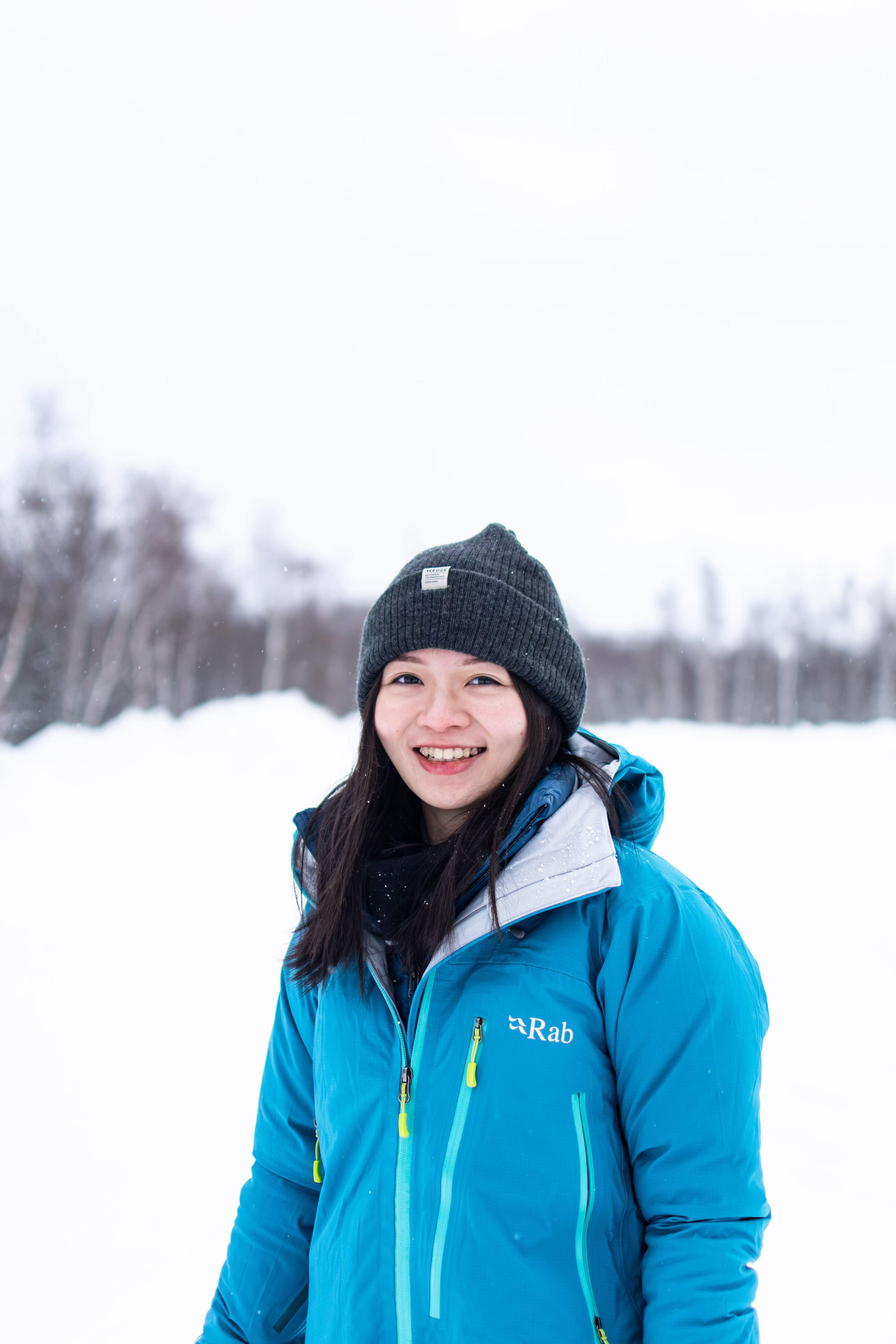}}]{Hui En Pang}
  is a PhD student in Computer Science at Nanyang Technological University. She holds a B.S. in Biology and a Master’s in Geographic Information Systems from National University of Singapore. She has publications in NeurIPS and work experience in Grab, AI Singapore, and SenseTime.
\end{IEEEbiography}

\vspace{-12mm}

\begin{IEEEbiography}[{\includegraphics[width=1in,height=1.25in,clip,keepaspectratio]{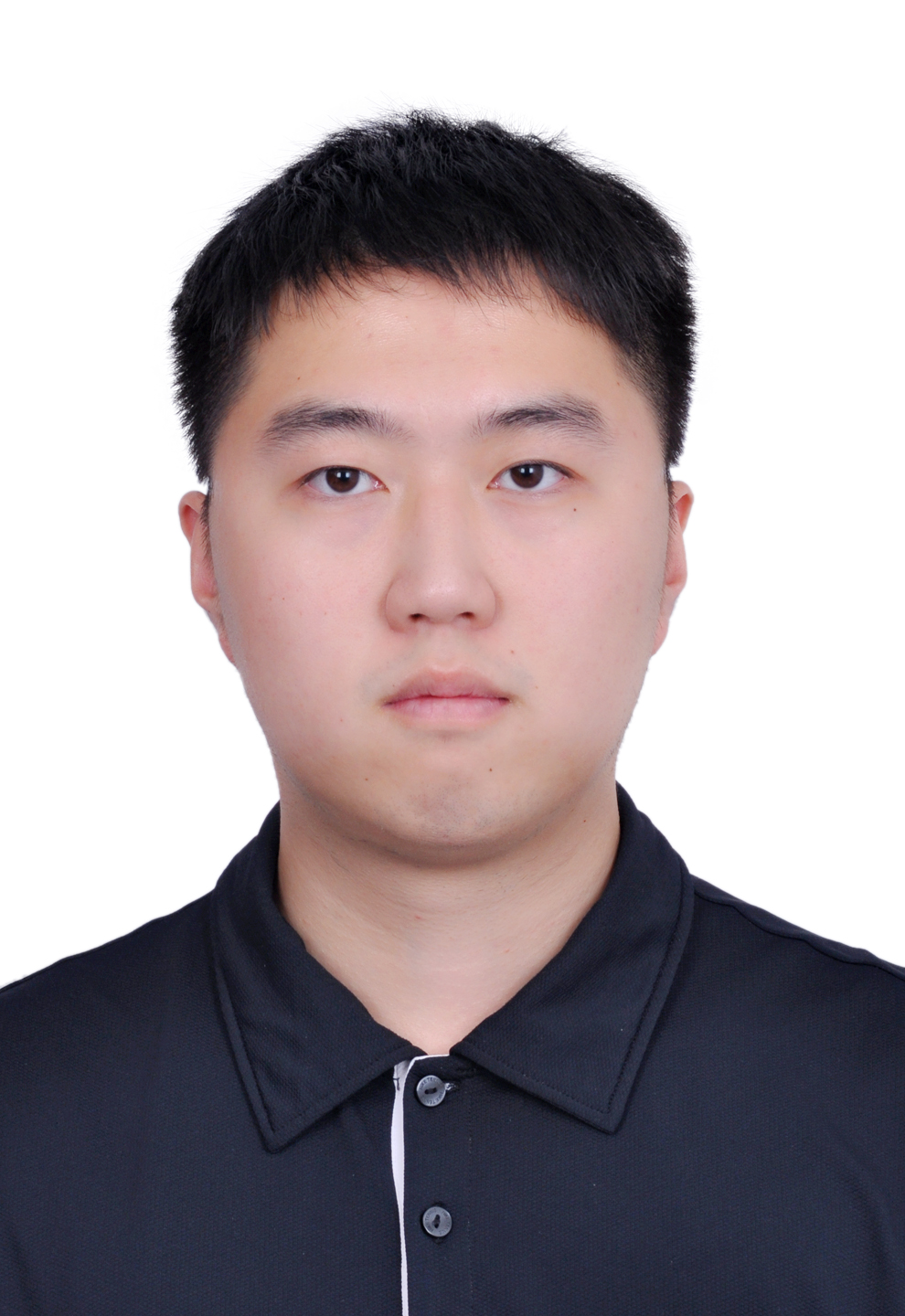}}]{Mingyuan Zhang}
  received a B.S. degree in computer science and engineering from Beihang University, China. He is currently pursuing a Ph.D. degree at MMLab@NTU, advised by Prof. Ziwei Liu. His research interests in computer vision include motion synthesis, 3D pose estimation, and scene understanding. He has published papers on top-tier conferences, including CVPR, ECCV, ICCV, ICLR, and AAAI.
\end{IEEEbiography}

\vspace{-12mm}

\begin{IEEEbiography}[{\includegraphics[width=1in,height=1.25in,clip,keepaspectratio]{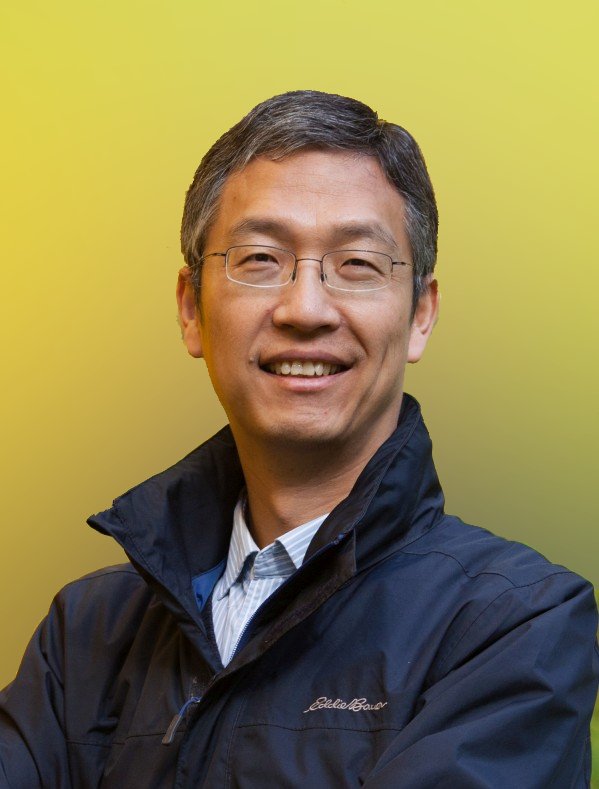}}]{Lei Zhang}
 (Fellow, IEEE) received the Ph.D. degree in computer science from Tsinghua University, Beijing, China, in 2001. He is currently the chief scientist of computer vision and robotics with the International Digital Economy Academy (IDEA) and an adjunct professor at the Hong Kong University of Science and Technology, Guangzhou, China. Before his current post, he was a principal researcher and research manager with Microsoft. He has authored or coauthored more than 150 technical papers, and holds more than 60 U.S. patents in his research field, which include computer vision and machine learning, with a particular focus on generic visual recognition at a large scale. He was an editorial board member for IEEE Transactions on Multimedia, IEEE Transactions on Circuits and Systems for Video Technology, and Multimedia System Journal and as the area chair of many top conferences.
\end{IEEEbiography}

\vspace{-8mm}

\begin{IEEEbiography}[{\includegraphics[width=1in,height=1.25in,clip,keepaspectratio]{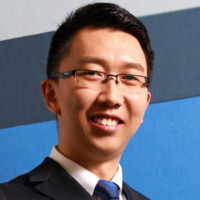}}]{Chen Change Loy}
 (Senior Member, IEEE) received the PhD degree in computer science from the Queen Mary University of London, in 2010. He is a professor with the College of Computing and Data Science , Nanyang Technological University. Prior to joining NTU, he served as a research assistant professor with the Department of Information Engineering, The Chinese University of Hong Kong, from 2013 to 2018. His research interests include computer vision and deep learning with a focus on image/video restoration and enhancement, generative tasks, and representation learning. He serves as an Associate Editor of the Computer Vision and Image Understanding (CVIU), International Journal of Computer Vision (IJCV) and IEEE Transactions on Pattern Analysis and Machine Intelligence (TPAMI). He also serves/served as the Area Chair of top conferences such as ICCV, CVPR, ECCV, ICLR and NeurIPS.
\end{IEEEbiography}

\vspace{-8mm}

\begin{IEEEbiography}[{\includegraphics[width=1in,height=1.25in,clip,keepaspectratio]{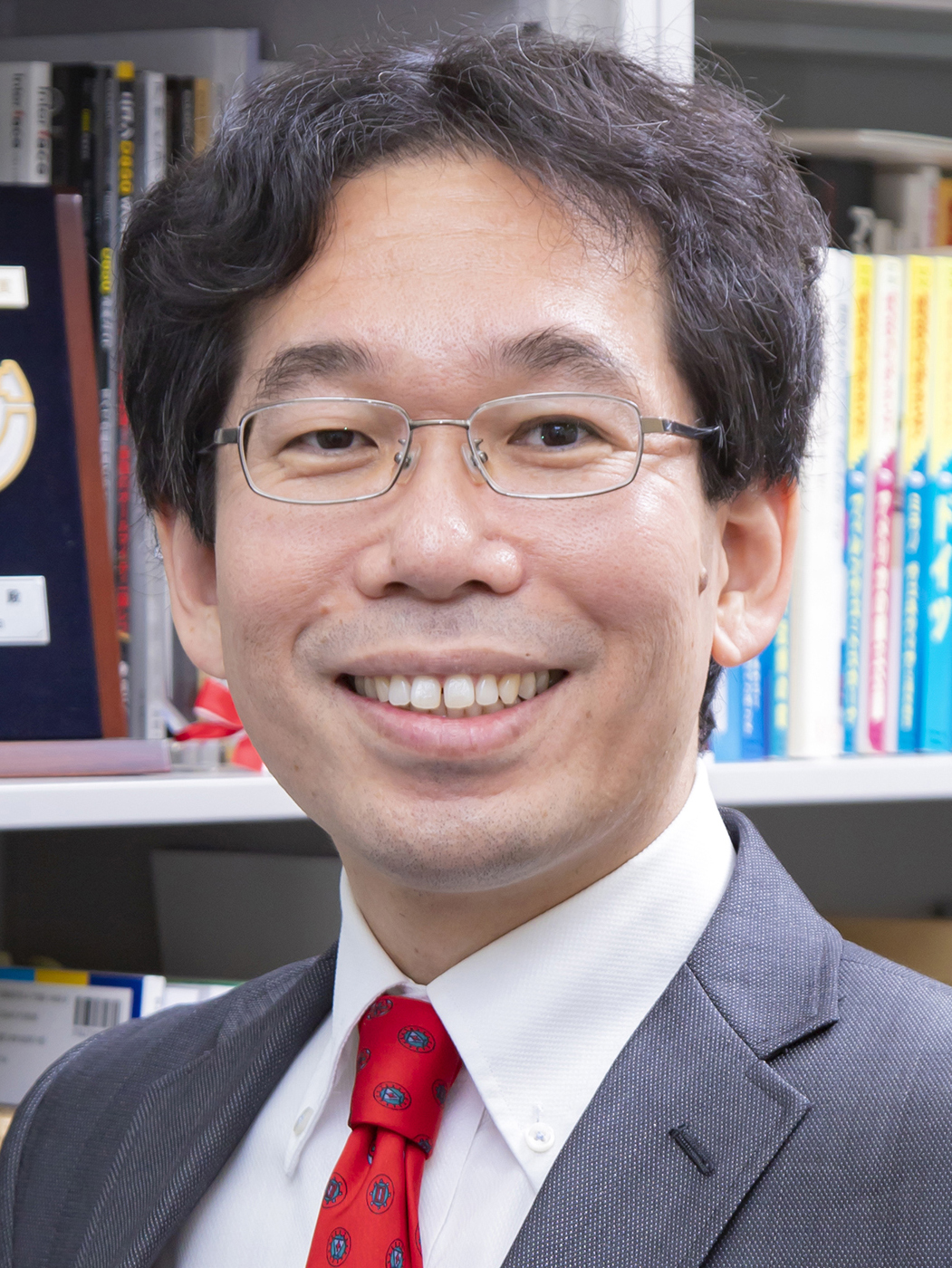}}]{Atsushi Yamashita} 
    (Senior Member, IEEE) is a professor at the University of Tokyo. He received B.E., M.E., and Ph.D. degrees in Engineering from the University of Tokyo, Japan. Previously, he was an Associate Professor at Shizuoka University, Japan, and a Visiting Associate at the California Institute of Technology, USA. His research interests include robotics, image processing, and computer vision.
\end{IEEEbiography}

\vspace{-8mm}

\begin{IEEEbiography}[{\includegraphics[width=1in,height=1.25in,clip,keepaspectratio]{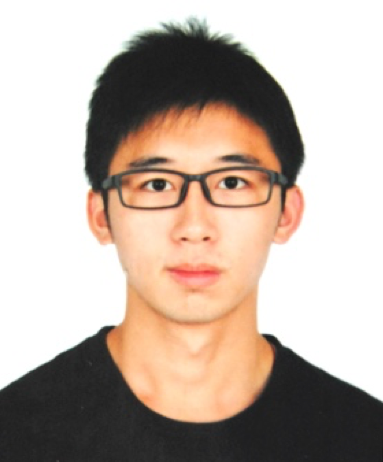}}]{Lei Yang}
  is currently a Research Director at SenseTime Group Inc. Lei received his Ph.D. degree from the Chinese University of HongKong in 2020, advised by Prof. Dahua Lin. Prior to that, Lei obtained his B.E. degree from Tsinghua University in 2015. He has authored over 30 papers on top conferences or journals in relevant fields, including CVPR, ICCV, ECCV, NeurIPS, SIGGRAPH, IJCV and T-PAMI.
\end{IEEEbiography}

\vspace{-8mm}

\begin{IEEEbiography}[{\includegraphics[width=1in,height=1.25in,clip,keepaspectratio]{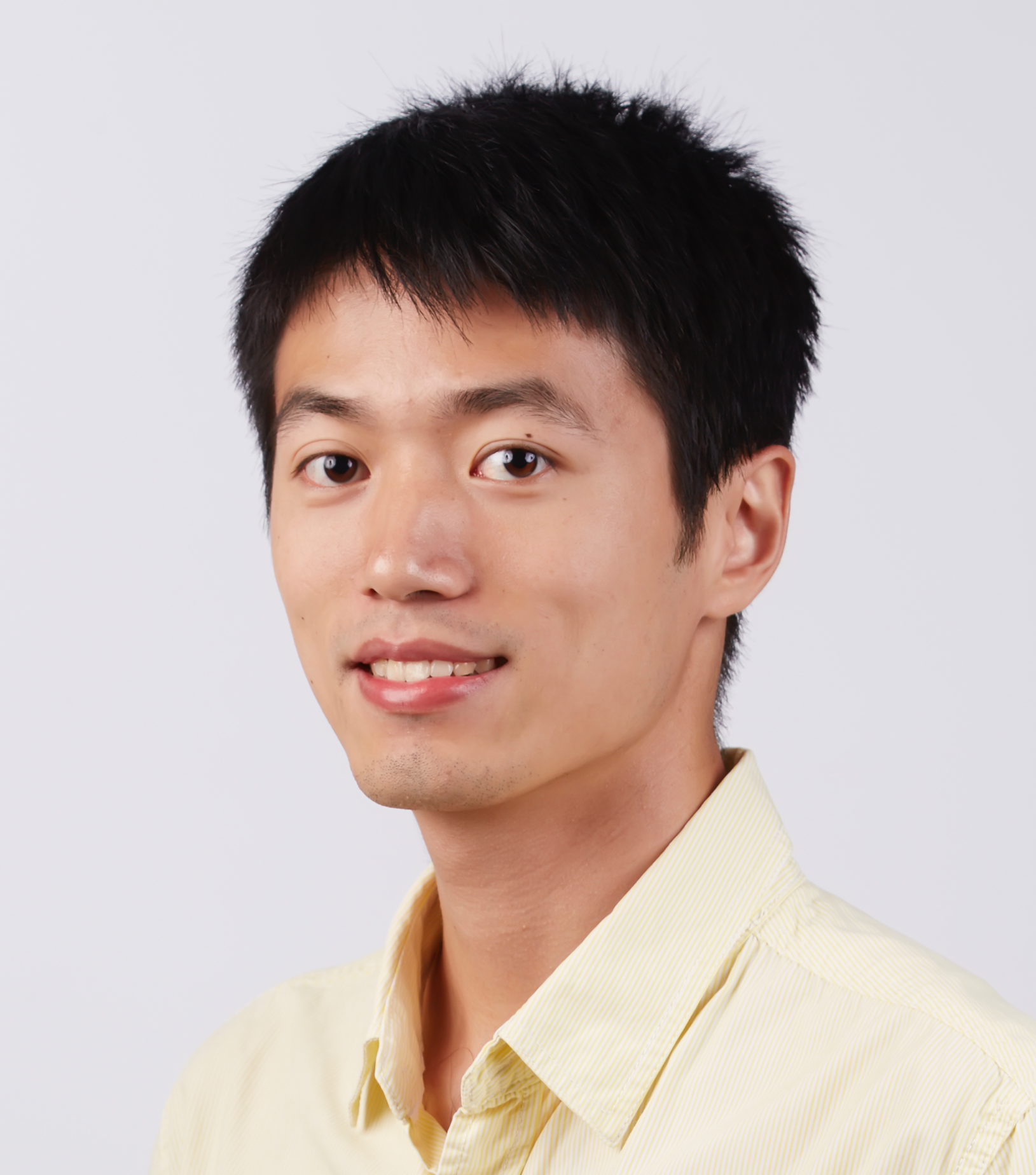}}]{Ziwei Liu}
  is currently an Assistant Professor at Nanyang Technological University, Singapore. Previously, he was a senior research fellow at the Chinese University of Hong Kong and a postdoctoral researcher at University of California, Berkeley. Ziwei received his PhD from the Chinese University of Hong Kong. His research revolves around computer vision, machine learning and computer graphics. He has published extensively on top-tier conferences and journals in relevant fields, including CVPR, ICCV, ECCV, NeurIPS, ICLR, ICML, TPAMI, TOG and Nature - Machine Intelligence. He is the recipient of Microsoft Young Fellowship, Hong Kong PhD Fellowship, ICCV Young Researcher Award and HKSTP Best Paper Award. He also serves as an Area Chair of ICCV, NeurIPS, and ICLR.
\end{IEEEbiography}

\vfill

\end{document}